%% file: main.tex
\documentclass[10pt]{article} 
\usepackage[accepted]{styles/tmlr}

\input{commands/local}
\input{commands/nomclat}

\usepackage{hyperref}
\usepackage{url}
\usepackage{microtype}
\usepackage{graphicx}
\usepackage{booktabs} 
\usepackage{multirow}
\usepackage{adjustbox}
\usepackage{hyperref}
\usepackage{wrapfig}
\usepackage[inline]{enumitem}
\usepackage{tikz}
\usepackage{caption}
\usepackage{subcaption}
\usepackage{graphicx}
\usetikzlibrary{positioning, arrows, shapes,snakes, calc}
\usetikzlibrary{shadows,shadings,shapes.symbols}

 \PassOptionsToPackage{numbers, compress}{natbib}

\usepackage{amsmath}
\usepackage{amssymb}
\usepackage{mathtools}
\usepackage{amsthm}

\usepackage[capitalize,noabbrev]{cleveref}
\theoremstyle{plain}

\theoremstyle{definition}

\theoremstyle{remark}

\usepackage[textsize=tiny]{todonotes}

\title{Optimizing Time Series Forecasting Architectures: \\ A Hierarchical Neural Architecture Search Approach}

\author{%
  \name Difan Deng \email \href{mailto:d.deng@ai.uni-hannover.de}{d.deng@ai.uni-hannover.de} \\
  \addr Institute of Artificial Intelligence \\
  Leibniz University Hannover 
  \AND
  \name Marius Lindauer \email \href{mailto:m.lindauer@ai.uni-hannover.de}{m.lindauer@ai.uni-hannover.de} \\
  \addr Institute of Artificial Intelligence \\
  Leibniz University Hannover \\
  L3S Research Center
}

\newcommand{\textupdate}[1]{\textcolor{blue}{#1}}
\renewcommand{\textupdate}[1]{#1}

\def\openreview{\url{https://openreview.net/forum?id=Ym2wqojm4e}} 

\begin{document}

\maketitle
\begin{abstract}
The rapid development of time series forecasting research has brought many deep learning-based modules to this field. However, despite the increasing number of new forecasting architectures, it is still unclear if we have leveraged the full potential of these existing modules within a properly designed architecture. In this work, we propose a novel hierarchical neural architecture search space for time series forecasting tasks. With the design of a hierarchical search space, we incorporate many architecture types designed for forecasting tasks and allow for the efficient combination of different forecasting architecture modules. Results on long-term time series forecasting tasks show that our approach can search for lightweight, high-performing forecasting architectures across different forecasting tasks.
\end{abstract}

\section{Introduction}
Time series forecasting techniques are widely applied in different fields, e.g., energy consumption~\citep{trindade-uci15a}, business~\citep{makridakis-ijf22a}, or traffic planning~\citep{lana-itsm18a}. However, unlike computer vision (CV) and natural language processing (NLP) tasks that are dominated by the CNN~\citep{he-cvpr16a,  liu-iclr18a, zoph-cvpr18a} and Transformer~\citep{brown-neurips20a, devlin-acl19a, liu-iccv21a, Vaswani-neurips17a} families, there is no clearly dominating architecture in time series forecasting tasks. Although there are lots of transformer-based approaches applied to forecasting models~\citep{ansari-arxiv24a, das-arxiv23a, wu-neurips21a, liu-iclr22a, zhou-aaai21a}, many of them might even be outperformed by a simple linear baseline~\citep{zeng-aaai23a}.

The success of transformer models in CV and NLP tasks is based on their ability to capture long-term dependencies with the help of tokens with fruitful semantic information, i.e., each word embedding already contains lots of information, while an image patch with many pixels can already tell us a lot of information. All this information relaxes the requirement for the models to grab the temporal information within the input series. However, this information is usually crucial in time series forecasting tasks, given that a single value at each time step only contains a limited amount of information. The strong ability of the transformer family to capture long-term dependencies might not compensate for its poor ability to build temporal connections within the input series. This finding is evident in~\citet {zeng-aaai23a}, where a simple linear layer could outperform many state-of-the-art transformer models.

Some recent works show the efficiency of transformers~\citep{liu-iclr24c, nie-iclr23a} on long-time forecasting tasks. However, their approach could still not overcome the temporal dependencies issues for transformers, e.g., PatchTST~\citep{nie-iclr23a} augments the information within each token by constructing a patch with the data from multiple time steps. While iTransformer~\citep{liu-iclr24c} simply encodes the entire input sequence into a token and tries to construct the connections among different variables.

 On the other hand, many architectures were constructed for mining the local dependencies, such as CNNs~\citep{bai-arxiv18a, luo-iclr24a} and RNNs~\citep{hewamalage-ijf21a, hochreiter-icann01a}. These architectures might not work well on long-term dependencies due to the limited receptive field (for CNN), latent bottlenecks~\citep{didolkar-neurips22a}, or vanishing gradients (for RNN families). Recent work, such as ModernTCN~\citep{luo-iclr24a}, showed that an increased convolutional kernel size and improved microarchitecture could lead to a more accurate model. However, this approach results in huge memory and computation consumption. Here we provide another perspective: one could combine these architectures with other operations, such as transformer layers, to develop a new architecture that combines the best of two worlds ~\citep{didolkar-neurips22a, lai-sigir18a, lim-ijf21a}.

Nevertheless, it is still unclear 
\begin{enumerate*}[label=(\roman*)]
\item which type of architecture we would like to construct, given the variability of different forecasting models~\citep{deng-ecml22a, oreshkin-iclr20a, salinas-ijf20a, zeng-aaai23a} and 
\item how to connect different operations to form a new architecture. 
\end{enumerate*}
Designing a new architecture from scratch for each task might take a lot of human expert effort and tedious trial-and-error. Neural architecture search (NAS) is a technique that automatically searches for the optimal architecture given a new task.

Previous NAS frameworks mainly focused on single network backbone types such as CNN or Transformer networks~\citep{ Chen-iccv21b, liu-iclr18a, zoph-cvpr18a}. It is still unclear how to optimize the forecasting architectures due to their internal complexity. For instance, an encoder-decoder architecture~\citep{wu-neurips21a, zhou-aaai21a} might work well on some tasks, while the other tasks might prefer encoder-only architectures~\citep{liu-iclr24c} or even MLP-only architectures~\citep{oreshkin-iclr20a, zeng-aaai23a}. This provides another challenge for designing a search space for time series forecasting tasks. In this work, we will address this challenge by designing a unified search space for time series forecasting tasks.

Our contributions are summarized as follows:
\begin{enumerate}
    \item We design a hierarchical search space that contains most forecasting architecture design decisions and allows any sort of architecture layers to be combined to form new architectures. 
    \item We show that by applying DARTS-PT~\citep{wang-iclr21a}, a differentiable neural architecture search approach, to our search space. The resulting architectures, dubbed \Ourname{}, are comparable to the state-of-the-art models with much less computational resource requirements. 
    \item We provide an analysis of our search space, showing that our search space has different properties compared to the existing CNN-based NAS search space, and provides further challenges for the NAS research. 
\end{enumerate}

\label{submission}

\section{Related Work}
Although many different deep learning architectures are proposed to solve time series tasks, there is little work that applies neural architecture search to search for a new architecture within the existing frameworks. In this section, we will provide a brief overview of deep learning-based forecasting frameworks and neural architecture search techniques.

\subsection{Deep Learning-based Time Series Forecasting\label{sec:dl4ts}}
Time series forecasting aims to predict the future values of target variables given their historical data. Because of its importance, much work has been investigated for a more accurate forecasting model in this research field. Previous work mainly focused on traditional statistical local approaches that train an individual model for each series~\citep{hyndman-book21a, box-book15a}. However, these approaches might not fit well in the era of big data, where a dataset could contain thousands of series. On the other side, the machine learning-based model trains a single global model across all the series and uses this model to predict all the series in the dataset~\citep{godahewa-neuripsdbt21a, makridakis-ijf20a, makridakis-ijf22a}. More recently, deep learning-based forecasting models~\citep{bai-arxiv18a,   hewamalage-ijf21a, lai-sigir18a, salinas-ijf20a, shi-neurips15a,  wen-tsw17a}, or even the zero-shot foundation models~\citep{ansari-arxiv24a, das-arxiv23a}, have gradually become mainstream in this research field.

Time series forecasting models need to work with sequential inputs~\citep{alexandrov-jmlr20a, beitner-github20a}. Overall, these networks can be categorized into two families: $\SeqNet$ and $\FlatNet$~\citep{deng-ecml22a}. Given a batch of sequences with shape $[B, L, N]$, where $B$ is the batch size, $L$ is the sequence length, and $N$ is the number of time series variables. A $\SeqNet$, such as RNNs~\citep{cho-emnlp14a, hochreiter-nc97a}, TCNs~\citep{bai-arxiv18a, Oord-iscassw16a}, and Transformers~\citep{li-neurips19a, liu-iclr22a, Vaswani-neurips17a, wu-neurips21a, zhou-aaai21a}, computes the correlations across different time steps without breaking the structure of the input sequence. On the contrary, a $\FlatNet$, such as MLP~\citep{zeng-aaai23a} and N-BEATS~\citep{oreshkin-iclr20a}, decomposes the variables into independent single-variable series: $[B \times  N, L]$.\footnote{However, in practice, we would still like to preserve the correlations between multi-variant series with, for instance, batch normalization~\citep{ioffe-icml15a}. In this case, the input series becomes $[B, N, L]$ }. After that, this variable is passed to another network. This strategy was previously introduced such that the time series can be handled by the machine learning framework designed for tabular datasets, such as MLP layers~\citep{zeng-aaai23a} and LightGBM~\citep{ke-neurips17a, makridakis-ijf22a}. Some recent works, such as PatchTST~\citep{nie-iclr23a}, release the correlation between different variables within a multi-variable series and consider it as a collection of independent series, and make predictions for each series independently, which also belongs to this type of architecture. However, these approaches might be too expensive for datasets with many variables since a forward pass is required for each series. Hence, in this work, we mainly focus on the MLP-based $\FlatNet$ to only search for lightweight architectures.

Diving deeper into the $\SeqNet$ architecture families, we find another two main branches: encoder-decoder architectures and encoder-only architectures\footnote{Sometimes we might also have decoder-only architectures; however, for the sake of simplicity, we consider both as part of encoder-only architectures.}. Encoder-decoder architectures~\citep{sutskever-neurips14a, Vaswani-neurips17a} maintain two individual networks that embed the information from the past and future correspondingly. These architectures have shown great success in time series forecasting tasks~\citep{wu-neurips21a, zhou-aaai21a}. On the other hand, encoder-only architectures only apply a $\Seq$ encoder that maps the past information into a latent feature map and utilizes another linear layer to provide the prediction with the latent feature map~\citep{liu-iclr24c, nie-iclr23a, salinas-ijf20a}. 

Many $\SeqNet$ models, especially the transformer family~\citep{lim-ijf21a, wu-neurips21a,zhou-aaai21a}, are designed for solving series input. However, a study by~\citet{zeng-aaai23a} showed that these transformers might even be outperformed by a linear model. This inspires us to seek other uncovered modules that might perform well within a properly designed architecture, esp. architectures with more than one type of operation.  

Many forecasting architectures are homogenous and only contain one type of operation layer~\citep{ bai-arxiv18a, li-neurips19a, liu-iclr22a, luo-iclr24a, Oord-iscassw16a} and stack this layer repeatedly to construct a new architecture. However, some recent work has also shown the efficiency of combining architectures from different model families. LSTNet~\citep{lai-sigir18a} stacks an RNN on top of a CNN layer. ConvLSTM~\citep{shi-neurips15a} constructs a convolutionary operation within an LSTM cell. Temporal fusion transformer~\citep{lim-ijf21a} stacks an explainable multi-head attention layer on top of an LSTM encoder-decoder model. All these works suggest the efficiency of combining operations from different architecture families. However, it is often tedious to find these combinations manually. Neural architecture search (NAS) is a technique that automatically searches for the optimal architecture for a given task. In the following section, we will briefly overview the NAS framework.

\subsection{Neural Architecture Search}
Previous NAS research mainly considered network training as a black-box process, training every configuration from scratch until convergence~\citep{deng-ecml22a, jin-sigkdd19a, white-aaai21a, zimmer-tpami21a, zoph-iclr17a}. However, training a network is very expensive and requires lots of resources. To overcome this problem, the One-Shot NAS~\citep{pham-icml18a, zoph-cvpr18a} approach defines a supernetwork where all the child architectures' weights are inherited from the supernet. DARTS~\citep{liu-iclr19a} further relaxes the discrete operation search space to continuous parameters and optimizes these values with model weights jointly with gradient descent. Finally, the optimal operations and paths are selected based on the architecture parameter values. 

DARTS may be unstable during the search process: the architectures suggested by the DARTS might be dominated by skip connections~\citep{chu-eccv20a,  jiang-neurips23a, wang-iclr21a, zela-iclr20a}. Robust DARTS~\citep{zela-iclr20a} showed that this instability is caused by the high validation loss curvature in the search space, while operation-level early stopping DARTS~\citep{jiang-neurips23a} found that the instability arises when the network weights are overfitted to the training sets. 
DARTS-PT~\citep{wang-iclr21a} provides a perturbation-based approach to measure the importance of each operation and use it to replace the architecture parameter-based approach introduced in \cite{liu-iclr19a}. 

ONE-NAS~\citep{lyu-asc23a} is an online architecture search framework that applies evolutionary algorithms to search for the optimal RNN networks on online forecasting tasks. On the other hand, SNAS4MTF~\citep{chen-arxiv21b} proposes to use architecture search to form an end-to-end forecasting architecture framework. Furthermore, Auto-PyTorch TS~\citep{deng-ecml22a}  provides a uniform search space that includes many forecasting modules. The optimizers can freely assemble these modules to form new architectures. However, these works still focus on homogeneous architecture designs, where the type of $\Seq$ decoders is restricted by the decision of $\Seq$ Encoders. For instance, an RNN encoder only allows RNN decoders ($\Seq$ decoder) or MLP decoders ($\Flat$ decoder).  
As the current neural architecture tends to focus more on the one-shot weight-sharing approaches~\citep{jiang-neurips23a, liu-iclr19a, wang-iclr21a, zoph-cvpr18a}, in this work, we will show how to design a general one-shot model for time series forecasting that fits most of the forecasting models, making automated deep learning for time series forecasting substantially faster than before. 

Evaluating an architecture could take lots of resources. One-shot NAS approaches alleviate this by constructing a supernet where the weights of all the candidate architectures are inherited from this super network. However, training a supernet still requires lots of resources. Zero-Cost (ZC) proxies~\citep{chen-openr22a, krishnakumar-neurips22a} attempt to address this problem by estimating the performance of the target model without updating the model weights, thereby greatly accelerating the search process. However, nearly all existing zero-cost proxies~\citep{abdelfattah-iclr22a, lee-iclr19a, lin21-iccv21a, ning-neurips21a, tanaka-neurips20a, turner-19a, wang-iclr19a} are developed for computer vision NAS tasks that only contain one architecture type (convolutional operations). It is still unclear if the ZC proxies can be generalized to compare the performances between different operation types.

\section{Problem Setting}
Time series forecasting tasks aim to predict the values of the target variables for a number of iterations after a certain time step, with the observed same variables and several other feature variables. Formally, given a dataset that is composed of multiple series $\dset = \{\dsetseq\}_{\seqidx=1}^\numseq$, where each series is composed of the past observed targets $\targetseq{\observedintervalseq}$, past observed features $\featurespastseq{\observedintervalseq}$, and known future features $\featuresfutureseq{\forecastintervalseq}$. Given a required forecasting horizon $\horizon$, the model is asked to predict the target variables $\predictionseq{\forecastintervalseq}$ with all the available information:
\begin{equation}
    \predictionseq{\forecastintervalseq} = f(\targetseq{\observedintervalseq}, \featurespastseq{\observedintervalseq}, \featuresfutureseq{\forecastintervalseq}; \weights )
\end{equation}

Two model families can mainly handle multi-horizon forecasting tasks: the autoregressive approach and the non-autoregressive approach. Autoregressive approaches~\citep{box-book15a, salinas-ijf20a} only predict one step within one forward pass and iteratively use the predicted value as the known feature that can be further fed to the model. On the other hand, non-autoregressive models~\citep{nie-iclr23a, zeng-aaai23a, zhou-aaai21a} directly generate multiple forecasting values within one forward pass. In this work, we mainly focus on searching for non-autoregressive architectures. 

Neural architecture search aims at finding the optimal architecture $\archp_{*}$ for a given task:  
\begin{align}
    \min_{\archp} \mathcal{L}_{val}(\weights^{*}, \archp)\ \ \ \ \mathrm{ s.t.   }\ \ \ \ \ {\weights^{*}} \in \argmin_{\weights}\mathcal{L}_{train}(\weights, \archp)
\end{align}

The search space in previous NAS work still focused on the homogeneous search space, where all the operations belong to the same architecture families. This work presents a unified heterogeneous search space containing most of the potential architectures applied for time series forecasting tasks.

\section{Search Space Design~\label{sec:searchspacedesign}}

It is a common observation and a foundational assumption of NAS that no single architecture family always outperforms the others; based on the results of \citet{deng-ecml22a}, we believe that this also holds for time series forecasting tasks. We are unlikely to define a single type of model that works for all datasets. Additionally, some of the architecture might be dependent on the other decision choices. For instance, if we decide to have an encoder-only $\SeqNet$, there is no need for us to search for a $\Seq$ decoder. Here, we propose a hierarchical search space that incorporates most of the forecasting architecture families described in Section~\ref{sec:dl4ts}. We will start from the most basic operation level and gradually decrease the granularity until our search space contains all the required components. 

\subsection{Operation Level\label{sec:ssd_ol}}

\begin{wrapfigure}[17]{r}{0.5\textwidth}
\centering
    \begin{subfigure}[t]{0.225\textwidth}
    \centering
        \input{figures/tikz/search_space/cells_searchspace}
\end{subfigure}
    \begin{subfigure}[t]{0.225\textwidth}
        \centering
        \input{figures/tikz/search_space/cell_searched}
\end{subfigure}
    \caption{Operation level Search Space. The nodes $0$ and $1$ are input nodes that receive the network inputs or the outputs from the other cells. Node $4$ is the output node. Each colored edge represents an operation. The Search space (Left) is a fully connected directed acyclic graph. Once we have finished the search, we get the final architecture (Right).}
    \label{fig:ss_operations}
\end{wrapfigure}

The first level, the operation level, describes the operations that can be used in the network. 
As shown in Figure ~\ref{fig:ss_operations}, our search space follows the DARTS~\citep{liu-iclr19a} search space, a cell-based architecture where each cell is a directed acyclic graph that contains $N$ nodes, including $N_{in}$ input nodes. Each node represents a latent feature map, and the edges that connect the nodes are the operations applied to the latent feature maps. DARTS defines two types of cells: normal cells and reduction cells. These two cell types share the same form of input feature maps. Therefore, they can be easily concatenated to form an architecture. However, this is not the case for forecasting tasks. $\SeqNet$ and $\FlatNet$ transform the input features differently. We cannot easily connect a $Seq$ cell after a $Flat$ cell and vice versa. Hence, we provide a search space for each of the model families. Detailed information about each operation can be found in the appendix~\ref{sec:nn_components}.

For the $\SeqNet$ families, we consider the following operations:
\begin{enumerate*}
    \item MLPMixer~\citep{chen-tmlr23a}, an all-MLP architecture that applies a linear layer to feature and time dimensions, respectively,
    \item LSTM~\citep{hochreiter-nc97a},
    \item GRU~\citep{cho-emnlp14a},
    \item Transformer~\citep{Vaswani-neurips17a},
    \item TCN~\citep{bai-arxiv18a}, 
    \item SepTCN~\citep{luo-iclr24a},
    and
    \item skip connections.
\end{enumerate*}
These operations are spread to encoder and decoder architectures, which will be discussed in the following section.

For the $\FlatNet$ families, we have: 
\begin{enumerate*}
    \item Linear~\citep{zeng-aaai23a}, a single linear layer that encodes the past information to the future variable,
    \item N-BEATS~\citep{oreshkin-iclr20a}, a repeatedly stacked MLP block, where each block contains a set of fully connected (FC) backbone layers, a forecast, and a backcast head, and
    \item skip connections.
\end{enumerate*}
There are several variants in N-BEATS modules: generic model, trend model, and seasonality model. We incorporate all these modules into our search space, providing another two operations. Since the only difference between different N-BEATS modules is their forecast and backcast head, we ask these N-BEATS modules to share the same FC layers backbones. In total, we have five operations for each edge in the $\FlatNet$ cell.

Unlike the previous search space that focused on one single architecture type~\citep{klyuchnikov-arxiv20a, krishnakumar-neurips22a, mehta-iclr22a,ying-icml19a}, the operations contained in ~\Ourname{} come from different families. Therefore, the outputs from each operation might have completely different distributions or even different output formats for the same input. The zero-cost proxies that are proven to be efficient on other benchmarks, such as ~\textit{flops} or ~\textit{number of parameters}~\citep{krishnakumar-neurips22a, wan-iclr22a}, might no longer work well within this search space.

\subsection{Micro Network Level~\label{sec:micro_level}}

$\Flat$ operations only receives the past information $\targetseq{\observedintervalseq}$. Therefore, we stack several $\Flat$ cells as a $\FlatNet$. As shown in Figure~\ref{fig:search_space_flat}, the past targets are first transposed and then fed to the encoder layers. The transposed target, i.e., the backcast part, and a zero tensor whose length is equal to the forecasting horizon, i.e., the forecast part, are fed to the $\Flat$ encoders. Finally, the forecast output is fed to the forecasting head to predict the target values.\footnote{We note that the head here does not necessarily need to be a network module since $\Flat$ net only predicts one variable each time.}

Different from the $\FlatNet$, we decompose the $\Seq$ architectures into two parts: encoders and decoders. The encoders encode the past observed values into an embedding and feed them to the decoder networks. The design of the $\Seq$ encoder is similar to the $\Flat$ Encoder, and we stack the encoder cells to form the encoder network. However, as discussed in Section~\ref{sec:dl4ts}, two potential ways exist to transform the encoder latent features to the forecasting heads. Therefore, we design the following two types of decoder networks: $\Seq$ decoder and $\Flat$ decoder for $\Seq$ encoder. 

The design of the $\Seq$ Encoder and $\Seq$ Decoder architecture has been widely applied in previous architecture works~\citep{lim-ijf21a, sutskever-neurips14a, Vaswani-neurips17a}. However, architecture decoders might require different information from the encoder network. For instance, a Transformer decoder~\citep{Vaswani-neurips17a} only requires the output from the last layer of the corresponding encoder; an RNN decoder would need the hidden states from the corresponding encoder layer. We record the following information stored by each edge:
\begin{enumerate}
    \item The output hidden feature map of this edge. It will then be contacted with the corresponding decoder feature maps and fed to the TCN decoder network.
    \item The last step's feature map. This value is considered a hidden state that can be fed to the corresponding GRU and LSTM layers to initialize their states. 
    \item The cell gate state that is applied to initialize the hidden cell states of the corresponding LSTM layer. If the encoder is not an LSTM network, following the idea of stitchable network~\citep{pan-cvpr23a} that uses a linear layer to stitch two networks with different shapes, we use a linear layer that transforms the hidden states into the cell gate states.
\end{enumerate}
Hence, we record all the related information as intermediate states during the forward pass. This information is then used to inform the decoder networks of the information provided by the encoder. 

On the other hand, we might not want a complex decoder architecture since the information provided to the decoder and most of the features presented in the past are no longer available in the future. Hence, here we define another type of decoder for $\Seq$ network: the linear (or flat) decoder for $\SeqNet$. We apply one linear layer across the time series dimension~\citep{chen-tmlr23a, zeng-aaai23a} that maps the encoder output and the available future information to the decoder output feature. This feature is then fed to the forecasting head to generate the final prediction result. The overall $\SeqNet$ architecture is presented in Figure~\ref{fig:search_space_seq}.

We also consider the choice of the $\SeqNet$ decoder as part of the architecture search procedure. Hence, we assign another set of architecture parameters to the output of the two decoder architectures. This architecture is jointly optimized with the other operations architecture weights introduced in Section~\ref{sec:ssd_ol}.

\begin{figure}[t]
\centering
    \begin{subfigure}[t]{0.485\textwidth}
       \input{figures/tikz/search_space/flat_net}
       \caption{Search Space for $\Flat$ cells\label{fig:search_space_flat}}
    \end{subfigure}
    \begin{subfigure}[t]{0.485\textwidth}
\centering
    \input{figures/tikz/search_space/seq_net}
    \caption{Search Space for $\Seq$ cells\label{fig:search_space_seq}}
    \end{subfigure}

    \caption{Micro-level search space design for Flat and Seq cells.\label{fig:ss_micro}}

\end{figure}

\subsection{Macro Architecture Level~\label{sec:macro_level}}

The $\FlatNet$ and $\SeqNet$ families defined above encode the input data from different perspectives: $\FlatNet$ families encode the input series along the time series dimension, while $\SeqNet$ families encode the input series across different variations. Hence, the two architectures can complement each other, and we concatenate the two architectures sequentially. 

\begin{wrapfigure}[9]{r}{0.6\textwidth}
\vspace{-25pt}
\centering
       \input{figures/tikz/search_space/hybrid}
       \caption{ Marco Search Space. \label{fig:ss_macro}}
\end{wrapfigure}

As shown in Figure~\ref{fig:ss_macro}, the past target values are first fed to the $\FlatNet$, which results in a backcast and forecasting feature maps. The forecast feature maps are then concatenated with the known future features and fed to the $\Seq$ decoder. Finally, the final forecasting result is the weighted sum of both $\FlatNet$ and $\SeqNet$\footnote{For the sake of simplicity, we omit the feature variables $\featurespastseq{\observedintervalseq}$ and $\featuresfutureseq{\forecastintervalseq}$ that are fed to the $f_{model}$ and $f_{seq}$}:

\begin{equation}
     f_{model}(\targetseq{\observedintervalseq}) = w_{seq}f_{seq}(\targetseq{\observedintervalseq}, f_{flat}(\targetseq{\observedintervalseq}))  + w_{flat}f_{flat}(\targetseq{\observedintervalseq})
 \end{equation}

These weights are considered an architecture parameter that can be jointly learned with the other architecture parameters described in the aforementioned sections.

\section{Searching for the Optimal Architectures~\label{sec:search_strategy}}
Here, we provide an exemplary searching strategy, i.e., DARTS~\citep{liu-arxiv19a, wang-iclr21a}, to search within our search space. DARTS assigns a weight for each of the operations within its search space and optimizes these architecture weights jointly with the model weights using gradient descent. Some of the modules, such as Dropout~\citep{srivastava-jmlr14a} and Batch Normalization~\citep{ioffe-icml15a}, behave differently during training and inference time. We thus switch these modules to the $eval$ mode when we update the architecture weights with validation losses to simulate the evaluation process. 

During the search phase, we divide the official training/validation splits of the raw dataset into training and validation sets of the same size, such that they do not overlap with the test sets. The training set and validation sets are then used to optimize the architecture weights and architecture parameters, respectively. We follow the common practice for the training-validation split in time series forecasting tasks: the validation set is located at the tail of the training set. The first half of the dataset is considered the training set that is used to optimize the weights of the supernet, while the second part of the dataset is the validation set, which is used to optimize the architecture parameters. We apply RevIN~\citep{kim-iclr22a} during both the searching and training phases to ensure that the input features fed to the networks stay in the same distribution. 

\subsection{Hierarchical Pruning of the One-Shot Model}

Vanilla DARTS are shown to be unstable during the search process and might prefer to select architectures that are dominated by skip connections~\citep{chu-eccv20a, jiang-neurips23a, wang-iclr21a, zela-iclr20a}. Hence, once the weights and architecture parameters are trained, as the last step, we select the optimal operations and edges using the perturbation-based approach~\citep{wang-iclr21a}. 

Given that our search space is a hierarchical search space, some of the operations might be dependent on others, and we have to consider that for the pruning phase. For instance, if we select a Linear Decoder, then we do not need to further select the operations in the $\Seq$ decoder. However, on the other hand, this also indicates that our estimate will be biased if we select the choice of decoder before selecting any edge operations in the $\Seq$ Decoder. Hence, we propose pruning our network from the lowest granularity level and gradually increasing the granularity level until we prune the operations in our search space.
Once all the operations are selected, we further prune the edges of our network using the same perturbation-based approach. Finally, only two edges are preserved for each node, including the cell output node.\footnote{We note that this setting is different from the traditional NAS framework, where all the edges towards the output nodes are preserved. This approach helps us to reduce the architecture size and required latency further.}

\section{Experiments~\label{sec:exp}}
In this section, we first demonstrate that \Ourname{} can identify the optimal architecture, which is comparable to many other handcrafted architectures on various datasets. We then provide a brief analysis of our search space to show the potential challenges in our search space. After that, we provide an analysis of the latency of the optimal architecture. Finally, we show some optimal architectures as examples. 
\subsection{Time series forecasting tasks}
We evaluate our architecture search framework on the popular long-term forecasting datasets introduced by \citet{wu-neurips21a} and  \citet{zeng-aaai23a}: Weather, Traffic, Exchange Rate (Exchange), Electricity (ECL), and four ETT datasets. Additionally, we evaluate our approach on the four PEMS datasets ~\citep{chen-trr01a} that record the public traffic network data in California.

\begin{table}[t]
    \centering
    \scalebox{0.7}{\input{tables/mean_table}}
    \caption{The results of the mean performance over all the long-term forecasting datasets. The best models are marked in red, while the second-best model is marked with underlines.}
    \label{tab:mean_res}
\end{table}

\begin{table}
    \centering
    \scalebox{0.7}{\input{tables/mean_pems}}
    \caption{The results of the mean performance over all the PEMS datasets. The best models are marked in red, while the second-best model is marked with underlines.}
    \label{tab:mean_res_pems}
\end{table}

Following the experiment setup from the other works, we set the forecasting horizon $\horizon$ for ETT, ECL, Exchange Rate, Weather, and Traffic datasets as $\{96, 192, 336, 720\}$. For the four PEMS datasets, we set these values as $\{12, 24, 48, 96\}$.  
We compare our results with the following baselines: PathTST~\citep{nie-iclr23a}, 
ModernTCN~\citep{luo-iclr24a},
DLinear~\citep{zeng-aaai23a}, TSMixer~\citep{chen-tmlr23a}. iTransformer~\citep{liu-iclr24c}, Autoformer~\citep{wu-neurips21a} and TimesNet~\citep{wu-iclr23a}.
For the sake of fair comparison, we follow the setup from PatchTST~\citep{nie-iclr23a} and set the input sequence length for all models to $336$. \textupdate{The training/validation/test splits of each dataset follow the rules defined in the Time Series Library\footnote{\url{https://github.com/thuml/Time-Series-Library}}, where a fixed ratio of the series is described as test sets, i.e., each test set might contain more time steps than the required forecasting horizons. i.e., we perform rolling origin evaluation by continuously moving the forecasting origin until we iterate through the entire test set.} We ran each experiment 5 times with different seeds and recorded their mean and standard deviation, respectively.\footnote{Our code can be found on \url{https://github.com/automl/OneShotForecastingNAS.git}.} \textupdate{Further experimental details, including the datasets information, searching and evaluation costs, hyperparameter settings, and GFlops changes after each stage, can be found under the appendix ~\ref{sec:exp_detail}. }


The results for long-term forecasting tasks are shown in Table~\ref{tab:mean_res}. The best results are marked in red. The full results can be found in the appendix ~\ref{sec:full_res}, ~\textupdate{where we also provide a statistical analysis against the best baseline models on each dataset}. Our network achieves the best or comparable results on the ECL, ETTm, Traffic, and Weather Datasets. Overall, we show that \Ourname{}
 automatically found architectures that are comparable to or better than many other hand-crafted architectures specifically designed for forecasting tasks with only the vanilla series modules. 

 While on another benchmark, the PEMS dataset, \Ourname{} outperforms all the other baselines for all the tasks, as shown in Table~\ref{tab:mean_res_pems}.  This shows that \Ourname{}  could adapt to different tasks and suggests optimal architectures for different time series distributions.

\subsection{Search Space Analysis~\label{sec:search_space_analysis}}
Unlike the search space in other tasks, which contains only a single operation type~\citep{klyuchnikov-arxiv20a, krishnakumar-neurips22a, mehta-iclr22a, ying-icml19a}, our search space encompasses operations from different architectures. This differs from the previous architecture search space and may present additional challenges to the optimizers. To provide a basic understanding of our search space, we randomly sample $3,000$ configurations from our search space and evaluate them on the ECL and Traffic datasets. 

\begin{figure}[h]
    \centering
    \includegraphics[width=0.3\linewidth]{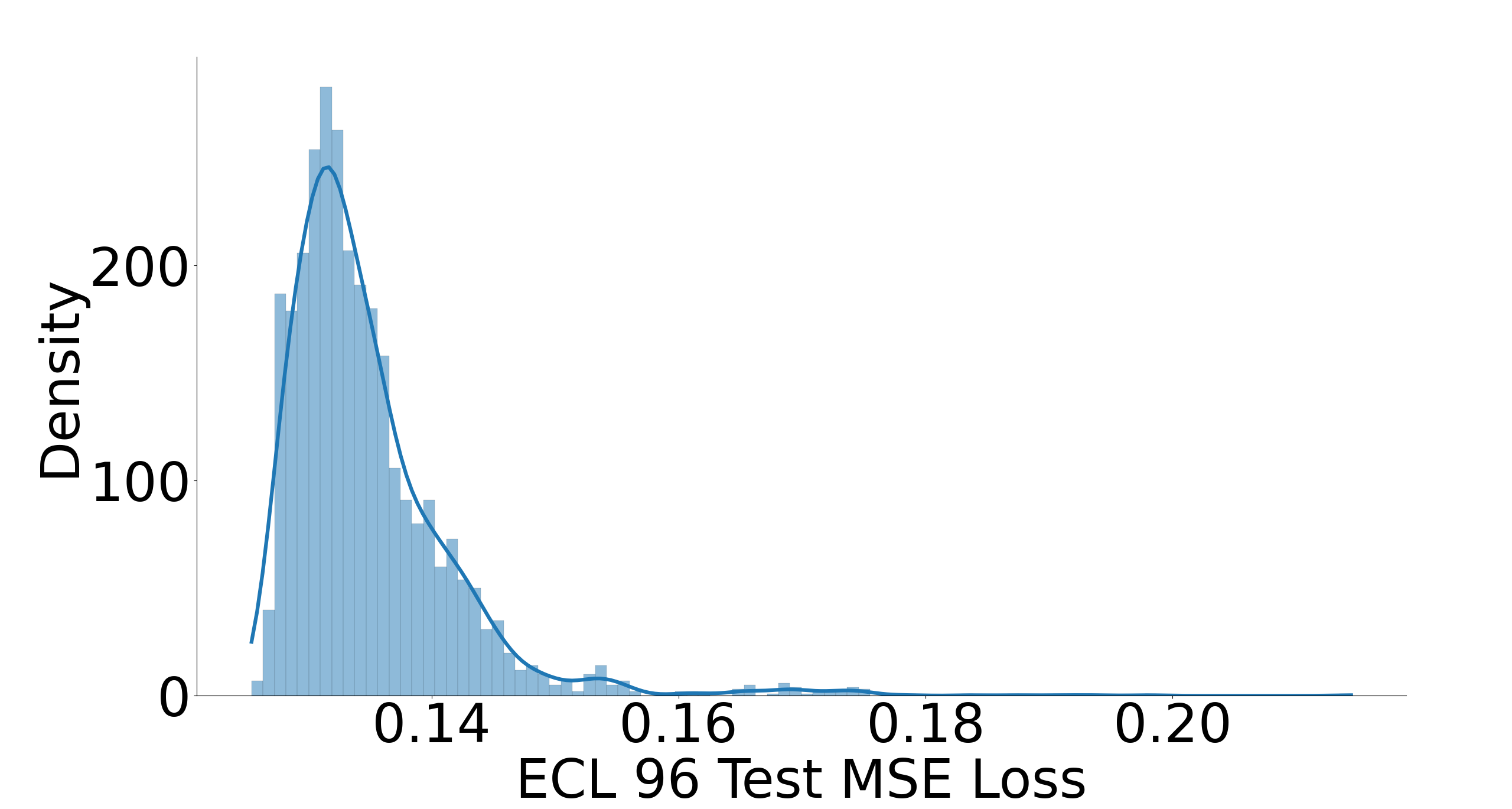}
    \includegraphics[width=0.3\linewidth]{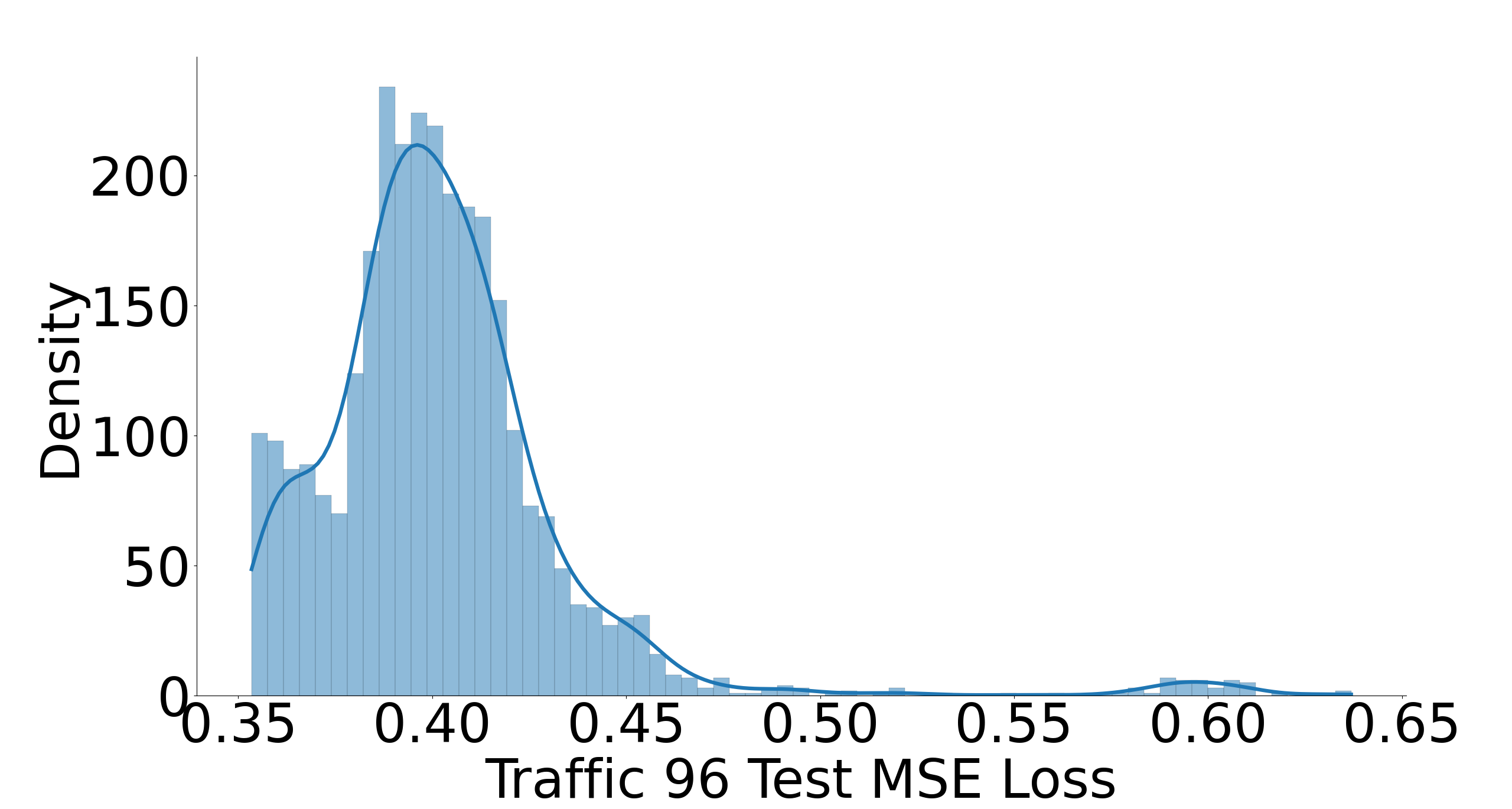}
    \includegraphics[width=0.3\linewidth]{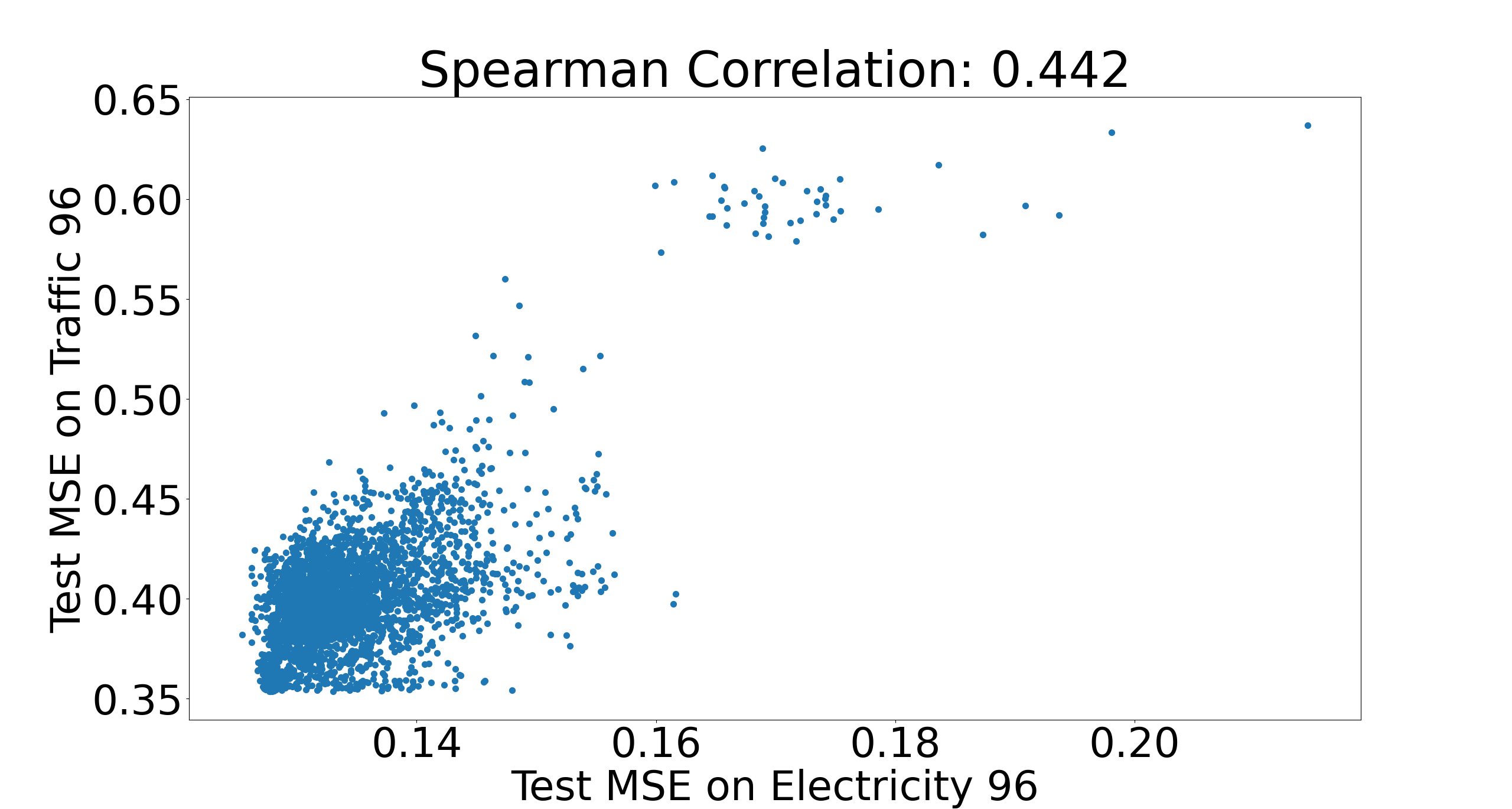}
    \caption{Random Search Evaluation Results on the ECL and Traffic datasets. (Left), MSE loss distributions on the ECL dataset. (Middle), MSE loss distributions on the Traffic dataset. (Right), loss distributions on the Traffic ECL dataset from the same architecture configurations}
    \label{fig:RandomStats}
\end{figure}

The random configuration performances are shown in Figure~\ref{fig:RandomStats}. For the same set of hyperparameter configurations, the MSE losses on these two tasks follow different distributions: losses on the ECL datasets are more skewed towards the lower bounds, while the losses on the Traffic datasets are more uniformly distributed. This indicates different difficulty levels for the optimizers to search for the optimal architectures on different datasets. To check whether the performance of the same hyperparameter configurations could transfer to different tasks, we plot the performance of the same configurations from the two tasks in the right part of Figure~\ref{fig:RandomStats}. The overall trend shows the consistent performance of the worst-performing configurations on both datasets. However, the performance diverges as we move towards the near-optimal configurations. Hence, the optimal configuration for one task may not work equally well for another task. There is still a need to further search for the optimal model on the target dataset. However, since the worst configurations perform consistently on the two tasks, we could incorporate the runs from other tasks as a prior ~\citep{hvarfner-iclr22a, malik-neurips23a}.

Operations in our search space come from different architecture types. This provides further challenge for the optimizers as the performance difference among the operations in the search space can no longer be described with simple proxies. For instance, in the computer vision NAS benchmark search space, a $3\times 3$ convolutional layer could, in most cases, achieve a better performance than a $1\times 1$ layer, as the \textit{number of parameters} of $3\times 3$ convolutional layer is larger than that of $1\times 1$ convolutional layer. However, it is questionable if a Transformer layer (with $12 d^2$ parameters) will outperform an LSTM layer (with $8 d^2$ parameters) since these two operations follow different computational rules. Hence, the widely applied zero-cost proxies applied in the computer vision NAS benchmarks might no longer work in our search space.

\begin{figure}
    \centering
    \includegraphics[width=0.75\linewidth]{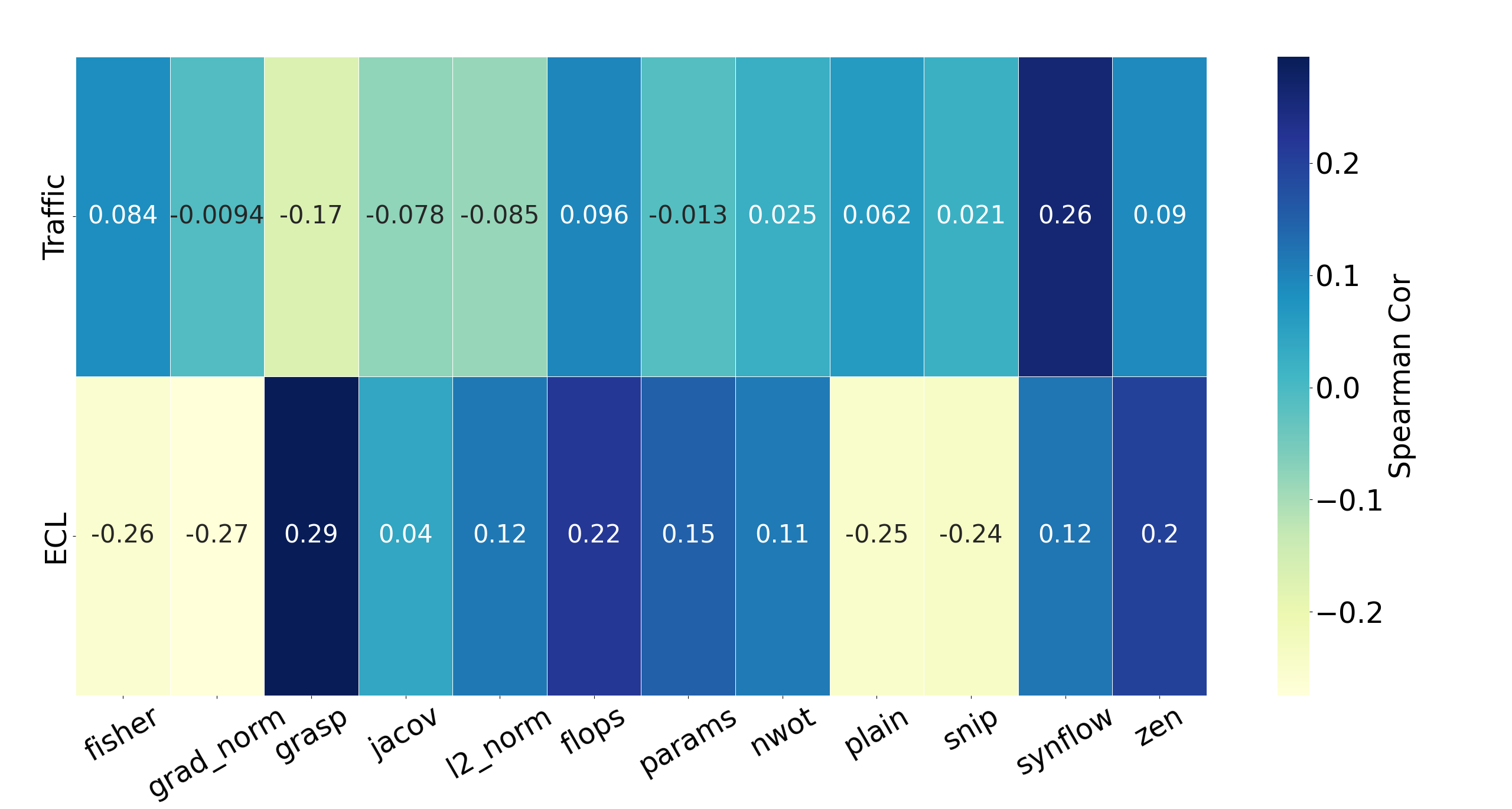}
    \caption{Spearman correlation between different ZC metrics and the evaluation test MSE losses}
    \label{fig:zc-spearman}
\end{figure}

We computed the zero-cost (ZC) metrics for the sampled architectures. Following the settings from NAS-Bench-Suite-Zero~\citep{krishnakumar-neurips22a}, we evaluate the following zero cost proxies: fisher~\citep{turner-19a}, flops, number of parameters~\citep{ning-neurips21a}, grad-norm, l2-norm, plain~\citep{abdelfattah-iclr22a}, grasp~\citep{wang-iclr19a},  jacov, nwot~\citep{mellor-icml20a}, snip~\citep{lee-iclr19a}, synflow~\citep{tanaka-neurips20a}, and zen-score~\citep{lin21-iccv21a}. We evaluate the ZC scores of all the sampled models and compute the Spearman correlation between the ZC scores and their test performance. The results are shown in Figure~\ref{fig:zc-spearman}. Supervisingly, even though many zero-cost proxies do not limit their application to pure CNN or image classification tasks, nearly all of the zero-cost proxies fail in our search space. This indicates that the existing zero-cost proxies might not generalize well to more complex search spaces. Additionally, as shown in Figure~\ref{fig:RandomStats}, even the same configuration might perform differently on different datasets. This further shows the necessity of defining new data-dependent zero-cost proxies that work well across different architecture types. 

\begin{figure}[h]
    \centering
    \includegraphics[width=0.45\linewidth]{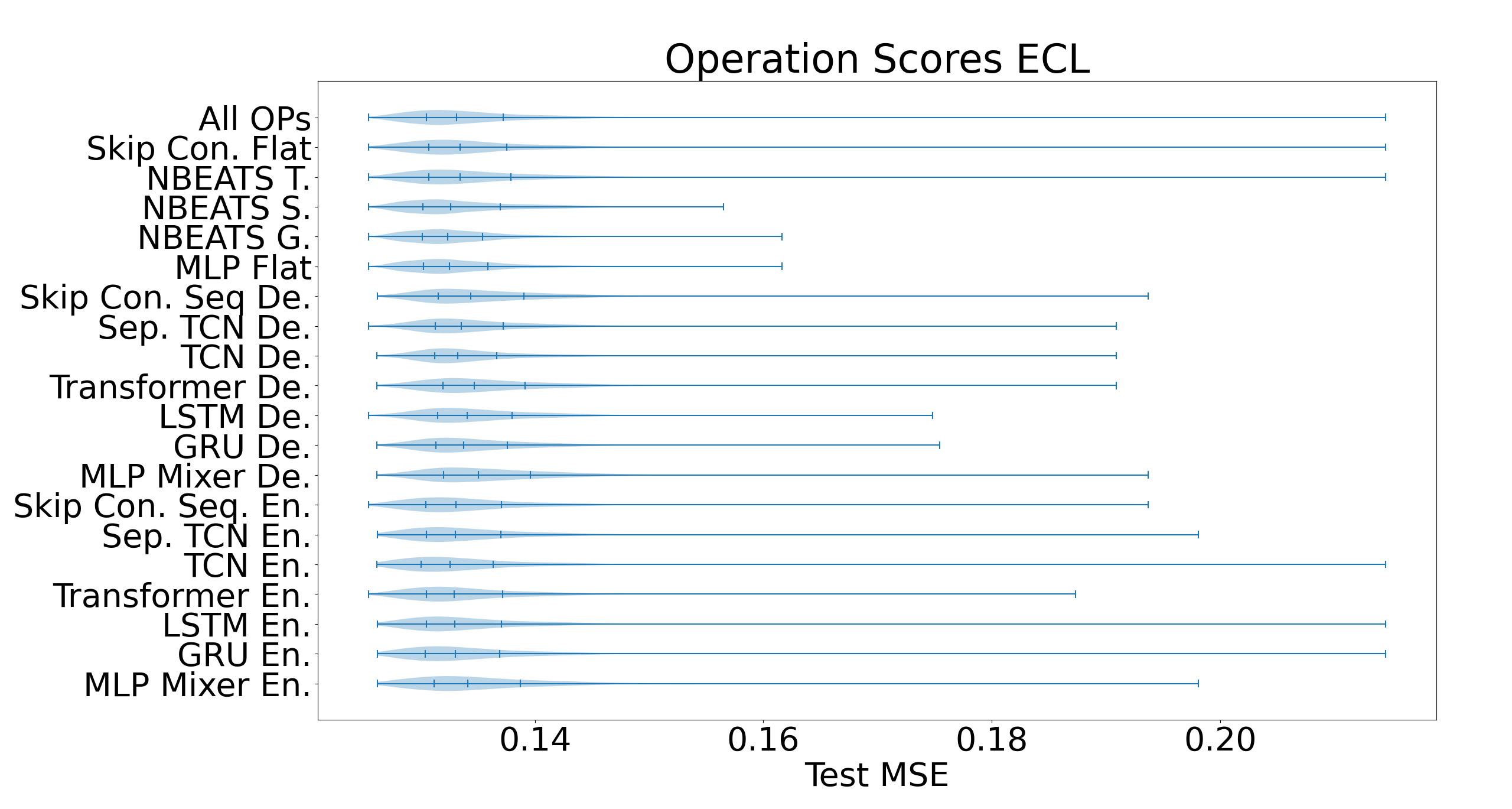}
    \includegraphics[width=0.45\linewidth]{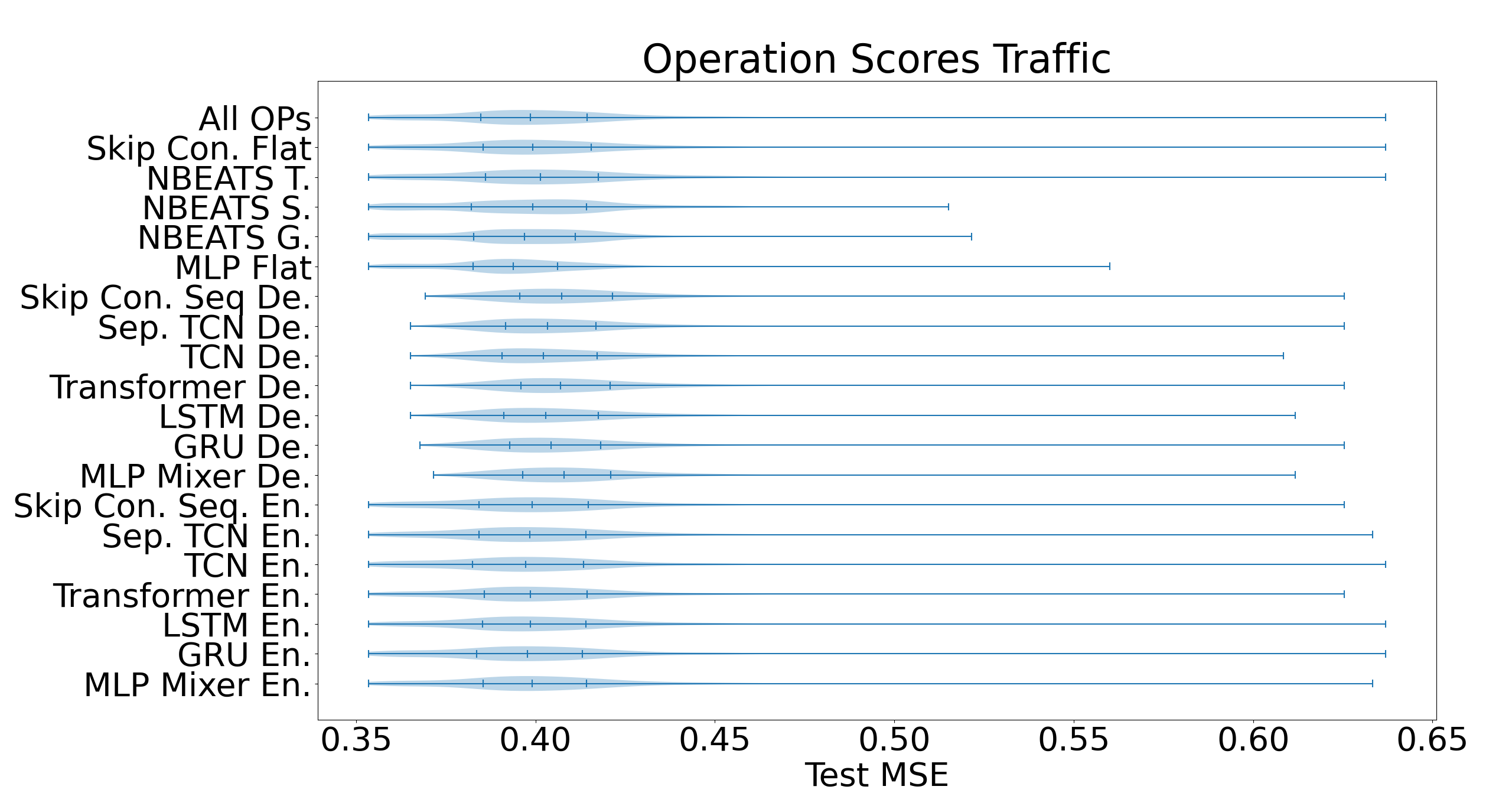}
    \caption{Performance of the randomly sampled models that contain at least one operation}
    \label{fig:randomOpsPerformance}
\end{figure}

To provide further insight into the design of ZC proxies for our search space, we check the performance of the architectures that contain at least one operation within our search space~\citep{lopes-arxiv23a}. This provides a preliminary estimation of the strength of each operation on the architecture performance. The result is shown in Figure~\ref{fig:randomOpsPerformance}. Although the sequential network with sequential decoder families performs similarly to the sequential network with linear decoder on the ECL dataset, the gap becomes larger when the same architecture is evaluated on the Traffic dataset. However, the linear decoders might also lead to a much worse model, as the worst-performing linear decoders' loss is much higher than that of the sequential decoders.

Among the sequence operations, the TCN families (TCN and separated TCN models) achieve better median and 25th quantile performance on both datasets, which indicates that the dataset requires the models to focus more on local dependencies. While transformer families dominate the other benchmark, their median and 25th quantile losses are higher than those of the other operations. Although the MLP mixer encoders are quite close (and sometimes are even optimal) to the other operations, MLP Mixer decoders perform worse than the other sequential decoding operations. This highlights the importance of incorporating various encoder-decoder architectures into forecasting architecture designs.  

For flat operations, MLP Flat and N-BEATS-Generic achieve better performance compared to N-BEATS-Seasonal and N-BEATS-Trend models. However, the top-performing N-BEATS-Seasonal models achieve a lower loss compared to the top-performing MLP flat layers. This might indicate that the inductive bias contained in the N-BEATS-Seasonal models, i.e., the forecasting results are periodic, would help the model achieve better performance.

\begin{figure}
    \centering
    \includegraphics[width=0.45\linewidth]{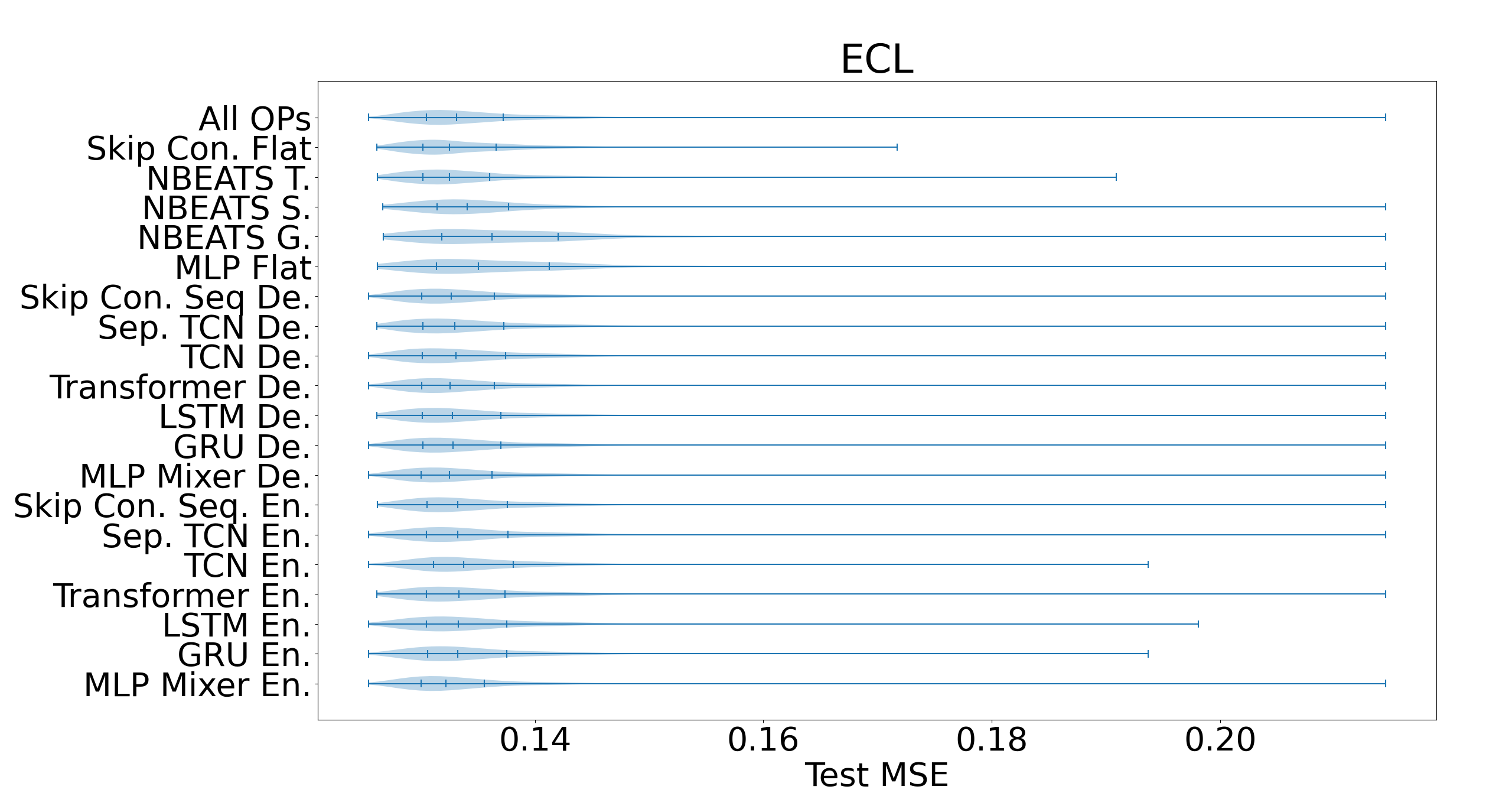}
    \includegraphics[width=0.45\linewidth]{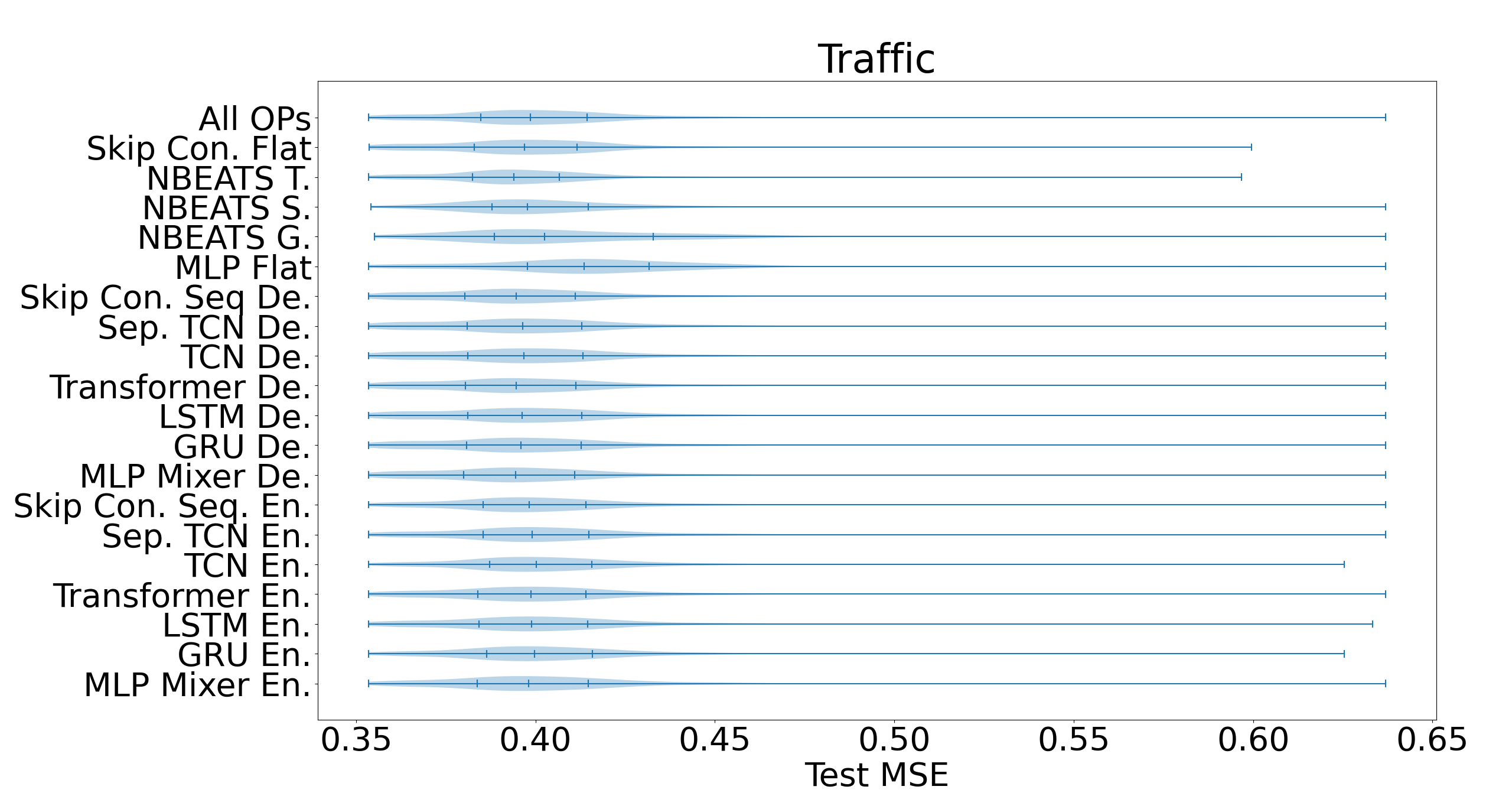}
    \caption{Performance of the randomly sampled models that do not contain the specific operation
    \label{fig:randomWOOpsPerformance}}
\end{figure}

We further show the scores of the architectures that do not contain the target operations in Figure~\ref{fig:randomWOOpsPerformance}. In this case, a higher score indicates that the target function is more important for the target task. Results show that many operations have different impacts on the searched architectures. For instance, missing N-BEATS-Trend results in a poorer optimal architecture for the ECL dataset, while a missing separated TCN results in a poorer mean-performing architecture on the same dataset. For the traffic dataset, removing the same operation could even yield a better-performing set of architectures. This shows that even the same operation might have a completely different influence on the architectures in the search space.

\subsection{Model Efficiency Analysis~\label{sec:model_eff}}
The growing demand for forecasting models has posed more challenges to the forecasting networks: the network should be fast, so that it can quickly predict the following trend. Additionally, networks need to contain fewer parameters and consume less memory so we can deploy them on embedded systems. 

To further show that our network could find an efficient and strong network, we ask all the networks to do a single forward pass and backpropagation with the series within the Traffic and ECL dataset, where each series contains 862 and 321 variables, respectively. We set the batch size of the series to 32 and the look-back window size to 96. The networks are then asked to predict the future series with a forecasting horizon of 96. This experiment is executed on one single Nvidia 2080 TI GPU with 11 GB GPU RAM\footnote{However, we search the one-shot model on an Nvidia A100 GPU with 40 GB GPU RAM.}. Due to this memory constraint, some of the networks, such as ModernTCN, cannot fit into this GPU with our setup and do not appear in this comparison. 

\begin{figure}[t]
    \centering
    \includegraphics[width=0.485\textwidth]{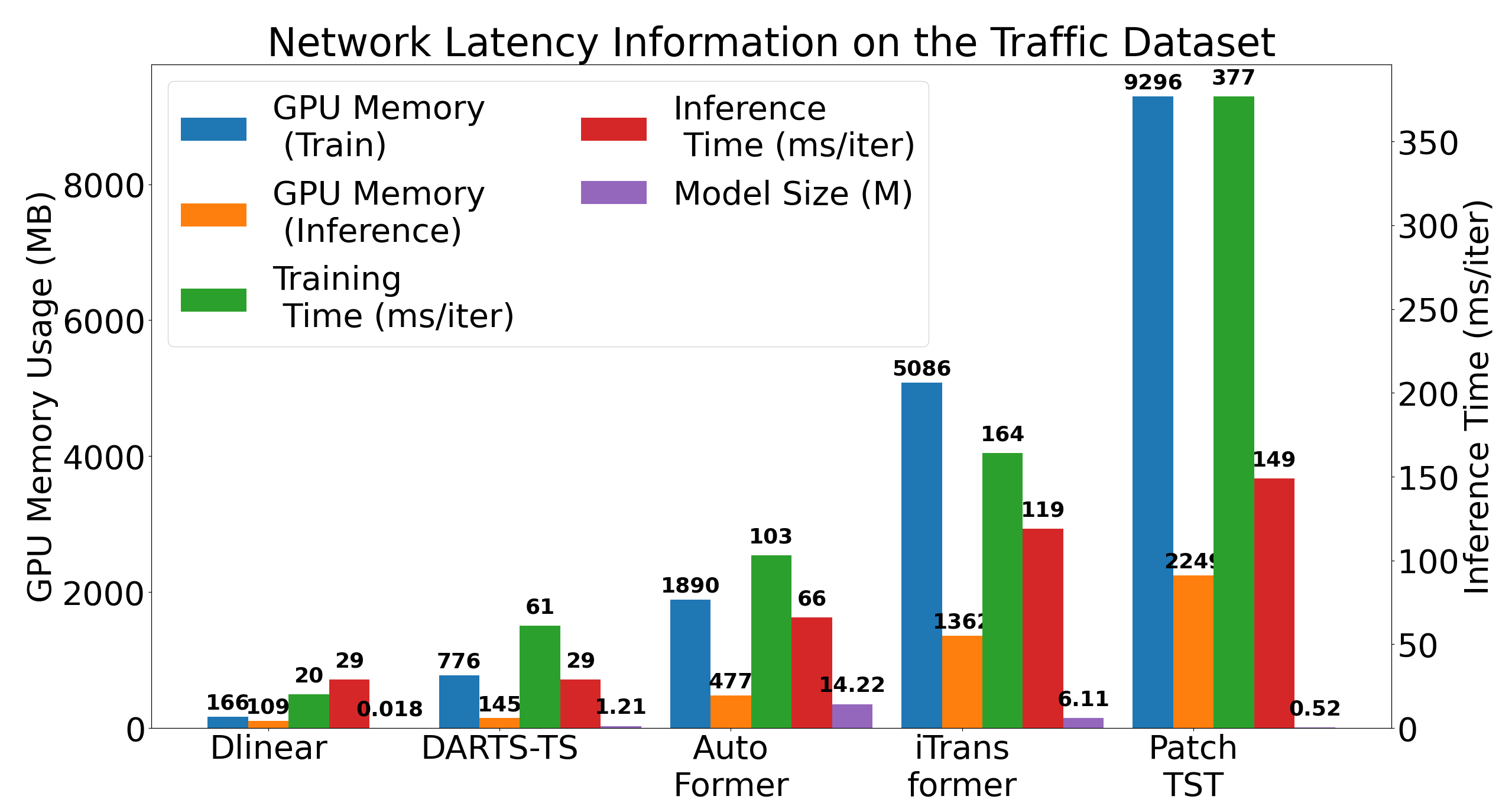}
    \includegraphics[width=0.485\textwidth]{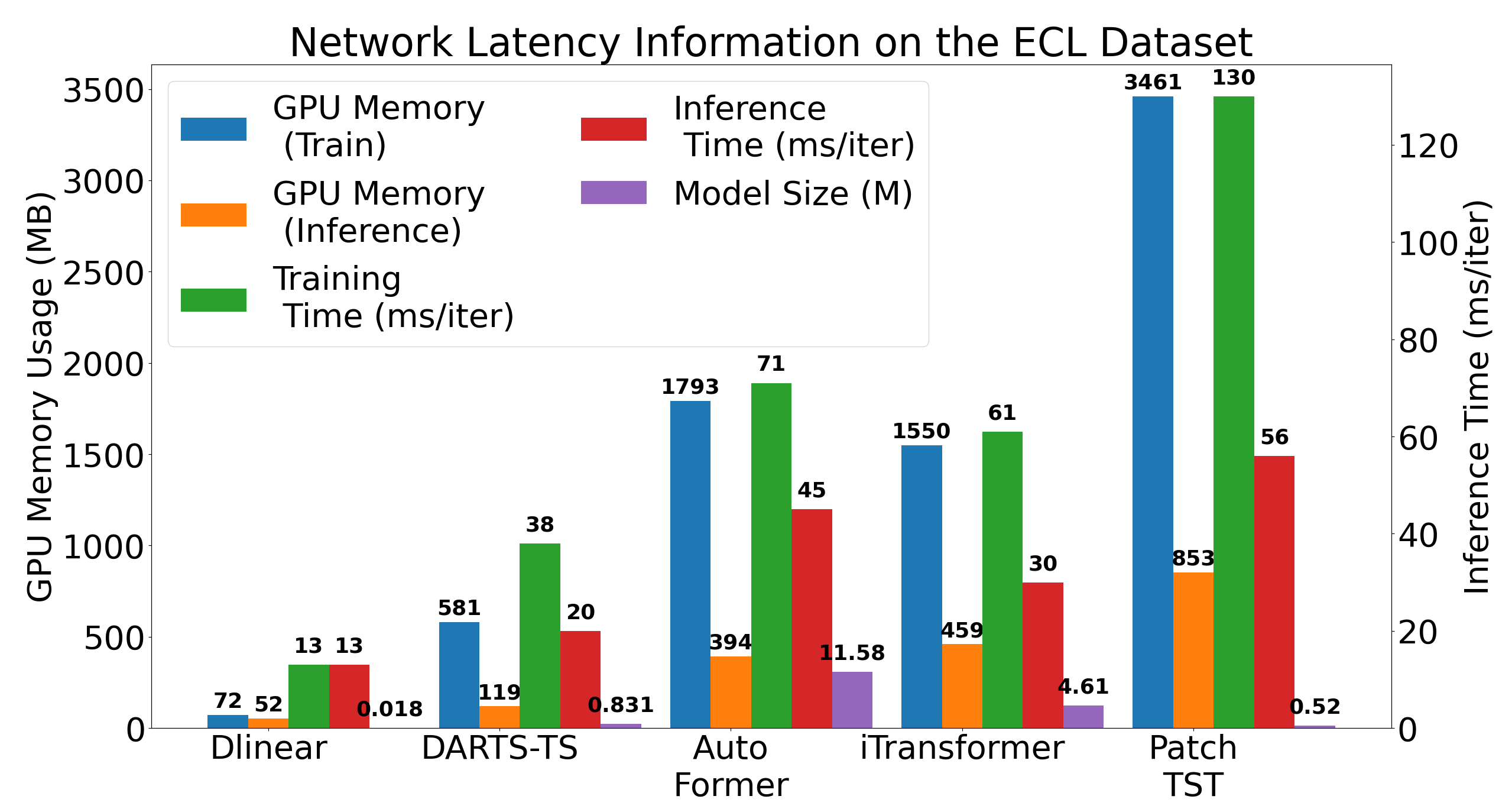}
    \caption{Latency information of different networks on the Traffic (left) and ECL (right) dataset. We set the look-back window size and the forecast horizon to 96. The batch size for both training and inference is set as 32.}
    \label{fig:latency_traffic}
\end{figure}

As shown in Figure~\ref{fig:latency_traffic}, while having a comparable performance with iTransformer and PatchTST on the traffic dataset, our approach requires around 2x less GPU memory and is faster than the iTransformer and 3x less GPU memory and speed up compared to PatchTST during the training phases. During the test phases, the required GPU memory is even further reduced to around 300 MB, which is 4x less than the iTransformer and 6x less than the PatchTST, and only requires \mbox{3x} more memory compared to a linear layer. A similar trend can be observed on the ECL dataset: \Ourname{} requires 7x less GPU memory and is 2.8x faster than PatchTST to achieve a better performance. 

\textupdate{Although we do not optimize for the latency as our optimization objects explicitly, the combination of different operations already allows us to achieve a similar performance with much less computational power required. e.g., to cover the same receptive field, we might need to stack multiple TCN layers or increase their kernel size. Both approaches result in an increased amount of computational time and memory consumption. However, if we instead apply a transformer or RNN layer on top of a TCN layer, the model could still learn the global and local information with less computational power required. Additionally, models such as PatchTST~\citep{nie-iclr23a} decompose one single multi-variant series instance into multiple uni-variant series instances and therefore, need to run the forward pass multiple times for a single multi-variant time series instance. ~\Ourname{} only performs one forward pass to get the forecasting results, which could significantly reduce the required computational power and the memory requirements. Hence, ~\Ourname{} could search for a more lightweight architecture, even if latency information is not the optimization target. However, we could also apply other optimization approaches, such as multi-objective differentiable architecture search~\citep{sukthanker-arxiv24a}, to search for the architecture that concerns both latency and accuracy.}

\begin{figure}[ht]
    \centering
    \includegraphics[width=0.4\textwidth]{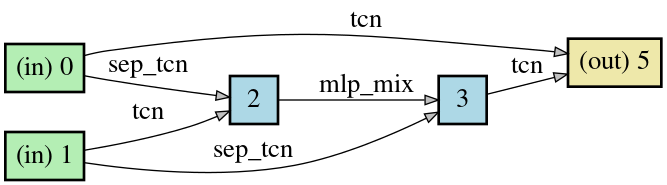}
    \centering
    \includegraphics[width=0.4\textwidth]{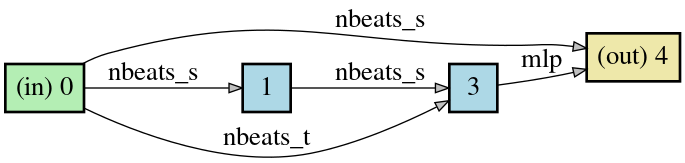}
    \caption{One optimal architecture on the ECL dataset, this is an encoder-only architecture and will be trained with an MSE loss}
    \label{fig:arch_ecl}
\end{figure}

\begin{figure}[ht]
    \centering
    \includegraphics[width=0.4\textwidth]{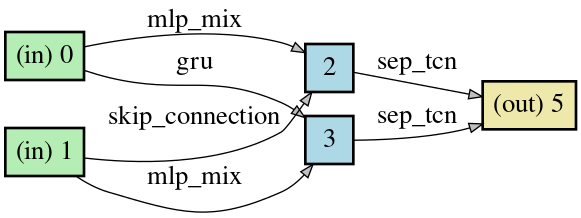}
    \centering
    \includegraphics[width=0.4\textwidth]{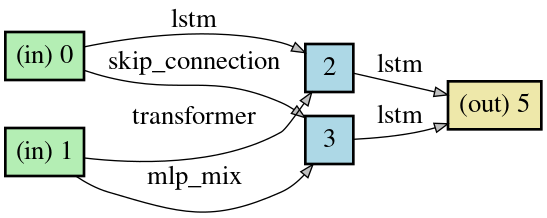}
    \centering
    \includegraphics[width=0.7\textwidth]{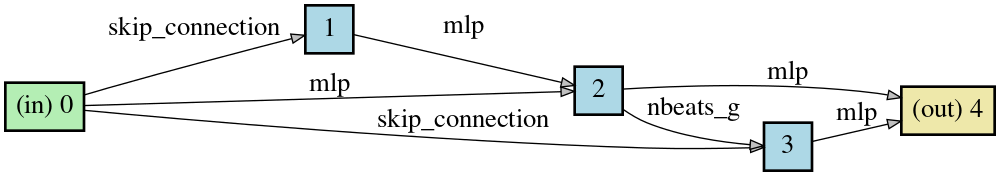}
    \caption{One optimal architecture on the ETTm2 dataset, this is an encoder-decoder architecture and will be trained with a quantile loss}
    \label{fig:arch_ettm2_1}
\end{figure}

\subsection{The optimal architectures~\label{sec:opt_arch}}

We show one of the optimal architectures found on the ECL dataset in Figure~\ref{fig:arch_ecl}. Its $\SeqNet$ is an encoder-only architecture that is composed of MLPMixer, TCN, and Separate TCN modules. 
TCN modules are still preferable over the other components, showing that the ECL dataset might prefer a model that focuses on the local correlation. While the $\FlatNet$ contains lots of N-BEATS-Seasonal modules, this indicates the strong seasonal and little trend signal that the dataset contains.

We present another encoder-decoder architecture in Figure~\ref{fig:arch_ettm2_1}. This architecture is optimized on the ETTm2 dataset. It applies two MLP-Mixer layers to the input node to first collect the global information from the raw input sequence and then apply two Separate TCN modules on top of that. Our optimizer also selects many LSTM modules for the decoder architectures, even if the corresponding encoder edges do not provide any hidden states. Additionally, no TCN module in the decoder layer. This is different from the optimal encoder architecture, where lots of TCN family components are selected. This indicates that the decoder networks would require modules that provide a global perspective to utilize all the information from the encoder networks, and therefore, the priority of TCN modules might decrease. More optimized architectures can be found in the appendix~\ref{sec:other_opt_archs}.

\section{Discussion and Future Work~\label{sec:conclusion}}
This work proposes a general search space for time series forecasting tasks. Our search space allows the components in the search space to freely connect to each other and form a new network. This search space contains most of the forecasting architectures and can be easily extended to other frameworks. For instance, iTransformer~\citep{liu-iclr24c} can be considered as a special case of the $\FlatNet$ and searched jointly with the other modules from this family. Decomposing multi-variant series into single variant series and applying a special kernel result in PatchTST~\citep{nie-iclr23a}, and then we can search for the optimal architecture jointly with the other $\SeqNet$. 

In Section~\ref{sec:exp}, we showed that \Ourname{} can search for a lightweight architecture while keeping strong performance on various datasets. However, unlike traditional supervised problems where all the sample instances are i.i.d., time series data might have the problem of distribution shift. The optimal model searched on the validation set might no longer work well on the test set. This provides a future challenge for the AutoML forecasting frameworks from the meta-level, i.e., they need to be able to pre-determine the optimal approach to evaluate the generalization ability. 

\input{appendix/acknowledgement}





\bibliography{bibtex/strings, bibtex/lib, bibtex/bib_local, bibtex/proc_local, bibtex/proc}
\bibliographystyle{plainnat}


\newpage
\appendix

\section{Operations Deatails~\label{sec:nn_components}}
 In section ~\ref{sec:ssd_ol}, we briefly introduced the operations within our search space. Here, we will provide the details of these operations. 
\subsection{\SeqNet}
For $\Seq$ net, encoders and decoders share the same operation sets. Overall, we have the following operations:
\begin{itemize}
    \item TSMixer~\citep{chen-tmlr23a}, a full MLP-based Sequential operation. Each TSMixer operation is constructed by the time and feature mixing blocks. The time mixing blocks use a fully connected (FC) layer to mix the information across different time steps. In contrast, the feature mixing block uses another set of FC layers to enhance the information within each channel. Our TSMixer encoders follow the design from ~\citet{chen-tmlr23a}, and the size of the feature mixing layer is set as $2\times d_{model}$. For TSMixer decoders,  we first construct the input feature map with the encoder network outputs. This concatenated feature is then provided to the time mixing modules to recover its size and return it to the forecasting horizon. Additionally, we use LayerNorm~\citep{ba-arxiv16a} instead of BatchNorm~\citep{ioffe-icml15a} in TSMixer to ensure that the operations within an edge generate the feature maps that follow the same distribution (since all other components in our module used LayerNorm to normalize the feature maps). 
    \item LSTM~\citep{hochreiter-nc97a} is an RNN model. It maintains a set of cell gate states to control the amount of information passed to the next time steps. Therefore, it suffers less from the known gradient explosion problems in the RNN families. However, the introduction of the gates brings lots of additional parameters to the modules. As described in Section~\ref{sec:micro_level}, the hidden states of the LSTM decoder are initialized by the last time step of the encoder feature from the corresponding layer and another feature generated with a linear embedding (or directly from the corresponding LSTM encoder).
    \item GRU~\citep{chung-arxiv15a} is yet another type of RNN network. It only maintains one hidden state and therefore requires a much smaller number of parameters and computations compared to the LSTM. The setting of the GRU Encoder/Decoder is nearly the same as the LSTM families. The only difference is that we do not maintain an additional state to initialize the GRU decoders.
    \item Transformer~\citep{Vaswani-neurips17a} has attracted lots of attention from different research fields and is therefore widely applied in time series forecasting tasks. Here, we use the vanilla Transformer implemented in PyTorch~\citep{pytorch-neurips19a} with the hidden size to be $4\times d_{model}$. 
    \item TCN~\citep{bai-arxiv18a} is a type of CNN network that can capture the local correlations among different time steps. However, the receptive field of the TCN network is restricted by its kernel size. To efficiently increase the receptive field without introducing too much computation overhead with a larger kernel, TCN implemented dilated convolution operations. Here we implement a similar approach,  for edge $i \leftarrow j$ that starts from node $j$ to node $i$ as cell $k$, we set its dilation as $2^{(j + k - n_{in})}$, where $n_{in}$ is the number of inputs of the current cell. Hence, the deeper convolutional layers will have a larger receptive field. For the TCN decoders, we concatenate the feature maps from the corresponding encoder layer with our input feature and feed them together to our TCN network. This idea is similar to U-Net~\citep{ronneberger-miccai15a}, where features with similar levels should be gathered together. 
    \item SepTCN~\citep{luo-iclr24a} is a variation of the vanilla TCN. We replace the full convolutional operations in TCN with a combination of a separated TCN model and another $1 \times 1$ linear layer.
    \item Skip Connection, an identity layer that passes its input to the next level. However, for the skip connection encoder, we still have a linear layer to provide initial cell gate states to the corresponding LSTM decoder layers.
\end{itemize}

Another type of $\Seq$ decoder is a linear decoder. We apply a linear layer that transforms the encoder's output feature map with size $R^{[B, L, N]}$ to $R^{[B, H, N]}$ and feed it further to the forecasting heads. 

 \subsection{\FlatNet}
 We only consider the MLP families in our $\FlatNet$ to minimize the computational overhead when applying our approaches to problems with higher series amounts. Given an input feature series with shape $[B, N, L]$, we first concatenate it with a zero tensor with shape $[B, N, H]$ that represents the prediction results. Then this concatenated tensor is fed to the $\Flat$ encoder.
 
 This architecture family includes:
 \begin{itemize}
     \item a simple Linear model~\citep{zeng-aaai23a}. This linear layer maps its input features with shape $[B, N, L + H]$ to a feature map whose size is equal to the forecasting horizon: $[B, N, H]$. Then the output feature is concatenated with the first part of the input feature maps. If the network only contains skip connections for all but the last layer, then this network becomes a DLinear model~\citep{zeng-aaai23a}. If the operation is not the output layer, we attach an activation and normalization layer to introduce some non-linearity.
     \item N-BEATS~\citep{oreshkin-iclr20a} modules. N-BEATS is a hierarchical module where each model is composed of multiple stacks. Each stack contains multiple blocks. Each block has an FC stack with multiple FC layers, a forecasting head, and a backcasting head. In our search space, each N-BEATS edge corresponds to an N-BEATS block. N-BEATS provides three variations: generic, trend, and seasonal. We include them all in our search space. Since the only difference between these variations is their prediction heads. We ask the models to share the same FC layer backbones and only diverge at the forecasting heads. 
     \item Skip Connection, a skip connection layer.  
     
 \end{itemize}

\subsection{Forecasting Heads}
We also consider the forecasting heads as part of the operations within our graph. Each of these heads 
is composed of one or multiple linear layers that map the $\SeqNet$~\footnote{For $\FlatNet$, there is no need to have an additional head if the loss only requires one output} output feature maps to the desired multiple-variable target values. These linear layers are then trained with a set of specific training losses. 
Let's assume that the target value is $\target$ and prediction value is $\prediction$
\begin{itemize}
    \item Quantile loss~\citep{lim-ijf21a, wen-tsw17a} predicts the percentiles of the target values. A quantile head can be composed of multiple heads, and each of the heads is asked to predict a $q$ quantile. Given a required quantile value $q$, the quantile loss is computed by $\mathcal{L}_q = \max (q(\target - \prediction) + (1-q)(\prediction - \target))$. In our network, we used the following quantile values: $\{0.1, 0.5, 0.9\}$. The final prediction is given by the $0.5$ quantile values. A quantile head is then a set of linear layers whose size is the number of quantile values
    
    \item MSE loss is yet another popular choice in time series forecasting tasks. It is computed by $\mathcal{L}_{MSE}=(\prediction - \target)^2$. An MSE head is a single linear layer whose weights are updated with MSE loss.
    
    \item MAE loss is similar to MSE loss. However, instead of computing the mean square error from MSE, it computes the mean absolute error: $\mathcal{L}_{MAE}=|\prediction - \target|$. Similar to the MSE layer, an MAE head is also composed of one linear layer, but its weights are updated with MAE losses. 
\end{itemize}

We stack these forecasting heads on top of the $\SeqNet$ decoders and optimize their architecture weights with validation losses. 

\section{Experiment Details~\label{sec:exp_detail}}
We show detailed information on the dataset that we applied in Table~\ref{tab:datainfo}. Given the great discrepancy in variable size between different datasets, we divide the datasets into two groups: we assign a smaller model to the datasets with fewer variables, such as Weather, Exchange, and ETTs. We then attach a normalization layer within the linear decoder. For the other datasets, we design an architecture search space with a relatively large model. Additionally, we removed the normalization layer in the linear decoder since the number of target variables is larger than our model size, and the final forecasting head does not need to recover the distribution of the target variable from the normalized feature maps. 

We run our experiments on a Cluster equipped with Nvidia A100 40 GB GPUs and AMD Milan 7763 CPUs. For each dataset, we perform the search over the smallest forecasting horizons and evaluate the same optimal model on all the other forecasting horizons. 
Each task is repeated 5 times. 
The resources spent on each evaluation depend on the size of the dataset. The exact GPU hours spent for each task and stage are presented in Table~\ref{tab:time_used}. The search is divided into three stages. The first stage is to jointly optimize the network weights and parameters, and the second stage is to select the optimal operations within each edge. Finally, the third stage is applied to only preserve two edges toward each node. Overall, it takes roughly 4 hours to evaluate smaller datasets such as ETThs and Exchange. Other tasks might require up to 10 GPU hours. The ablation study requires another 500 GPU hours. Overall, it takes roughly 1200 GPU hours to finish the experiments. 

\begin{table}[h]
    \centering
    \scalebox{0.95}{
    \input{tables/search_time/search_time}}
    \caption{GPU hours used for the search stage}
    \label{tab:time_used}
\end{table}

For all the baselines,  we use their official implementation from PatchTST\footnote{\url{https://github.com/yuqinie98/PatchTST}}, ModernTCN\footnote{\url{https://github.com/luodhhh/ModernTCN}}, and  TSMixer\footnote{\url{https://github.com/google-research/google-research/tree/master/tsmixer}}, while for the other baselines, we take the implementation from Time-Series-Library\footnote{\url{https://github.com/thuml/Time-Series-Library}}.

\begin{table}[h]
    \centering
    \begin{tabular}{c|c|c}
        \toprule
         Dataset &  $\#$ \ Time\ steps & $\#$ \  Variables\\
         \midrule
         ECL & 26304 & 321 \\
         Traffic & 17544 & 862 \\
         Weather & 62696 & 21 \\
         Exchange & 7588 & 9 \\
         ETTh1 & 17420 & 7\\
         ETTh2 & 17420 & 7 \\
        \bottomrule
    \end{tabular}
    \begin{tabular}{c|c|c}
        \toprule
         Dataset &  $\#$ \ Time\ steps & $\#$ \  Variables\\
         \midrule
         ETTm1 & 69680 & 7 \\
         ETTm2 & 69680 & 7\\
         PEMS03 & 26208 & 358 \\
         PEMS04 & 16992 & 307 \\
         PEMS07 & 28224 & 883 \\
         PEMS08 & 17856 & 170 \\
        \bottomrule
    \end{tabular}
    \caption{Data set information}
    \label{tab:datainfo}
\end{table}

Our architecture search framework is a two-stage approach. In the first stage, we search for the optimal architecture, while in the second stage, we 
train the proposed network from scratch. We preserve most of the searching hyperparameters from ~\citet{liu-iclr18a}. However, we apply ADAM instead of SGD during the test phases to optimize the network weights. All the optimizers are applied with CosineAnnealingWarmRestarts~\citep{loshchilov-iclr17a} that restarts learning rates every 20 epochs. We also apply a smaller learning rate and stronger weight decay to the smaller datasets to avoid overfitting. The concrete hyperparameter settings during search and evaluation phases are presented in Table~\ref{tab:hyperp}

\begin{table}[h]
    \centering
     \scalebox{0.8}{
    \begin{tabular}{c|c|c|c|c}
        \toprule
         HP Names & \multicolumn{2}{c}{Search HP Values} & \multicolumn{2}{c}{Evaluation HP Values} \\ & Weights Optimizers & Architecture Optimizers &  big datasets & small datasets \\ 
         \midrule 
         Epochs & 40 & 40 & 100  & 100 \\
         Gradient Clip & 0.1& 0.1 & 0.1 & 0.1 \\
        \midrule
         Type & SGD &  Adam & Adam  & AdamW \\
         Learning Rate & 0.025 & 0.001 & 0.001 & 0.0002 \\
         Weight Decay & 0 & 0.001 & 0 & 0.1 \\
         Momentum & 0.9 & & & \\
         Betas &  & (0.5, 0.999) & (0.9, 0.999) & (0.9, 0.999) \\
        \bottomrule
    \end{tabular}}
    \caption{Training Hyparameters (HP) during searching and training phases}
    \label{tab:hyperp}
\end{table}

All the look-back window sizes are set to 336 (for long-term forecasting tasks) and 96 (for PEMS tasks) for both the searching and testing phases. We divide the datasets into two groups based on their number of variables: \begin{enumerate*}
    \item The PEMSs, traffic, and ECL dataset belongs to the big dataset
    \item The remaining datasets, including Weather and ETTs, are small datasets.
\end{enumerate*}

Both datasets share nearly the same architecture, which is a mixed network introduced in Section~\ref{sec:macro_level}. The number of $\Seq$ and $\Flat$ cells is both set to $2$, and they could receive $2$ and $1$ input variables, respectively. However, for the big dataset, we set the number of $\SeqNet$  hidden dimensions as $32$, while this value for the small dataset is $8$. This approach is also applied to the hidden NBATS dimension of $\FlatNet$: we set this value as $256$ for the big datasets and $96$ for the small datasets. The hyperparameters applied during training and validation are listed in Table~\ref{tab:hyperp}

Since the $\Seq$ encoder has two input nodes, instead of feeding the raw input value to both input nodes, we decompose the input nodes into trend-cyclical components by using an average moving~\citep{wu-neurips21a, zeng-aaai23a} to ask the edges in the cell to focus on different levels of information.

\begin{table}[h]
    \centering
    \input{tables/flops_changes}
    \caption{GFLOPS changes after different stages}
    \label{tab:flops_changes}
\end{table}

We now study how FLOPS change within each Prune stage. ~\Ourname{} search process involves three stages: weights optimization, operation pruning, and topology pruning. We show how the flops change after each stage in Table~\ref {tab:flops_changes}. Most of the FLOPS are substantially reduced after the operation pruning stage, where only one operation is preserved for each edge. This value is further reduced with the Topology Prune stage, where each node only preserves at most two edges as inputs.

\section{Results on all the forecasting horizons~\label{sec:full_res}}
\begin{table}[t]
    \centering
    \scalebox{0.4}{\input{tables/full_res}}
    \caption{The full evaluation results on long-term forecasting datasets. We evaluate each model five times and take the mean of the final results. The optimal models are marked in red, and we underline the models that are not significantly worse than the optimal.}
    \label{tab:res_all}
        
\end{table}

\begin{table}[t]
    \centering
    \scalebox{0.4}{\input{tables/full_res_pems}}
    \caption{The full evaluation results on PEMS datasets. The optimal models are marked in red, and we underline the models that are not significantly worse than the optimal.}
    \label{tab:res_all_pems}
        
\end{table}


Table ~\ref{tab:res_all} and Table ~\ref{tab:res_all_pems} show the results w.r.t. each forecasting horizon. While \Ourname{} requires a much smaller amount of resources and parameters, it still shows comparable performances on many datasets.


\subsection{Statstical Test}
\begin{table}[h]
    \centering
     \scalebox{0.6}{\input{tables/statistical_test_LTSF}}
    \caption{Statistical Test against the non-NAS architectures on the LTSF datasets}
    \label{tab:stastic_test_ltsf}
\end{table}

\begin{table}[h]
    \centering
     \scalebox{0.6}{\input{tables/statisitical_test_pems}}
    \caption{Statistical Test against the non-NAS architectures on the PEMS datasets}
    \label{tab:stastic_test_pems}
\end{table}

\textupdate{
Table~\ref {tab:stastic_test_ltsf} and Table~\ref{tab:stastic_test_pems} show the statistical test against the best non-NAS baseline for each dataset. For the LTSF datasets, the optimal baseline models differed across different datasets. With a significant level of $0.05$, ~\Ourname{} is significantly better than the baseline models on most of the benchmarks for ECL, ETTm1, and ETTm2 datasets. While for the other benchmarks, ~\Ourname{} is many times comparable to the best baselines, with the forecasting losses slightly higher or lower than the baselines. Additionally, the optimal baselines differ across different datasets, which also shows that the architectures need to adapt to different tasks, further highlighting the importance of heterogeneous forecasting architectures.  }

\textupdate{While for the PEMS datasets (Table~\ref{tab:stastic_test_pems}), the optimal baselines concentrates mostly on the iTransformer~\citep{liu-iclr24c} and TimesNet~\citep{wu-iclr23a}. However, ~\Ourname{} is still significantly better than the baseline models on the PEMS04 and PEMS07 datasets. }

\section{Comparing with zero-shot forecasting foundation models}
Time series forecasting foundation models~\citep{ ansari-arxiv24a, bian-icml24a, das-arxiv23a, liu-arxiv24a} have become the trend for recent forecasting tasks. Instead of training multiple networks for different tasks separately, forecasting foundation models train one single model across multiple datasets and directly test it on the target sequence without further training the model on that dataset.

Here, we compare ~\Ourname{} with another two forecasting foundation models: TimesFM~\citep{das-arxiv23a} (with 200M parameters) and Moirai-MOE~\citep{liu-arxiv24a} (moirai-moe-1.0-R-small with 117M parameters and 11M active parameters). Both approaches consider the input series independently and patchify each single variant time series by integrating neighboring data points into one single patch~\citep{nie-iclr23a}. Since no further training process is required for these two models, and the evaluation costs can be expensive, we only evaluate these two models with one random seed.

The result is shown in Table~\ref{tab:res_llm_lsft} and Table~\ref{tab:res_llm_pems}. We run all the foundation models on Nvidia H100 GPUs with 80GB GPU memory for at most 24 hours. Any run that fails to finish the prediction (either by memory out or out of time) will be marked with $NaN$. \Ourname{} achieves the optimal performance on the ETT datasets and is only slightly worse than TimesFM on the ECL datasets with forecasting horizons of $96$ and $192$. TimesFM achieves a better performance on the traffic, weather, and ECL datasets. However, these three datasets are also part of the training set from TimesFM, and therefore, the forecasting results on these datasets can no longer be considered as a zero-shot forecasting setup. While for the other dataset, where TimesFM is not trained, including the four PEMS datasets,  \Ourname{} still outperforms TimesFM's zero-shot forecasting results. 

\begin{table}[h]
    \centering
    \scalebox{0.6}{\input{tables/llm_based/res_with_llm}}
    \caption{Results against forecasting foundation models in Long Term Forecasting datasets}
    \label{tab:res_llm_lsft}
\end{table}

\begin{table}[h]
    \centering
    \scalebox{0.6}{\input{tables/llm_based/res_with_llm_pems}}
    \caption{Results against forecasting foundation models in PEMS dataset}
    \label{tab:res_llm_pems}
\end{table}

\begin{figure}[t]
    \centering
    \includegraphics[width=0.485\textwidth]{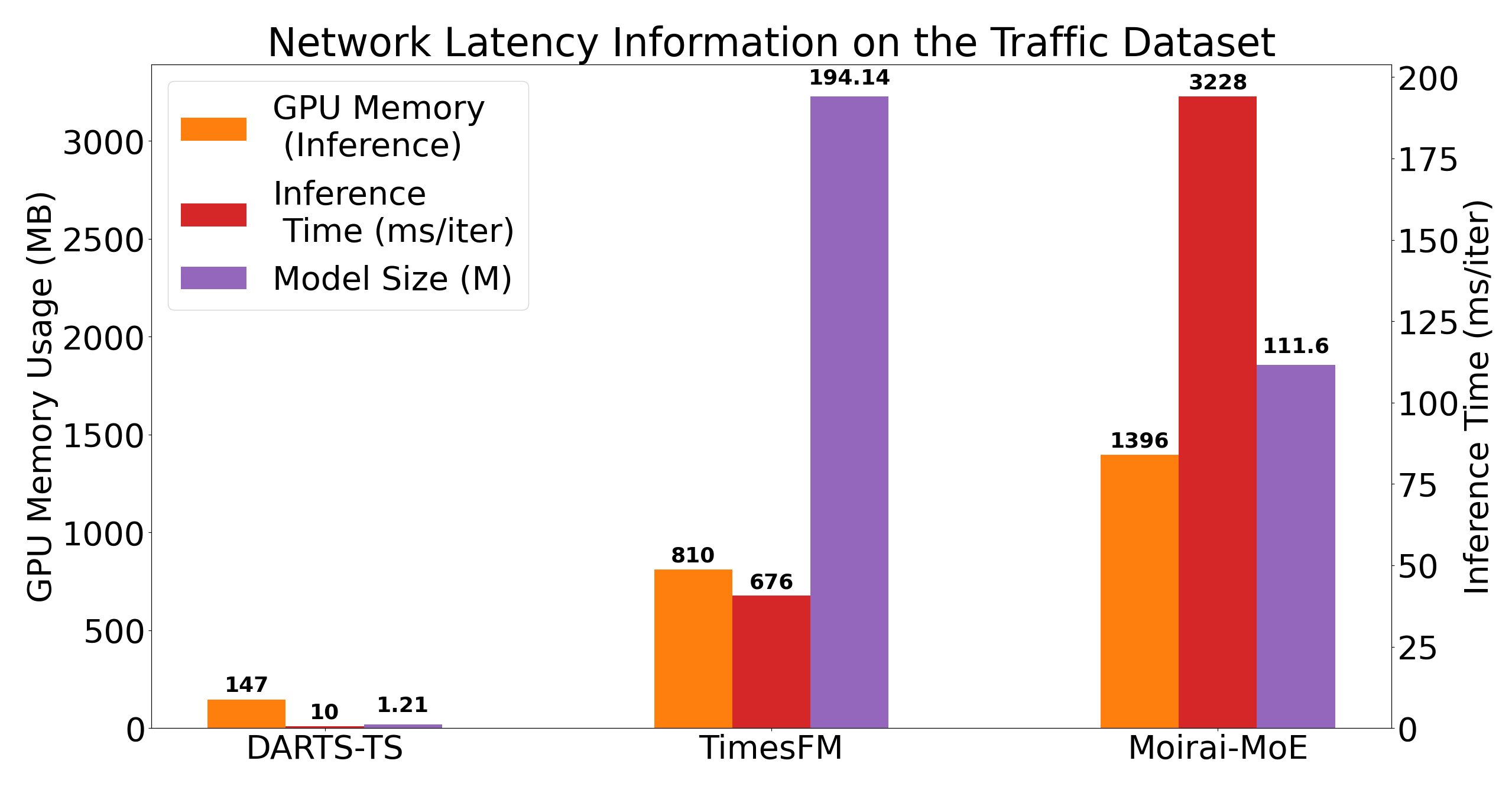}
    \includegraphics[width=0.485\textwidth]{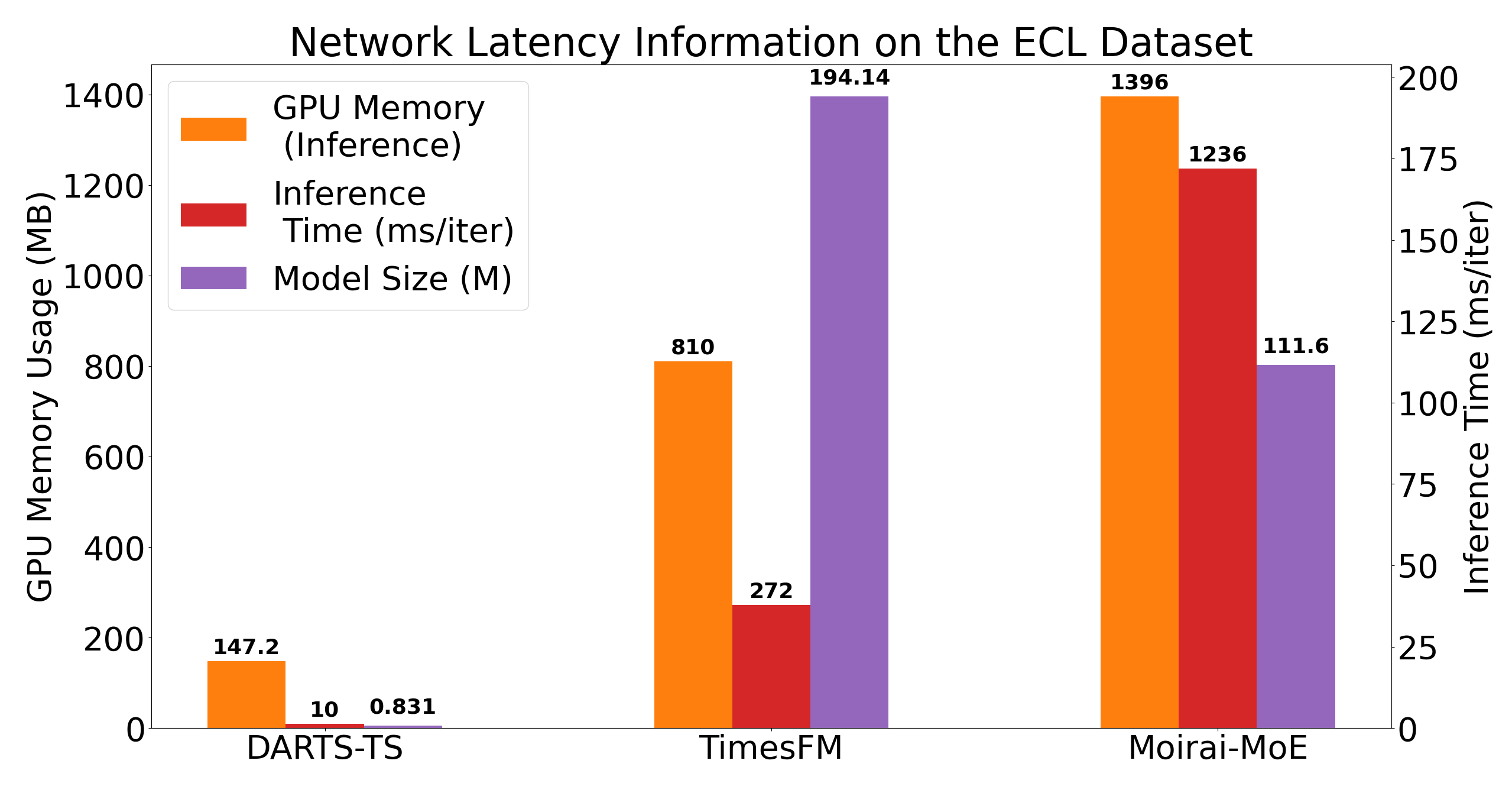}
    \caption{Latency information of different networks on the Traffic (left) and ECL (right) dataset. We set the look-back window size and the forecast horizon as 96 and the batch size as 1.}
    \label{fig:latency_llms}
\end{figure}

We also provide a comparison of the latency cost during inference time between \Ourname{}, TimesFM, and Moirai-MOE. Due to the memory limitation,  we only evaluate with a batch size of 1 and set the number of samples for Moirai-MoE to 10 instead of 100 random samples in its original setup. The result is shown in Figure~\ref{fig:latency_llms}. ~\Ourname{} provides a 67 times speed up on Traffic and a 27 times speed up on ECL compared to TimesFM. This speed-up increases to 322 times on Traffic and 123 times on ECL compared to Moirai-MoE. Showing the efficiency of ~\Ourname{} compared to the LLM-based approaches. 

\section{Ablation Study}

\subsection{The impact of window size~\label{sec:ablation_ws}}
In our experiments, we fixed our window size to 336 and applied this window size to predict different forecasting horizons. However, since the look-back window size is an important hyperparameter in forecasting tasks, it is interesting to see if the model should always stick to the window where it is trained. To answer this, we ask the optimizer to search for an architecture that requires the input window size $\{96, 192, 336, 720\}$. We then evaluate each of these found architectures with the window sizes mentioned above.  

\begin{figure}
    \centering
    \includegraphics[width=0.7\textwidth]{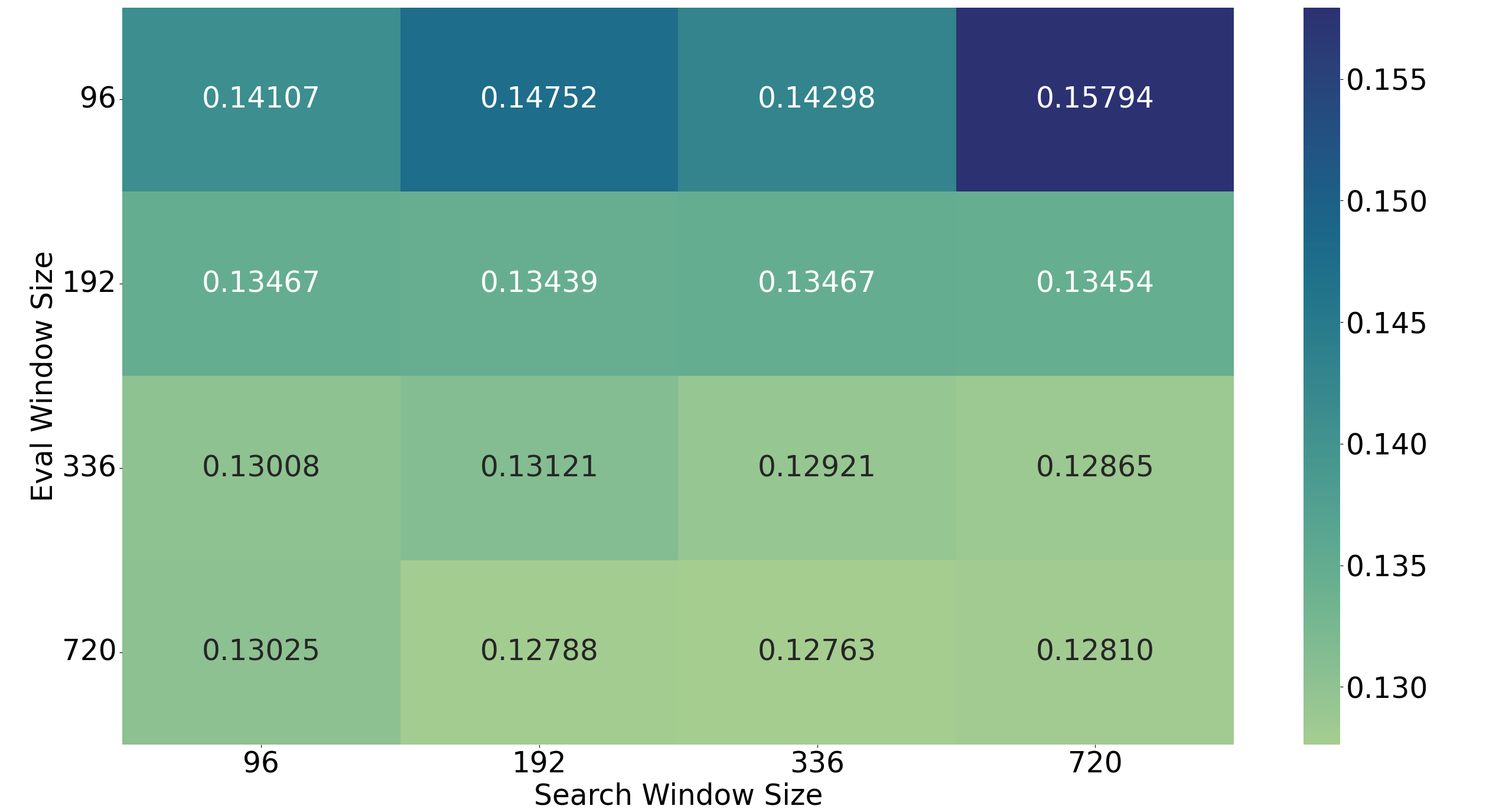}
    \caption{The impact of window size during searching and evaluations on the ECL-96 task}
    \label{fig:abla_winsize}
\end{figure}

We run this task on the ECL-96 dataset. The result is shown in Figure~\ref{fig:abla_winsize}. We see that our model will perform better with the increase in search window size in general. While the search window size also influences the final evaluations, a model searched with a window size of 720  performs the worst when the architecture suggested by this optimizer is asked to make a prediction with a window size of 96 and vice versa. However, this gap becomes smaller if the search window size is closer to the eval window size. This indicates that the window size we used to search for the network should not be too far away from the actual window size used to evaluate the model.  

\subsection{Forecasting Horizon}
\begin{figure}
    \centering
    \includegraphics[width=0.7\textwidth]{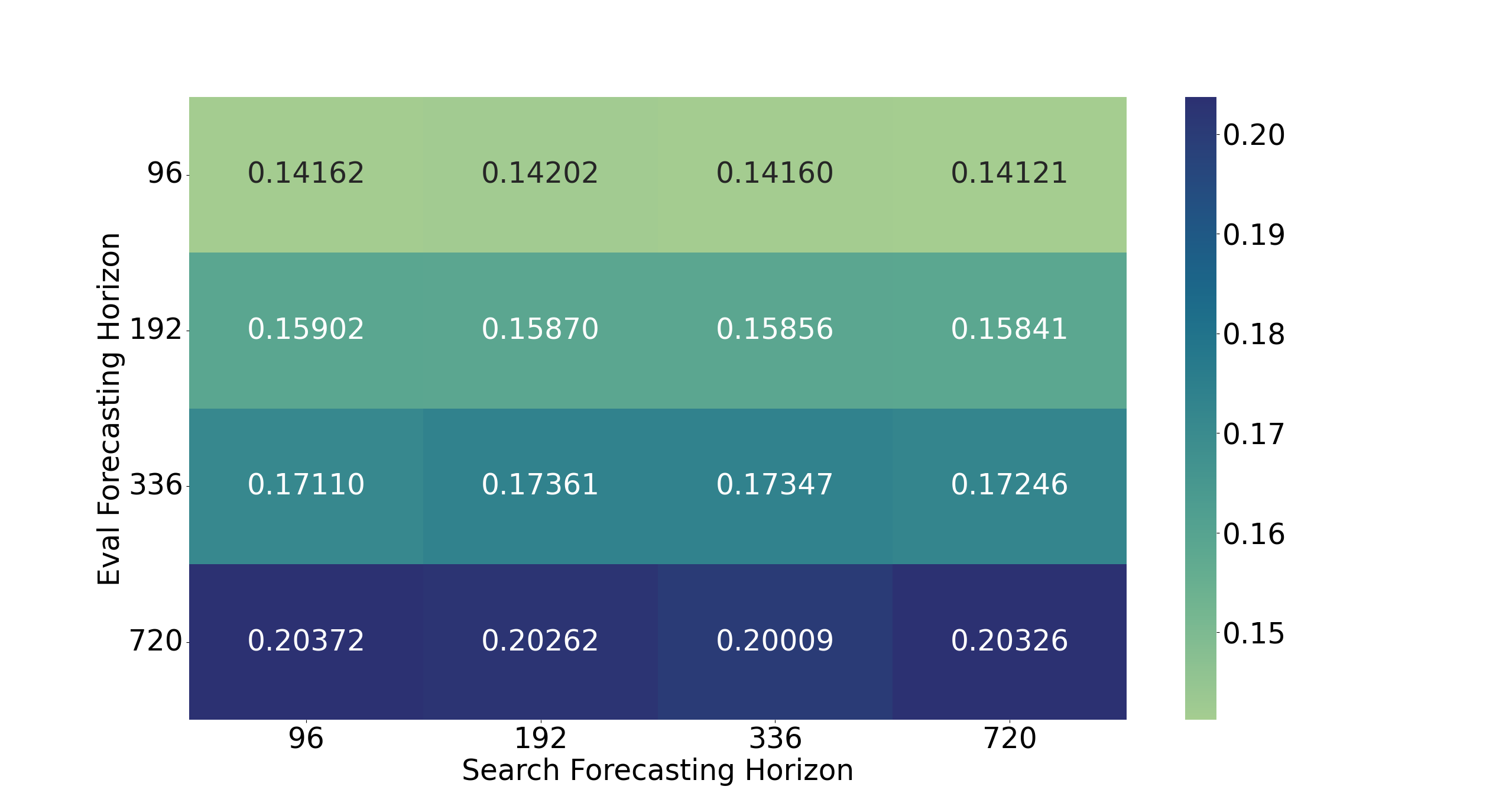}
    \caption{The impact of forecasting horizon during searching and evaluations on the ECL-96 task}
    \label{fig:abla_horizon}
\end{figure}

In our experiments, we only searched with the forecasting horizon of 96 and applied them to the other forecasting horizon tasks. Here, we check how the optimal architecture searched for one target forecasting horizon can be transferred to another forecasting component. Similar to Section~\ref{sec:ablation_ws}, we ask the optimizer to search for the models with forecasting horizon of size $\{96, 192, 336, 720\}$. We then evaluate each of these optimal architectures with the other forecasting horizons mentioned above. The sliding window size is set $96$ for all tasks in this scenario. The result is shown in Figure~\ref {fig:abla_horizon}. The search forecasting horizons could also be transferred to the other forecasting horizons during evaluation, as different searching forecasting horizons provide similar performance under the same evaluation forecasting horizon.

\subsection{Forecasting components}
We constructed a hierarchical search space in Section~\ref{sec:searchspacedesign}. Here, we will show the efficiency of each component. We provide the following variations:
\begin{itemize}
\item \Ourname{} CV. This variation has the same architecture as ~\Ourname{}. However, the training validation dataset splits strategy during its search phase is 5-fold cross-validation instead of the holdout strategy described in Section~\ref{sec:search_strategy}
\item Flat Only. This variation only contains the $\FlatNet$ in the search space. 
\item Seq Only. This variation only contains the $\SeqNet$ in the search space. 
\item Parallel. This variation is similar to our approach. However, the $\SeqNet$ receives only feature variables instead of the output of the $\FlatNet$
\item Concat Seq First. This variation first feeds the input data to the $\SeqNet$ and then the output of $\SeqNet$ is fed to the $\FlatNet$
\item No Weights. This variation removed the weighted sum approach described in Section ~\ref{sec:macro_level}
\end{itemize}

\begin{table}[t]
    \centering
    \scalebox{0.45}{\input{tables/res_ablation_components}}
    \caption{Ablation over the components in \Ourname}
    \label{tab:res_ablation_components}
        
\end{table}

The result is shown in Table~\ref {tab:res_ablation_components}. Despite that \Ourname{} CV achieves nearly the same performance on some datasets, such as ECL, ETTm2, Exchange, and Weather, there is still a performance gap between ~\Ourname{} Holdout (i.e., ~\Ourname{} in the table) and ~\Ourname{} CV on the other datasets. The Parallel approach is slightly worse than \Ourname{} on many datasets. However, there is a huge gap between Parallel and \Ourname{} on the Traffic dataset. While the Flat Only approach is generally worse than \Ourname{} on the ETT datasets. Overall, we show that the architectural design of \Ourname{} generally provides us with architectures that are robust across many datasets.

\subsection{Optimizazion Epochs}
In Table~\ref {tab:hyperp}, we optimize the supernet for 40 epochs and then prune the network with the pre-trained architectures. Here, we study the impact of the number of optimization epochs.

\begin{figure}
    \centering
    \includegraphics[width=0.875\linewidth]{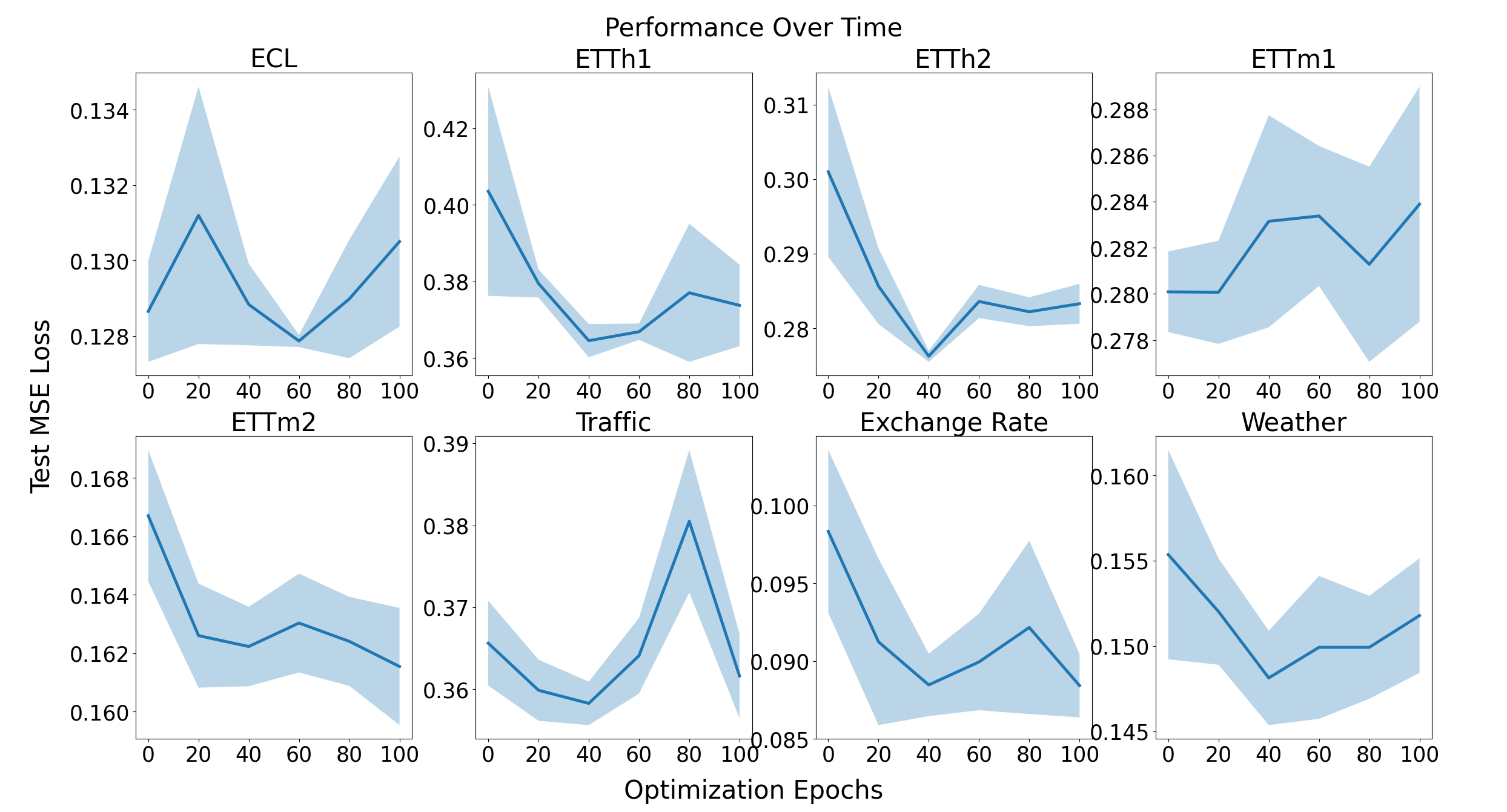}
    \caption{Performance Over Number of Supernet Optimization Epochs}
    \label{fig:ablation_performance_overtime}
\end{figure}

The result is shown in Figure~\ref{fig:ablation_performance_overtime}. Each model is optimized for $\{0, 20, 40, 60, 80, 100\}$ epochs, respectively. Since we will also update the architecture weights and parameters during the pruning stage, even optimizing the models for zero epochs might also return a well-performing model. However, training the model for too long might overfit the validation set~\citep{zela-iclr20a}. Overall, the optimal number of epochs might differ across different datasets, while optimizing the super net for $40$ or $60$ epochs might result in a better average performance.

\textupdate{\subsection{Operations}
As described in Section~\ref{sec:nn_components}, our search space contains many different operations. To verify the importance of each operation, here we remove each operation from the search space and check its performance on each of the target datasets.}

\begin{table}[h]
    \centering
    \scalebox{0.55}{\input{tables/res_ablation_ops}}
    \caption{Ablation over the Operations in our search space}
    \label{tab:res_ablation_ops}
        
\end{table}

\textupdate{The result is shown in Table~\ref {tab:res_ablation_ops}. Despite that, removing certain operations from the search space might provide better results on certain tasks. For instance, removing Separated TCN on ECL datasets results in a model with $0.128$ MSE loss instead of the $0.129$ MSE loss found by the Full search space. However, removing the same operation on the ETTh1 dataset increases the loss from $0.365$ to $0.381$. This degeneration also appears for other operations, for instance, removing MLP increases the loss on the Traffic dataset from $0.358$ to $0.379$, while removing N-BEATS-Trend could further increase this value to $0.402$. Hence, the search space could benefit from all the operations described in Section~\ref{sec:nn_components} depending on the datasets to which the model is applied.  This further highlights the importance of exploring various architectures for different target tasks. }

\textupdate{\subsection{RevIN}
In Section~\ref {sec:search_strategy}, we applied RevIN~\citep{kim-iclr22a} in ~\Ourname{} during both the searching and testing phases; here, we present an ablation study on the RevIN for ~\Ourname{}. }

\begin{table}[h]
    \centering
    \scalebox{0.65}{\input{tables/res_ablation_revin}}
    \caption{Ablation for the RevIN in our search space}
    \label{tab:res_ablation_revin}
        
\end{table}

~\textupdate{Table~\ref{tab:res_ablation_revin} shows the impact of RevIN. Except for the Weather dataset, where the mean MSE loss increases from 0.147 to 0.148 by applying RevIN, the performance of all the other datasets can be improved with RevIN, especially on the ETTs and traffic datasets. Overall, RevIN enables more accurate forecasting models within our search space. }

\subsection{Separated Encoder-Decoder Search Space}
In Section~\ref{sec:micro_level}, we design a search space that allows the encoder-decoder pairs from the same edge to have different architecture types. This design decision enlarges the search space. Here, we demonstrate that the enlarged search space encompasses architectures that may outperform those with the same encoder-decoder architecture. 

Hence, we check the random configurations from Section~\ref{sec:search_space_analysis} and select all the configurations that contain $\Seq$ decoders. Then, for each decoder edge, instead of randomly sampling a new decoder component, we initialize the decoders with operations that are the same as those of the corresponding encoder nodes. We then reevaluate those architectures with the identical operations and compare them with the corresponding mixed architectures.

\begin{figure}
    \centering
    \includegraphics[width=0.425\linewidth]{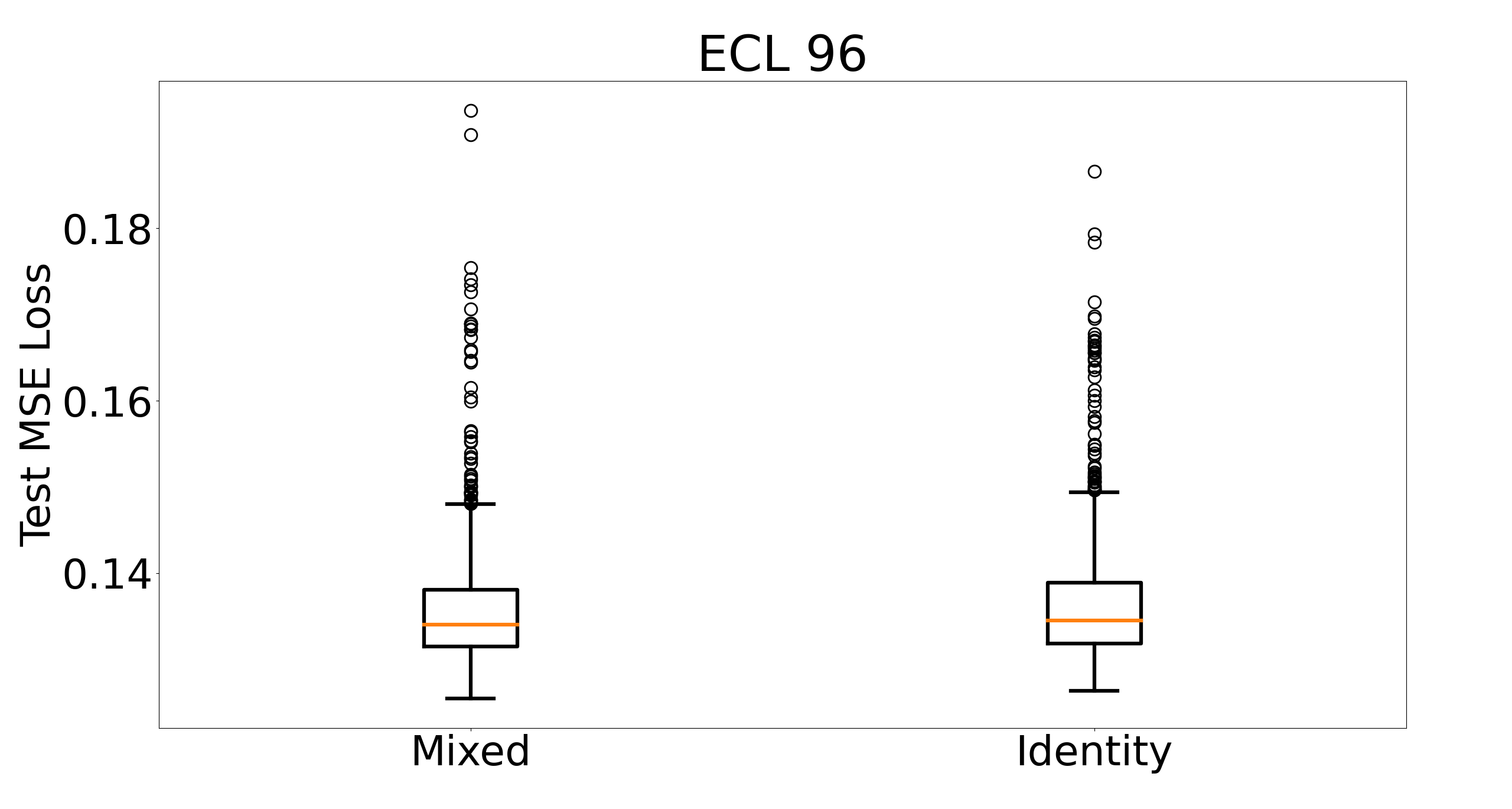}
    \includegraphics[width=0.425\linewidth]{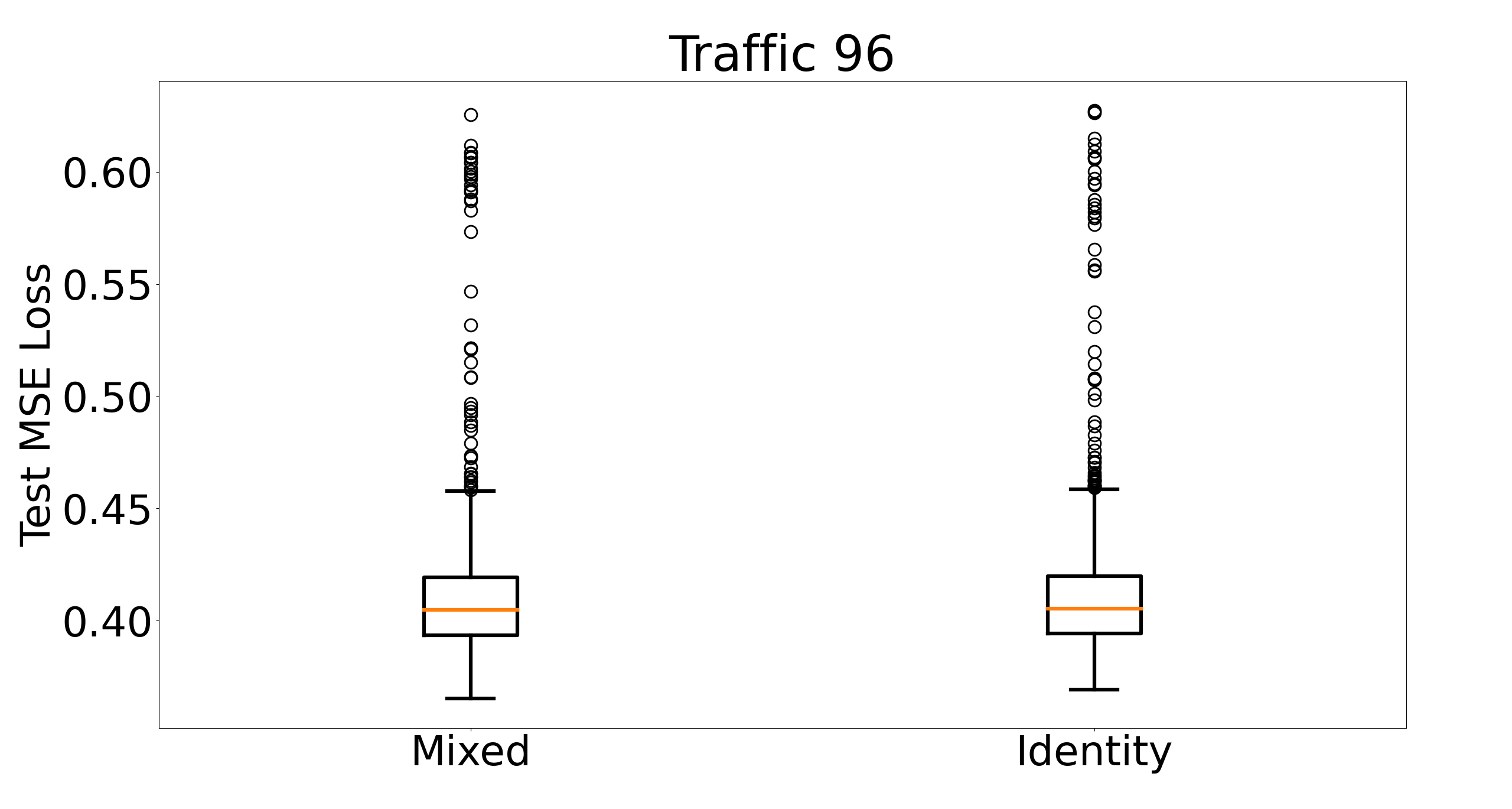}
    \caption{Performance of the architectures with mixed and identical  encoder-decoder architectures}
    \label{fig:ablation_mixedopsl}
\end{figure}

The result is shown in Figure~\ref{fig:ablation_mixedopsl}. On both datasets, architectures with mixed operations achieve a lower minimum test loss ($0.365 \ vs\ 0.369$ on the Traffic dataset and $0.125\ vs \ 0.126 $ on the ECL dataset). This shows that applying different operations to the encoders and decoders provides a potentially better model compared to the architectures that apply the same encoder and decoder operation within the search space.


\section{Optimal Architectures on other datasets~\label{sec:other_opt_archs}}
\begin{figure}[h]
    \centering
    \begin{subfigure}[b]{0.425\textwidth}
    \centering
    \includegraphics[width=\textwidth]{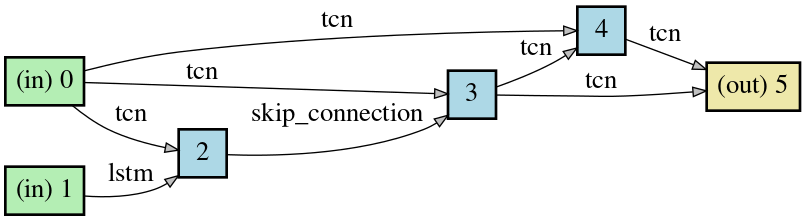} 
    \end{subfigure}
    \hfill
    \begin{subfigure}[b]{0.45\textwidth}
    \centering
    \includegraphics[width=\textwidth]{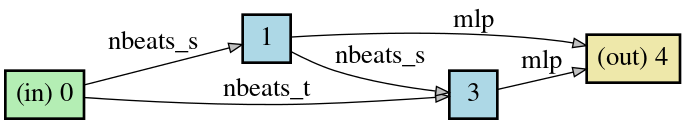}
    \end{subfigure}
    \caption{An optimal architecture on the ECL dataset. This is an encoder-only architecture.}
    \label{fig:opt_arch_ecl2}
\end{figure}

\begin{figure}[h!]
    \centering
    \begin{subfigure}[b]{0.425\textwidth}
    \centering
    \includegraphics[width=\textwidth]{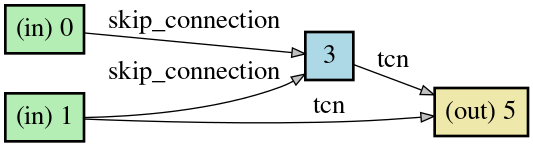} 
    \end{subfigure}
    \hfill
    \begin{subfigure}[b]{0.45\textwidth}
    \centering
    \includegraphics[width=\textwidth]{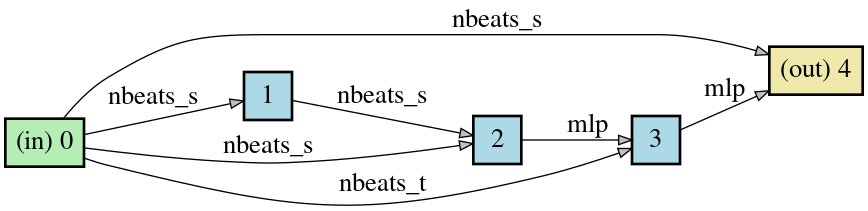}
    \end{subfigure}
    \caption{An optimal architecture on the Traffic dataset. This is an encoder-only architecture.}
    \label{fig:opt_arch_traffic}
\end{figure}

\begin{figure}[h!]
    \centering
    \begin{subfigure}[b]{0.425\textwidth}
    \centering
    \includegraphics[width=\textwidth]{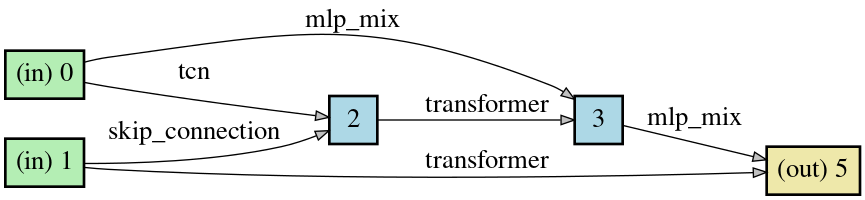} 
    \end{subfigure}
    \hfill
    \begin{subfigure}[b]{0.45\textwidth}
    \centering
    \includegraphics[width=\textwidth]{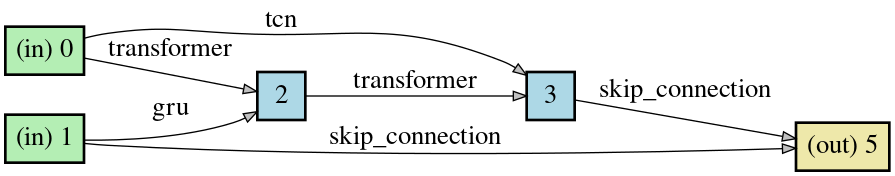}
    \end{subfigure}
    \begin{subfigure}[b]{0.45\textwidth}
    \centering
    \includegraphics[width=\textwidth]{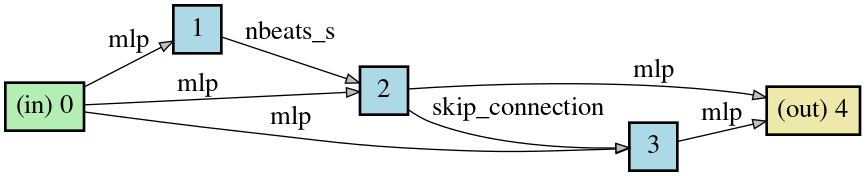}
    \end{subfigure}
    \caption{An optimal architecture on the ETTm1 dataset. This is an encoder-decoder architecture.}
    \label{fig:opt_arch_ettm1}
\end{figure}

\begin{figure}[h!]
    \centering
    \begin{subfigure}[b]{0.425\textwidth}
    \centering
    \includegraphics[width=\textwidth]{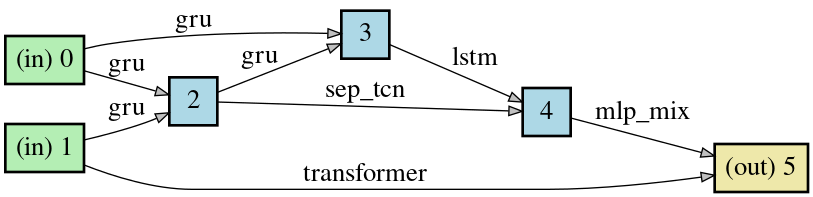} 
    \end{subfigure}
    \hfill
    \begin{subfigure}[b]{0.45\textwidth}
    \centering
    \includegraphics[width=\textwidth]{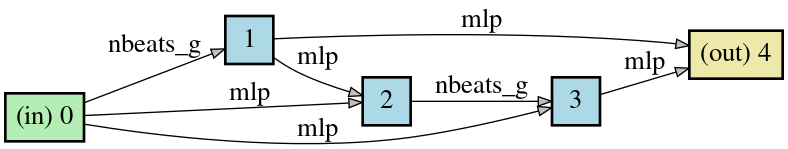}
    \end{subfigure}
    \caption{An optimal architecture on the ETTm2 dataset. This is an encoder-only architecture.}
    \label{fig:opt_arch_ettm2}
\end{figure}

\begin{figure}[h!]
    \centering
    \begin{subfigure}[b]{0.425\textwidth}
    \centering
    \includegraphics[width=\textwidth]{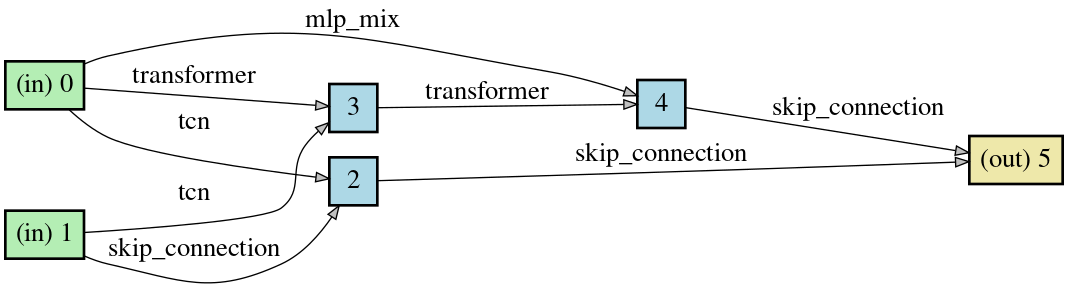} 
    \end{subfigure}
    \hfill
    \begin{subfigure}[b]{0.45\textwidth}
    \centering
    \includegraphics[width=\textwidth]{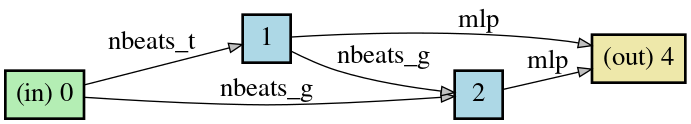}
    \end{subfigure}
    \caption{An optimal architecture on the Weather dataset. This is an encoder-only architecture.}
    \label{fig:opt_arch_weather}
\end{figure}

In Section~\ref{sec:opt_arch}, we showed one of the optimal architectures in the electricity dataset. In this section, we will provide more searched architectures. 

Figure~\ref{fig:opt_arch_ecl2} shows yet another optimal architecture on the ECL dataset. We can see that TCN and N-BEATS-Seasonal modules are still contained in the optimal modules, further confirming our conclusions in Section~\ref{sec:exp}.

We also show some of the optimal architectures in Figure~\ref{fig:opt_arch_traffic}, ~\ref{fig:opt_arch_ettm1}, ~\ref{fig:opt_arch_ettm2}, and ~\ref{fig:opt_arch_weather}.  We can see that no single operation dominates the optimal architecture, which shows the necessity of performing an architecture search for the optimal architecture.

\end{document}

%% file: commands/local.tex
\newcommand{\Seq}{\textit{Seq}}
\newcommand{\Flat}{\textit{Flat}}

\newcommand{\Ourname}{DARTS-TS}

\newcommand{\SeqNet}{\textit{Seq\ Net}}
\newcommand{\FlatNet}{\textit{Flat\ Net}}

\newcommand{\seqidx}{\mathit{i}}

\newcommand{\target}{\mathbf{y}}
\newcommand{\prediction}{\hat{\mathbf{y}}}

\newcommand{\featurespast}{\mathbf{x}}
\newcommand{\featuresfuture}{\hat{\mathbf{x}}}
\newcommand{\seqlength}{T}

\newcommand{\dsetseq}{\dset_{\seqidx}}

\newcommand{\targetseq}[2][\seqidx]{\target_{#1, #2}}
\newcommand{\predictionseq}[2][\seqidx]{\prediction_{#1, #2}}

\newcommand{\featurespastseq}[2][\seqidx]{\featurespast_{#1, #2}}
\newcommand{\featuresfutureseq}[2][\seqidx]{\featuresfuture_{#1, #2}}

\newcommand{\horizon}{H}

\newcommand{\seqlengthseq}[1][\seqidx]{\seqlength_{#1}}
\newcommand{\numseq}{N}

\newcommand{\observedintervalseq}{1:\seqlengthseq}
\newcommand{\forecastintervalseq}{\seqlengthseq +1 : \seqlengthseq + \horizon}

\newcommand{\archp}{\mathit{\alpha}}

%% file: commands/nomclat.tex
\usepackage{amsmath}
\usepackage{mathtools}  
\usepackage{amsfonts}

\DeclareMathOperator*{\argmin}{arg\,min}        

\newcommand{\dset}[0]{\mathcal{D}}                          

\newcommand{\weights}[0]{\mathbf{\theta}}                   



%% file: figures/tikz/search_space/cells_searchspace.tex
\begin{tikzpicture}[
every node/.style={scale=0.5},
    block/.style={rectangle, minimum width=10mm,minimum height=10mm, draw=black},
    optblock/.style={rectangle, minimum width=40mm, minimum height=4mm, draw=gray, dashed, rounded corners=.1cm},
    data/.style={rectangle, minimum width=45mm, minimum height=4.5mm, draw=black},
    bigoplus/.style={path picture={ \draw[black]
    (path picture bounding box.south) -- (path picture bounding box.north) (path picture bounding box.west) -- (path picture bounding box.east);
    }},
    dot/.style = {circle, fill, minimum size=3pt,inner sep = 0, outer sep=0pt,node contents={}},
    fill fraction/.style n args={2}{path picture={
 \fill[#1] (path picture bounding box.south west) rectangle
 ($(path picture bounding box.north west)!#2!(path picture bounding box.north
 east)$);}},]
    \node[block, fill=gray!5] (n0) {0};
    \node[block, fill=gray!5, right=8mm of n0,] (n1) {\large 1};

    \node[block, fill=gray!20] at ($(n0.west) + (-0.4, 1.2)$) (n2) {\large 2};

    \node[block, fill=gray!20] at ($(n1.east) + (0.45, 1.5)$) (n3) {\large 3};

    \node[block, fill=lightgray] at ($(n0.east) + (0.5, 2.7)$) (n4) {\large 4};

   \draw[red!50, ->] ($(n0.north) + (-0.15, 0.0)$) to [bend left=3] ($(n2.south west) + (0.05, 0.0)$);
   \draw[ blue!50, ->] ($(n0.north) + (-0.05, 0.0)$) to [bend left=3] ($(n2.south west) + (0.15, 0.0)$);

   \draw[red!50, ->] ($(n1.north) + (-0.1, 0.0)$) to [bend right=3]  ($(n2.south east) + (-0.3, 0.0)$);
   \draw[blue!50, ->] ($(n1.north) + (0.0, 0.0)$) to [bend right=3] ($(n2.south east) + (-0.2, 0.0)$);
    
   \draw[red!50, ->] ($(n0.north) + (0.0, 0.0)$) to [bend left=3]  ($(n3.south west) + (0.1, 0.0)$);
   \draw[blue!50, ->] ($(n0.north) + (0.1, 0.0)$) to [bend left=3] ($(n3.south west) + (0.2, 0.0)$);

    \draw[red!50, ->] ($(n1.north) + (0.0, 0.0)$) to [bend right=3]  ($(n3.south) + (0.1, 0.0)$);
   \draw[blue!50, ->] ($(n1.north) + (0.1, 0.0)$) to [bend right=3] ($(n3.south) + (0.2, 0.0)$);

    \draw[red!50, ->] ($(n2.east) + (0.0, 0.0)$) to [bend right=1] ($(n3.west) + (0.0, 0.1)$);
   \draw[blue!50, ->] ($(n2.east) + (0.0, -0.1)$) to[bend right=1] ($(n3.west) + (0.0, 0.0)$);

    \draw[red!50, ->] ($(n0.north) + (-0.1, 0.0)$) to [bend left=1] ($(n4.south) + (-0.15, 0.0)$);
   \draw[ blue!50, ->] ($(n0.north) + (-0.0, 0.0)$) to [bend left=1] ($(n4.south) + (-0.05, 0.0)$);

    \draw[red!50, ->] ($(n1.north) + (0.0, 0.0)$) to [bend right=1] ($(n4.south) + (0.05, 0.0)$);
   \draw[ blue!50, ->] ($(n1.north) + (0.1, 0.0)$) to [bend right=1] ($(n4.south) + (0.15, 0.0)$);

    \draw[red!50, ->] ($(n2.north) + (0.0, 0.0)$) to [bend left=1] ($(n4.west) + (0.0, -0.1)$);
   \draw[ blue!50, ->] ($(n2.north) + (0.1, 0.0)$) to [bend left=1] ($(n4.west) + (0.0, -0.2)$);

\draw[red!50, ->] ($(n3.north) + (-0.1, 0.0)$) to [bend left=1] ($(n4.east) + (0.0, -0.2)$);
   \draw[ blue!50, ->] ($(n3.north) + (0.0, 0.0)$) to [bend left=1] ($(n4.east) + (0.0, -0.1)$);

 \end{tikzpicture}

%% file: figures/tikz/search_space/cell_searched.tex
\begin{tikzpicture}[
every node/.style={scale=0.5},
    block/.style={rectangle, minimum width=10mm,minimum height=10mm, draw=black},
    optblock/.style={rectangle, minimum width=40mm, minimum height=4mm, draw=gray, dashed, rounded corners=.1cm},
    data/.style={rectangle, minimum width=45mm, minimum height=4.5mm, draw=black},
    bigoplus/.style={path picture={ \draw[black]
    (path picture bounding box.south) -- (path picture bounding box.north) (path picture bounding box.west) -- (path picture bounding box.east);
    }},
    dot/.style = {circle, fill, minimum size=3pt,inner sep = 0, outer sep=0pt,node contents={}},
    fill fraction/.style n args={2}{path picture={
 \fill[#1] (path picture bounding box.south west) rectangle
 ($(path picture bounding box.north west)!#2!(path picture bounding box.north
 east)$);}},]
    \node[block, fill=gray!5] (n0) {0};
    \node[block, fill=gray!5, right=5mm of n0,] (n1) {\large 1};

    \node[block, fill=gray!20] at ($(n0.west) + (-0.4, 1.2)$) (n2) {\large 2};

    \node[block, fill=gray!20] at ($(n1.east) + (0.45, 1.5)$) (n3) {\large 3};

    \node[block, fill=lightgray] at ($(n0.east) + (0.5, 2.7)$) (n4) {\large 4};

   \draw[red!50, ->] ($(n0.north) + (-0.15, 0.0)$) to [bend left=3] ($(n2.south west) + (0.05, 0.0)$);

   \draw[blue!50, ->] ($(n1.north) + (0.0, 0.0)$) to [bend right=3] ($(n2.south east) + (-0.2, 0.0)$);
    

    \draw[red!50, ->] ($(n1.north) + (0.0, 0.0)$) to [bend right=3]  ($(n3.south) + (0.1, 0.0)$);




    \draw[red!50, ->] ($(n2.north) + (0.0, 0.0)$) to [bend left=1] ($(n4.west) + (0.0, -0.1)$);

   \draw[ blue!50, ->] ($(n3.north) + (0.0, 0.0)$) to [bend left=1] ($(n4.east) + (0.0, -0.1)$);

 \end{tikzpicture}

%% file: figures/tikz/search_space/flat_net.tex
    \begin{tikzpicture}[
    every node/.style={scale=0.57},
    block/.style={rectangle, minimum width=40mm, minimum height=4mm, draw=black, rounded corners=.1cm},
    optblock/.style={rectangle, minimum width=60mm, minimum height=4mm, draw=gray, dashed, rounded corners=.1cm},
    data/.style={rectangle, minimum width=40mm, minimum height=5.5mm, draw=black},
    dot/.style = {circle, fill, minimum size=3pt,inner sep = 0, outer sep=0pt,node contents={}},]

    \node[data] (inputpast) {Past Targets};
    \node[data,  fill=blue!10, align=left, above=2mm of inputpast] (ptprocessed) {Transposed Past Targets};
    \node[data, fill=orange!10, right=0mm of ptprocessed] (futurepadding) {Future Placeholder};

    \draw[->] (inputpast) -> (ptprocessed) node [midway, right] () {Transposed};

    \node[block, above=2mm of ptprocessed.north east] (encoder1) {Encoder Layer 1};


    \draw[->] ($(ptprocessed.north east) + (-0.4, 0)$) -- ($(encoder1.south) + (-0.4, 0)$);
    \draw[->] ($(futurepadding.north west) + (0.4, 0)$) -- ($(encoder1.south) + (0.4, 0)$);

    \node[data, fill=blue!10, above=7mm of ptprocessed]  (backcast1) {Backcast Output Layer 1};
    \node[data, fill=orange!10, right=0mm of backcast1]  (forecast1) {Forecast Output Layer 1};

    \draw[->] ($(encoder1.north) + (-0.4, 0)$) -- ($(backcast1.south east) + (-0.4, 0)$);
    \draw[->] ($(encoder1.north) + (0.4, 0)$) -- ($(forecast1.south west) + (0.4, 0)$);

    \node[above =1.5mm of backcast1.north east] (dotsencoder) {\rotatebox{90}{\large $\cdots$}};

    \draw[->] ($(backcast1.north east) + (-0.4, 0)$) -- ($(backcast1.north east) + (-0.4, 0.15)$);
    \draw[->] ($(forecast1.north west) + (0.4, 0)$) -- ($(forecast1.north west) + (0.4, 0.15)$);
    \node[block, above=1.5mm of dotsencoder] (encodern) {Encoder Layer n};

    \draw[->] ($(encodern.south) + (0.4, -0.15)$) -- ($(encodern.south) + (0.4, 0)$);
    \draw[->] ($(encodern.south) + (-0.4, -0.15)$) -- ($(encodern.south) + (-0.4, 0)$);

    \node[data, fill=blue!10, above=13mm of backcast1]  (backcastn) {Backcast Output Layer N};
    \node[data, fill=orange!10, right=0mm of backcastn]  (forecastn) {Forecast Output Layer N};

    \draw[->] ($(encodern.north) + (-0.4, 0)$) -- ($(backcastn.south east) + (-0.4, 0)$);
    \draw[->] ($(encodern.north) + (0.4, 0)$) -- ($(forecastn.south west) + (0.4, 0)$);

    \node[block, minimum width=30mm, right= 3mm of forecastn] (head) {Forecasting Head};
    \node[data,  minimum width=30mm, above=3mm of head] (forecasting) {Forecasting};

    \draw[->] (forecastn) -> (head);
    \draw[->] (head) -> (forecasting) node [midway, right] () {Transposed};


    \node[data, minimum width=4mm, fill=blue!10!white] at ($(forecast1.east) + (0.5, 0.4)$) (legendbackcast) {};
    \node[right=0mm of legendbackcast.east, align=left] (legendbackcasttext) {Backcast \\ Feature Maps};
    \node[data, minimum width=4mm, below = 4.5mm of legendbackcast, fill=orange!15!white] (legendforecast) {};
    \node[right=0mm of legendforecast.east, align=left] (legendforecasttext) {Forecast \\ Feature Maps};

    \end{tikzpicture}

%% file: figures/tikz/search_space/seq_net.tex
\begin{tikzpicture}[
every node/.style={scale=0.57},
    block/.style={rectangle, minimum width=30mm, minimum height=4mm, draw=black, rounded corners=.1cm},
    optblock/.style={rectangle, minimum width=40mm, minimum height=4mm, draw=gray, dashed, rounded corners=.1cm},
    data/.style={rectangle, minimum width=30mm, minimum height=4.5mm, draw=black},
    dot/.style = {circle, fill, minimum size=3pt,inner sep = 0, outer sep=0pt,node contents={}},
    fill fraction/.style n args={2}{path picture={
 \fill[#1] (path picture bounding box.south west) rectangle
 ($(path picture bounding box.north west)!#2!(path picture bounding box.north
 east)$);}},
    fill fraction3/.style n args={3}{path picture={
 \fill[#1] (path picture bounding box.south west) rectangle
 ($(path picture bounding box.west)!#3!(path picture bounding box.east)$);
 \fill[#2] (path picture bounding box.west) rectangle
 ($(path picture bounding box.north west)!#3!(path picture bounding box.north east)$);}},
    ]
    
    \node[block] (encoder11) {Encoder Layer 1};

    \node[data, minimum width=10mm, right= 2 mm of encoder11, fill=cyan!10!white] (lout11) {Output Layer 1};
    \node[data, above=3mm of encoder11, fill={cyan!15}, fill fraction={blue!15}{0.85}] (sout1) {Output Encoder Layer 1};

    \node[block, minimum height=6mm, right=5.5mm of lout11] (decoder1) {Decoder Layer 1};

    \draw[->, dashed] (lout11.east) -- ($(lout11.east) + (0.1, 0)$)  |- ($(decoder1.west) + (0.0, -0.1)$);
    \draw[->, dashed] ($(sout1.south)!.5!(sout1.south east)$) -- ($(sout1.south)!.5!(sout1.south east) + (0, -0.25)$) -| ($(lout11.east) + (0.225, 0)$) -- (decoder1.west);
    
    \draw[->, dashed] ($(sout1.south)!.85!(sout1.south east)$) -- ($(sout1.south)!.85!(sout1.south east) + (0, -0.1)$) -| ($(lout11.east) + (0.35, 0.1)$) -- ($(decoder1.west) + (0, 0.1)$);
    
    \node[data, below=3mm of encoder11] (inputpast) {Past Input};
    
    \node[data] at (inputpast.east -| decoder1.south) (inputfuture) {Known Future Input};

    \node[above =2mm of sout1] (dotsencoder) {\rotatebox{90}{\large $\cdots$}};

    \node[block, above = 2mm of dotsencoder] (encoder21)  {Encoder Layer N};

    \node[data, minimum width=10mm, right= 2mm of encoder21, fill=cyan!10!white] (lout21) {Output Layer N};

    \node[data, above=3mm of encoder21, fill={cyan!15}, 
    fill fraction3={blue!15}{orange!15}{0.85}] (sout2) {Output Encoder Block N};

    \node[] at (dotsencoder.west -| decoder1.north) (dotsdecoder) {\rotatebox{90}{\large $\cdots$}};

    \node[block, minimum height=6mm] at (decoder1.north |- encoder21.east) (decoder2) {Decoder Block N};

    \node[data] at (sout1.east -| decoder1.north) (decoderout1) {Output Decoder Block 1};
    \node[data] at (sout2.east -| decoder2.north) (decoderout2) {Output Decoder Block N};

    \draw[->, dashed] (lout21.east) -- ($(lout21.east) + (0.1, 0)$)  |- ($(decoder2.west) + (0.0, -0.1)$);
    \draw[->, dashed] ($(sout2.south)!.5!(sout2.south east)$) -- ($(sout2.south)!.5!(sout2.south east) + (0, -0.25)$) -| ($(lout21.east) + (0.225, 0)$) -- (decoder2.west);
    
    \draw[->, dashed] ($(sout2.south)!.85!(sout2.south east)$) -- ($(sout2.south)!.85!(sout2.south east) + (0, -0.1)$) -| ($(lout21.east) + (0.35, 0.1)$) -- ($(decoder2.west) + (0, 0.1)$);

    \draw[->] (inputpast.north) -> (encoder11.south);
    \draw[->] (encoder11.north) -> (sout1.south);

     \draw[->] (encoder11.east) -> (lout11.west);

    \draw[->] (sout1.north) -> (dotsencoder.south);
    \draw[->] (dotsencoder.north) -> (encoder21.south);
    \draw[->] (encoder21.north) -> (sout2.south);

    \draw[->] (encoder21.east) -> (lout21.west);

    \draw[->] (inputfuture.north) -> (decoder1.south);
    \draw[->] (decoder1.north) -> (decoderout1.south);

    \draw[->] (decoderout1.north) -> (dotsdecoder.south);
    \draw[->] (dotsdecoder.north) -> (decoder2.south);

    \node[block, minimum width=20mm,above= 3mm of decoderout2] (head) {Forecasting Head};
    \draw[->] (decoder2.north) -> (decoderout2.south);
    \draw[->, dashed] (decoderout2) -> (head);

    \node[block] at (sout2.north |- head.east) (flatdec) {Flat Decoder};
    \draw[->, dashed] (sout2) -> (flatdec);
    \draw[->, dashed] (flatdec) -> (head);

        \node[data,minimum width=20mm, right=4.5mm of head] (res) {Forecasting};
    \draw[->] (head) -> (res);

    \node[data, minimum width=4mm, fill=cyan!10!white] at ($(decoderout1.east) + (0.3, 0.6)$) (legendrnn) {};
    \node[right=0mm of legendrnn.east, align=left] (legrnntext) {\footnotesize Feature Maps \\ \footnotesize for RNN \\ \footnotesize Decoders};
    \node[data, minimum width=4mm, below = 6mm of legendrnn, fill=orange!15!white] (legendtransformer) {};
    \node[right=0mm of legendtransformer.east, align=left] (legendtransformertext) {\footnotesize Feature Maps \\ \footnotesize for Transformer \\ \footnotesize Decoders};

    \node[data, minimum width=4mm,  below = 6mm of legendtransformer, fill=blue!15!white] (legendtcn) {};
    \node[right=0mm of legendtcn.east, align=left] (legendtcntext) {\footnotesize Feature Maps \\ \footnotesize for TCN \\ \small Decoders};

    \end{tikzpicture}

%% file: figures/tikz/search_space/hybrid.tex
\begin{tikzpicture}[
every node/.style={scale=0.50},
    block/.style={rectangle, minimum width=30mm, minimum height=4mm, draw=black, rounded corners=.1cm},
    optblock/.style={rectangle, minimum width=40mm, minimum height=4mm, draw=gray, dashed, rounded corners=.1cm},
    data/.style={rectangle, minimum width=45mm, minimum height=4.5mm, draw=black},
    bigoplus/.style={path picture={ \draw[black]
    (path picture bounding box.south) -- (path picture bounding box.north) (path picture bounding box.west) -- (path picture bounding box.east);
    }},
    dot/.style = {circle, fill, minimum size=3pt,inner sep = 0, outer sep=0pt,node contents={}},
    fill fraction/.style n args={2}{path picture={
 \fill[#1] (path picture bounding box.south west) rectangle
 ($(path picture bounding box.north west)!#2!(path picture bounding box.north
 east)$);}},]
    \node[data] (inputpast) {Past Targets};

    \node[block, above=3mm of inputpast] (flatnet) {$\FlatNet$};

    \node[data, above=3mm of flatnet] (flatoutbackcast)  {Backcast Output $\FlatNet$};
    \node[data,right=0mm of flatoutbackcast](flatoutforecast)  {Forecast Output $\FlatNet$};
    \node[data, above=0mm of flatoutforecast] (inputfuture) {Known Future Input};

     \node[block, minimum width=70mm, above=3mm of inputfuture.north west] (seqnet) {$\SeqNet$};

    \draw[->] (inputpast) -> (flatnet);
    \draw[->] ($(flatnet.north) + (-0.2,0.0) $) -- ($(flatoutbackcast.south) + (-0.2, 0.0)$);
    \draw[->] ($(flatnet.north) + (0.2,0.0)$) -- ($(flatnet.north) + (0.2,0.15)$) -| ($(flatoutforecast.south)$);

    \draw[->] (inputpast.west) -- ($(inputpast.west) + (-0.3, 0.0)$) |- ($(seqnet.south) + (-1.5, -0.3)$) -- ($(seqnet.south) + (-1.5, -0.0)$);
    \draw[->] (inputfuture) -- (inputfuture.north |- seqnet.south);

    \node[bigoplus,draw,circle,minimum width=2.5mm] at ($(seqnet.north east) + (0.7, 0.3)$) (plus) {};
    \draw[->] (flatoutforecast.east) -| (plus);
    \draw[->] (seqnet.north -| flatoutforecast.north) |- (plus);

    \node[data, minimum width=20mm,above=3mm of plus] (out) {Forecasting};
    \draw[->] (plus) -- (out);

 \end{tikzpicture}

%% file: tables/mean_table.tex
\begin{tabular}{lllllllllllllllll}
\toprule
 & \multicolumn{2}{c}{DARTS-TS} & \multicolumn{2}{c}{iTransformer} & \multicolumn{2}{c}{ModernTCN} & \multicolumn{2}{c}{PatchTST} & \multicolumn{2}{c}{TSMixer} & \multicolumn{2}{c}{DLinear} & \multicolumn{2}{c}{TimesNet} & \multicolumn{2}{c}{Autoformer} \\
 & MSE & MAE & MSE & MAE & MSE & MAE & MSE & MAE & MSE & MAE & MSE & MAE & MSE & MAE & MSE & MAE \\
\midrule
ETTm1 & \textcolor{red}{0.344} & \textcolor{red}{0.368} & 0.368 & 0.395 & 0.362 & 0.386 & \underline{0.352} & \underline{0.381} & 0.382 & 0.408 & 0.360 & 0.382 & 0.402 & 0.415 & 0.619 & 0.539 \\
ETTm2 & \textcolor{red}{0.253} & \textcolor{red}{0.306} & 0.273 & 0.330 & 0.261 & 0.319 & \underline{0.257} & \underline{0.315} & 0.446 & 0.477 & 0.267 & 0.329 & 0.290 & 0.339 & 0.423 & 0.441 \\
ETTh1 & \underline{0.413} & \underline{0.423} & 0.473 & 0.468 & \textcolor{red}{0.404} & \textcolor{red}{0.421} & 0.415 & 0.429 & 0.515 & 0.503 & 0.447 & 0.456 & 0.486 & 0.482 & 0.559 & 0.529 \\
ETTh2 & 0.351 & 0.388 & 0.387 & 0.415 & \underline{0.333} & \underline{0.385} & \textcolor{red}{0.330} & \textcolor{red}{0.379} & 0.571 & 0.548 & 0.422 & 0.439 & 0.399 & 0.435 & 0.739 & 0.621 \\
ECL & \textcolor{red}{0.156} & \textcolor{red}{0.245} & 0.166 & 0.261 & 0.163 & 0.257 & \underline{0.161} & \underline{0.254} & 0.168 & 0.272 & 0.166 & 0.264 & 0.203 & 0.302 & 0.221 & 0.334 \\
Exchange & \underline{0.378} & \textcolor{red}{0.409} & 0.411 & 0.440 & 0.525 & 0.505 & 0.385 & \underline{0.417} & \textcolor{red}{0.366} & 0.452 & 0.382 & 0.419 & 0.540 & 0.524 & 1.010 & 0.775 \\
Weather & 0.230 & \textcolor{red}{0.262} & 0.239 & 0.274 & 0.231 & 0.269 & \underline{0.229} & \underline{0.265} & \textcolor{red}{0.222} & 0.288 & 0.244 & 0.297 & 0.249 & 0.287 & 0.398 & 0.431 \\
Traffic & \underline{0.394} & \textcolor{red}{0.260} & \textcolor{red}{0.387} & 0.273 & 0.421 & 0.287 & 0.398 & \underline{0.267} & 0.541 & 0.413 & 0.434 & 0.295 & 0.624 & 0.336 & 0.671 & 0.412 \\
\bottomrule
\end{tabular}

%% file: tables/mean_pems.tex
\begin{tabular}{lllllllllllllllll}
\toprule
 & \multicolumn{2}{c}{DARTS-TS} & \multicolumn{2}{c}{iTransformer} & \multicolumn{2}{c}{ModernTCN} & \multicolumn{2}{c}{PatchTST} & \multicolumn{2}{c}{TSMixer} & \multicolumn{2}{c}{DLinear} & \multicolumn{2}{c}{TimesNet} & \multicolumn{2}{c}{Autoformer} \\
 & MSE & MAE & MSE & MAE & MSE & MAE & MSE & MAE & MSE & MAE & MSE & MAE & MSE & MAE & MSE & MAE \\
\midrule
PEMS03 & \textcolor{red}{0.137} & \textcolor{red}{0.241} & 0.458 & 0.408 & 0.408 & 0.419 & 0.199 & 0.291 & 0.162 & 0.281 & 0.264 & 0.358 & \underline{0.151} & \underline{0.248} & 0.554 & 0.541 \\
PEMS04 & \textcolor{red}{0.112} & \textcolor{red}{0.221} & \underline{0.127} & \underline{0.237} & 0.488 & 0.471 & 0.266 & 0.338 & 0.136 & 0.254 & 0.264 & 0.355 & 0.127 & 0.239 & 0.763 & 0.669 \\
PEMS07 & \textcolor{red}{0.102} & \textcolor{red}{0.204} & 0.504 & 0.477 & 0.304 & 0.374 & 0.209 & 0.293 & 0.150 & 0.251 & 0.311 & 0.373 & 0.132 & 0.233 & 0.361 & 0.438 \\
PEMS08 & \textcolor{red}{0.169} & \textcolor{red}{0.256} & 0.202 & 0.269 & 0.510 & 0.479 & 0.231 & 0.304 & 0.225 & 0.305 & 0.331 & 0.376 & \underline{0.191} & \underline{0.267} & 0.739 & 0.627 \\
\bottomrule
\end{tabular}

%% file: appendix/acknowledgement.tex
\section{Acknowledgement}
The authors gratefully acknowledge the computing time provided to them on the high-performance computers Noctua2 at the NHR Center PC2 under the project hpc-prf-intexml. These are funded by the Federal Ministry of Education and Research and the state governments participating on the basis of the resolutions of the GWK for the national high performance computing at universities (www.nhr-verein.de/unsere-partner).

Difan Deng was supported by the Federal Ministry of Education and Research (BMBF) under the project AI service center KISSKI (grant no.01IS22093C).

%% file: tables/search_time/search_time.tex
\begin{tabular}{lrrrrrrrr}
\toprule
 & ECL & ETTh1 & ETTh2 & ETTm1 & ETTm2 & Traffic & Exchange & Weather \\
\midrule
Optimization & 3.665 & 0.845 & 0.845 & 3.512 & 3.594 & 3.187 & 0.818 & 3.273 \\
Operation Prune & 2.672 & 0.730 & 0.738 & 2.984 & 3.024 & 2.445 & 0.584 & 2.316 \\
Topology Prune & 0.088 & 0.024 & 0.030 & 0.106 & 0.106 & 0.107 & 0.022 & 0.078 \\
Search & 6.424 & 1.599 & 1.612 & 6.602 & 6.724 & 5.738 & 1.424 & 5.666 \\
\bottomrule
\end{tabular}

%% file: tables/flops_changes.tex
\begin{tabular}{lrrrrrrrr}
\toprule
 & ECL & ETTh1 & ETTh2 & ETTm1 & ETTm2 & Traffic & Exchange & Weather \\
\midrule
Supernet & 26.241 & 0.750 & 0.750 & 0.750 & 0.750 & 61.841 & 0.778 & 1.149 \\
After OP Prune & 4.991 & 0.078 & 0.079 & 0.061 & 0.068 & 15.237 & 0.087 & 0.158 \\
After Topology Prune & 4.358 & 0.042 & 0.060 & 0.031 & 0.040 & 11.911 & 0.052 & 0.079 \\
\bottomrule
\end{tabular}

%% file: tables/full_res.tex
\begin{tabular}{llllllllllllllllll}
\toprule
 &  & \multicolumn{2}{c}{DARTS-TS} & \multicolumn{2}{c}{iTransformer} & \multicolumn{2}{c}{ModernTCN} & \multicolumn{2}{c}{PatchTST} & \multicolumn{2}{c}{TSMixer} & \multicolumn{2}{c}{DLinear} & \multicolumn{2}{c}{TimesNet} & \multicolumn{2}{c}{Autoformer} \\
 &  & MSE & MAE & MSE & MAE & MSE & MAE & MSE & MAE & MSE & MAE & MSE & MAE & MSE & MAE & MSE & MAE \\
\midrule
\multirow[t]{4}{*}{ECL} & 96 & \textcolor{red}{0.129 (0.00)} & \textcolor{red}{0.217 (0.00)} & 0.132 (0.00) & 0.228 (0.00) & 0.135 (0.00) & 0.231 (0.00) & \underline{0.130 (0.00)} & 0.223 (0.00) & 0.137 (0.00) & 0.241 (0.00) & 0.140 (0.00) & 0.237 (0.00) & 0.184 (0.00) & 0.287 (0.00) & 0.204 (0.01) & 0.320 (0.01) \\
 & 192 & \textcolor{red}{0.147 (0.00)} & \textcolor{red}{0.234 (0.00)} & 0.155 (0.00) & 0.249 (0.00) & 0.149 (0.00) & 0.243 (0.00) & 0.148 (0.00) & 0.241 (0.00) & 0.156 (0.00) & 0.260 (0.00) & 0.153 (0.00) & 0.250 (0.00) & 0.194 (0.00) & 0.295 (0.00) & 0.212 (0.00) & 0.327 (0.00) \\
 & 336 & \textcolor{red}{0.164 (0.00)} & \textcolor{red}{0.253 (0.00)} & 0.171 (0.00) & 0.266 (0.00) & \underline{0.165 (0.00)} & 0.259 (0.00) & \underline{0.165 (0.00)} & 0.259 (0.00) & 0.173 (0.00) & 0.281 (0.00) & 0.169 (0.00) & 0.267 (0.00) & 0.197 (0.00) & 0.299 (0.00) & 0.215 (0.01) & 0.328 (0.01) \\
 & 720 & \textcolor{red}{0.186 (0.00)} & \textcolor{red}{0.275 (0.00)} & 0.206 (0.01) & 0.299 (0.01) & 0.205 (0.00) & 0.295 (0.00) & 0.202 (0.00) & 0.292 (0.00) & 0.206 (0.00) & 0.308 (0.00) & 0.203 (0.000) & 0.301 (0.000)) & 0.236 (0.03) & 0.329 (0.02) & 0.254 (0.01) & 0.360 (0.01) \\
\cline{1-18}
\multirow[t]{4}{*}{ETTh1} & 96 & \textcolor{red}{0.365 (0.00)} & \textcolor{red}{0.385 (0.00)} & 0.404 (0.00) & 0.419 (0.00) & \underline{0.369 (0.00)} & 0.394 (0.00) & 0.375 (0.00) & 0.400 (0.00) & 0.387 (0.00) & 0.413 (0.01) & 0.379 (0.01) & 0.403 (0.01) & 0.443 (0.01) & 0.453 (0.01) & 0.496 (0.01) & 0.492 (0.01) \\
 & 192 & \textcolor{red}{0.405 (0.00)} & \textcolor{red}{0.411 (0.00)} & 0.451 (0.00) & 0.449 (0.00) & \underline{0.407 (0.00)} & \underline{0.415 (0.00)} & 0.413 (0.00) & 0.420 (0.00) & 0.428 (0.01) & 0.439 (0.01) & 0.415 (0.01) & 0.427 (0.01) & 0.486 (0.01) & 0.482 (0.01) & 0.532 (0.04) & 0.509 (0.02) \\
 & 336 & 0.437 (0.01) & 0.433 (0.01) & 0.471 (0.00) & 0.465 (0.00) & \textcolor{red}{0.392 (0.00)} & \textcolor{red}{0.413 (0.00)} & 0.427 (0.00) & 0.432 (0.00) & 0.505 (0.01) & 0.501 (0.01) & 0.470 (0.03) & 0.469 (0.03) & 0.482 (0.01) & 0.478 (0.01) & 0.544 (0.03) & 0.523 (0.01) \\
 & 720 & \underline{0.448 (0.01)} & \underline{0.462 (0.00)} & 0.565 (0.02) & 0.538 (0.01) & 0.450 (0.00) & \textcolor{red}{0.461 (0.00)} & \textcolor{red}{0.444 (0.00)} & \underline{0.463 (0.00)} & 0.741 (0.09) & 0.658 (0.03) & 0.524 (0.02) & 0.527 (0.01) & 0.534 (0.03) & 0.515 (0.02) & 0.662 (0.14) & 0.592 (0.06) \\
\cline{1-18}
\multirow[t]{4}{*}{ETTh2} & 96 & 0.276 (0.00) & \textcolor{red}{0.332 (0.00)} & 0.305 (0.00) & 0.361 (0.00) & \textcolor{red}{0.264 (0.00)} & 0.333 (0.00) & 0.275 (0.00) & 0.336 (0.00) & 0.370 (0.01) & 0.436 (0.01) & 0.283 (0.00) & 0.347 (0.00) & 0.363 (0.03) & 0.408 (0.02) & 0.517 (0.05) & 0.534 (0.04) \\
 & 192 & 0.344 (0.00) & \underline{0.377 (0.00)} & 0.389 (0.01) & 0.411 (0.00) & \textcolor{red}{0.322 (0.00)} & \textcolor{red}{0.377 (0.00)} & 0.339 (0.00) & 0.379 (0.00) & 0.494 (0.03) & 0.510 (0.02) & 0.366 (0.02) & 0.403 (0.01) & 0.411 (0.02) & 0.437 (0.01) & 0.565 (0.09) & 0.559 (0.05) \\
 & 336 & 0.377 (0.01) & 0.406 (0.00) & 0.420 (0.01) & 0.434 (0.01) & \textcolor{red}{0.315 (0.00)} & \textcolor{red}{0.377 (0.00)} & 0.328 (0.00) & 0.381 (0.00) & 0.586 (0.02) & 0.559 (0.01) & 0.428 (0.02) & 0.450 (0.01) & 0.394 (0.02) & 0.438 (0.01) & 0.757 (0.14) & 0.642 (0.07) \\
 & 720 & 0.406 (0.01) & 0.435 (0.00) & 0.436 (0.01) & 0.454 (0.00) & 0.429 (0.00) & 0.453 (0.00) & \textcolor{red}{0.378 (0.00)} & \textcolor{red}{0.420 (0.00)} & 0.837 (0.07) & 0.688 (0.03) & 0.610 (0.06) & 0.555 (0.03) & 0.429 (0.03) & 0.458 (0.01) & 1.115 (0.17) & 0.751 (0.06) \\
\cline{1-18}
\multirow[t]{4}{*}{ETTm1} & 96 & \textcolor{red}{0.283 (0.00)} & \textcolor{red}{0.330 (0.00)} & 0.305 (0.00) & 0.358 (0.00) & 0.296 (0.00) & 0.348 (0.00) & 0.290 (0.00) & 0.341 (0.00) & 0.308 (0.01) & 0.358 (0.01) & 0.301 (0.00) & 0.346 (0.00) & 0.330 (0.01) & 0.373 (0.00) & 0.497 (0.04) & 0.487 (0.02) \\
 & 192 & \textcolor{red}{0.320 (0.00)} & \textcolor{red}{0.354 (0.00)} & 0.343 (0.00) & 0.380 (0.00) & 0.348 (0.00) & 0.378 (0.00) & 0.333 (0.00) & 0.369 (0.00) & 0.347 (0.01) & 0.386 (0.01) & 0.337 (0.00) & 0.368 (0.00) & 0.430 (0.05) & 0.423 (0.02) & 0.591 (0.03) & 0.528 (0.01) \\
 & 336 & \textcolor{red}{0.357 (0.00)} & \textcolor{red}{0.377 (0.00)} & 0.380 (0.00) & 0.402 (0.00) & 0.376 (0.00) & 0.395 (0.00) & 0.367 (0.00) & 0.391 (0.00) & 0.398 (0.01) & 0.420 (0.01) & 0.374 (0.00) & 0.392 (0.00) & 0.397 (0.00) & 0.416 (0.00) & 0.682 (0.06) & 0.561 (0.02) \\
 & 720 & \underline{0.418 (0.01)} & \textcolor{red}{0.412 (0.00)} & 0.441 (0.00) & 0.438 (0.00) & 0.430 (0.00) & 0.421 (0.00) & \textcolor{red}{0.417 (0.00)} & 0.422 (0.00) & 0.474 (0.03) & 0.470 (0.02) & 0.427 (0.00) & 0.423 (0.00) & 0.450 (0.01) & 0.446 (0.01) & 0.703 (0.09) & 0.579 (0.03) \\
\cline{1-18}
\multirow[t]{4}{*}{ETTm2} & 96 & \textcolor{red}{0.162 (0.00)} & \textcolor{red}{0.244 (0.00)} & 0.179 (0.00) & 0.268 (0.00) & 0.170 (0.00) & 0.257 (0.00) & 0.165 (0.00) & 0.254 (0.00) & 0.182 (0.01) & 0.293 (0.01) & 0.166 (0.00) & 0.258 (0.00) & 0.190 (0.01) & 0.277 (0.00) & 0.332 (0.03) & 0.392 (0.02) \\
 & 192 & \underline{0.223 (0.00)} & \textcolor{red}{0.285 (0.00)} & 0.242 (0.00) & 0.312 (0.00) & 0.225 (0.00) & 0.297 (0.00) & \textcolor{red}{0.222 (0.00)} & 0.293 (0.00) & 0.284 (0.03) & 0.386 (0.03) & \underline{0.224 (0.00)} & 0.301 (0.00) & 0.247 (0.01) & 0.313 (0.00) & 0.380 (0.08) & 0.414 (0.04) \\
 & 336 & \textcolor{red}{0.274 (0.00)} & \textcolor{red}{0.320 (0.00)} & 0.292 (0.00) & 0.344 (0.00) & 0.283 (0.00) & 0.335 (0.00) & \underline{0.277 (0.00)} & 0.329 (0.00) & 0.499 (0.04) & 0.535 (0.02) & 0.280 (0.00) & 0.339 (0.01) & 0.312 (0.02) & 0.354 (0.01) & 0.432 (0.05) & 0.452 (0.03) \\
 & 720 & \textcolor{red}{0.354 (0.00)} & \textcolor{red}{0.373 (0.00)} & 0.380 (0.01) & 0.397 (0.00) & 0.367 (0.00) & 0.386 (0.00) & 0.364 (0.00) & 0.383 (0.00) & 0.818 (0.03) & 0.695 (0.01) & 0.397 (0.01) & 0.416 (0.00) & 0.413 (0.01) & 0.411 (0.00) & 0.547 (0.09) & 0.506 (0.05) \\
\cline{1-18}
\multirow[t]{4}{*}{Exchange} & 96 & \underline{0.088 (0.00)} & 0.210 (0.00) & 0.099 (0.00) & 0.227 (0.00) & 0.169 (0.00) & 0.305 (0.00) & 0.093 (0.00) & 0.213 (0.00) & 0.113 (0.00) & 0.259 (0.01) & \textcolor{red}{0.084 (0.00)} & \textcolor{red}{0.203 (0.00)} & 0.167 (0.01) & 0.305 (0.01) & 0.541 (0.12) & 0.560 (0.07) \\
 & 192 & 0.186 (0.00) & 0.308 (0.00) & 0.202 (0.00) & 0.326 (0.00) & 0.276 (0.00) & 0.389 (0.00) & 0.192 (0.00) & 0.312 (0.00) & 0.237 (0.03) & 0.378 (0.01) & \textcolor{red}{0.164 (0.01)} & \textcolor{red}{0.293 (0.00)} & 0.309 (0.02) & 0.415 (0.01) & 0.956 (0.24) & 0.770 (0.12) \\
 & 336 & \underline{0.350 (0.01)} & \textcolor{red}{0.428 (0.01)} & 0.397 (0.01) & 0.466 (0.01) & 0.449 (0.00) & 0.505 (0.00) & \textcolor{red}{0.350 (0.00)} & \underline{0.431 (0.00)} & 0.443 (0.08) & 0.519 (0.03) & \underline{0.355 (0.01)} & 0.453 (0.00) & 0.487 (0.02) & 0.534 (0.01) & 1.290 (0.23) & 0.903 (0.08) \\
 & 720 & 0.888 (0.03) & \underline{0.689 (0.01)} & 0.947 (0.01) & 0.740 (0.00) & 1.206 (0.02) & 0.821 (0.01) & 0.906 (0.00) & 0.713 (0.00) & \textcolor{red}{0.671 (0.11)} & \textcolor{red}{0.653 (0.04)} & 0.927 (0.05) & 0.727 (0.02) & 1.197 (0.07) & 0.842 (0.02) & 1.254 (0.03) & 0.866 (0.01) \\
\cline{1-18}
\multirow[t]{4}{*}{Traffic} & 96 & \underline{0.358 (0.00)} & \textcolor{red}{0.240 (0.01)} & \textcolor{red}{0.356 (0.00)} & 0.258 (0.00) & 0.398 (0.00) & 0.275 (0.00) & 0.367 (0.00) & 0.250 (0.00) & 0.488 (0.00) & 0.381 (0.00) & 0.410 (0.00) & 0.282 (0.00) & 0.605 (0.01) & 0.330 (0.00) & 0.682 (0.02) & 0.415 (0.02) \\
 & 192 & 0.386 (0.01) & \textcolor{red}{0.256 (0.00)} & \textcolor{red}{0.376 (0.00)} & 0.268 (0.00) & 0.412 (0.00) & 0.280 (0.00) & 0.385 (0.00) & \underline{0.259 (0.00)} & 0.524 (0.01) & 0.403 (0.00) & 0.423 (0.00) & 0.287 (0.00) & 0.616 (0.00) & 0.333 (0.01) & 0.669 (0.03) & 0.411 (0.02) \\
 & 336 & 0.398 (0.01) & \textcolor{red}{0.262 (0.00)} & \textcolor{red}{0.389 (0.00)} & 0.274 (0.00) & 0.424 (0.00) & 0.287 (0.00) & 0.399 (0.00) & \underline{0.267 (0.00)} & 0.556 (0.01) & 0.423 (0.01) & 0.436 (0.000) & 0.296 (0.000) & 0.624 (0.01) & 0.335 (0.01) & 0.674 (0.03) & 0.415 (0.01) \\
 & 720 & 0.434 (0.00) & \textcolor{red}{0.284 (0.00)} & \textcolor{red}{0.426 (0.00)} & 0.293 (0.00) & 0.452 (0.00) & 0.305 (0.00) & 0.439 (0.01) & \underline{0.292 (0.01)} & 0.596 (0.01) & 0.444 (0.01) & 0.466 (0.000) & 0.315 (0.000) & 0.650 (0.01) & 0.346 (0.00) & 0.661 (0.02) & 0.406 (0.01) \\
\cline{1-18}
\multirow[t]{4}{*}{Weather} & 96 & \underline{0.148 (0.00)} & \textcolor{red}{0.194 (0.01)} & 0.163 (0.00) & 0.212 (0.00) & 0.152 (0.00) & \underline{0.206 (0.00)} & 0.151 (0.00) & \underline{0.199 (0.00)} & \textcolor{red}{0.146 (0.00)} & 0.220 (0.00) & 0.174 (0.00) & 0.235 (0.00) & 0.165 (0.00) & 0.223 (0.00) & 0.295 (0.01) & 0.370 (0.01) \\
 & 192 & \underline{0.195 (0.00)} & \textcolor{red}{0.239 (0.01)} & 0.207 (0.00) & 0.253 (0.00) & 0.197 (0.00) & \underline{0.247 (0.00)} & 0.196 (0.00) & \underline{0.242 (0.00)} & \textcolor{red}{0.192 (0.00)} & 0.267 (0.01) & 0.216 (0.00) & 0.274 (0.00) & 0.216 (0.00) & 0.267 (0.00) & 0.382 (0.04) & 0.431 (0.03) \\
 & 336 & 0.252 (0.01) & \textcolor{red}{0.282 (0.01)} & 0.256 (0.00) & \underline{0.291 (0.00)} & 0.246 (0.00) & \underline{0.285 (0.00)} & 0.248 (0.00) & \underline{0.283 (0.00)} & \textcolor{red}{0.240 (0.00)} & 0.304 (0.01) & 0.262 (0.00) & 0.314 (0.00) & 0.278 (0.01) & 0.309 (0.01) & 0.424 (0.04) & 0.445 (0.02) \\
 & 720 & 0.323 (0.00) & \textcolor{red}{0.331 (0.01)} & 0.330 (0.00) & 0.340 (0.00) & 0.327 (0.00) & \underline{0.338 (0.00)} & \underline{0.319 (0.00)} & \underline{0.335 (0.00)} & \textcolor{red}{0.311 (0.01)} & 0.359 (0.01) & 0.325 (0.00) & 0.365 (0.00) & 0.338 (0.00) & 0.349 (0.00) & 0.493 (0.06) & 0.476 (0.03) \\
\cline{1-18}
\bottomrule
\end{tabular}

%% file: tables/full_res_pems.tex
\begin{tabular}{llllllllllllllllll}
\toprule
 &  & \multicolumn{2}{c}{DARTS-TS} & \multicolumn{2}{c}{iTransformer} & \multicolumn{2}{c}{ModernTCN} & \multicolumn{2}{c}{PatchTST} & \multicolumn{2}{c}{TSMixer} & \multicolumn{2}{c}{DLinear} & \multicolumn{2}{c}{TimesNet} & \multicolumn{2}{c}{Autoformer} \\
 &  & MSE & MAE & MSE & MAE & MSE & MAE & MSE & MAE & MSE & MAE & MSE & MAE & MSE & MAE & MSE & MAE \\
\midrule
\multirow[t]{4}{*}{PEMS03} & 12 & \textcolor{red}{0.066 (0.00)} & \textcolor{red}{0.171 (0.00)} & 0.069 (0.00) & 0.175 (0.00) & 0.112 (0.00) & 0.221 (0.00) & 0.079 (0.00) & 0.187 (0.00) & 0.075 (0.00) & 0.187 (0.00) & 0.105 (0.00) & 0.220 (0.00) & 0.085 (0.00) & 0.192 (0.00) & 0.277 (0.06) & 0.387 (0.04) \\
 & 24 & \textcolor{red}{0.097 (0.00)} & \textcolor{red}{0.206 (0.00)} & \underline{0.098 (0.00)} & \underline{0.209 (0.00)} & 0.173 (0.00) & 0.281 (0.00) & 0.124 (0.00) & 0.235 (0.00) & 0.113 (0.00) & 0.238 (0.01) & 0.182 (0.00) & 0.296 (0.00) & 0.110 (0.00) & 0.216 (0.00) & 0.422 (0.05) & 0.466 (0.03) \\
 & 48 & \textcolor{red}{0.152 (0.01)} & \textcolor{red}{0.257 (0.01)} & \underline{0.448 (0.57)} & \underline{0.416 (0.28)} & 0.307 (0.00) & 0.395 (0.00) & 0.223 (0.00) & 0.319 (0.00) & 0.195 (0.02) & 0.320 (0.02) & 0.318 (0.00) & 0.410 (0.00) & \underline{0.168 (0.01)} & \underline{0.263 (0.00)} & 0.806 (0.08) & 0.679 (0.04) \\
 & 96 & \textcolor{red}{0.234 (0.02)} & \underline{0.331 (0.02)} & 1.215 (0.62) & 0.831 (0.25) & 1.041 (0.02) & 0.779 (0.01) & 0.368 (0.00) & 0.425 (0.00) & 0.266 (0.01) & 0.380 (0.01) & 0.450 (0.00) & 0.507 (0.00) & \underline{0.242 (0.01)} & \textcolor{red}{0.321 (0.00)} & 0.710 (0.15) & 0.634 (0.07) \\
\cline{1-18}
\multirow[t]{4}{*}{PEMS04} & 12 & \textcolor{red}{0.073 (0.00)} & \textcolor{red}{0.176 (0.00)} & 0.081 (0.00) & 0.188 (0.00) & 0.132 (0.00) & 0.245 (0.00) & 0.101 (0.00) & 0.209 (0.00) & 0.085 (0.00) & 0.195 (0.00) & 0.115 (0.00) & 0.228 (0.00) & 0.088 (0.00) & 0.197 (0.00) & 0.562 (0.06) & 0.577 (0.03) \\
 & 24 & \textcolor{red}{0.091 (0.00)} & \textcolor{red}{0.198 (0.00)} & 0.124 (0.00) & 0.232 (0.00) & 0.244 (0.00) & 0.338 (0.00) & 0.161 (0.00) & 0.267 (0.00) & 0.112 (0.01) & 0.228 (0.01) & 0.189 (0.00) & 0.299 (0.00) & 0.104 (0.00) & 0.216 (0.00) & 0.637 (0.10) & 0.617 (0.05) \\
 & 48 & \textcolor{red}{0.120 (0.00)} & \textcolor{red}{0.232 (0.00)} & 0.135 (0.00) & 0.248 (0.00) & 0.452 (0.00) & 0.482 (0.00) & 0.294 (0.00) & 0.369 (0.00) & 0.159 (0.01) & 0.278 (0.01) & 0.323 (0.00) & 0.407 (0.00) & 0.138 (0.00) & 0.252 (0.01) & 1.002 (0.10) & 0.775 (0.04) \\
 & 96 & \textcolor{red}{0.165 (0.00)} & \textcolor{red}{0.278 (0.00)} & 0.169 (0.00) & \underline{0.280 (0.00)} & 1.127 (0.00) & 0.818 (0.00) & 0.507 (0.00) & 0.505 (0.00) & 0.190 (0.01) & 0.313 (0.01) & 0.428 (0.00) & 0.484 (0.00) & 0.179 (0.00) & 0.291 (0.00) & 0.853 (0.24) & 0.708 (0.08) \\
\cline{1-18}
\multirow[t]{4}{*}{PEMS07} & 12 & \textcolor{red}{0.060 (0.00)} & \textcolor{red}{0.155 (0.00)} & 0.066 (0.00) & 0.164 (0.00) & 0.085 (0.00) & 0.196 (0.00) & 0.076 (0.00) & 0.180 (0.00) & 0.070 (0.00) & 0.177 (0.00) & 0.100 (0.00) & 0.215 (0.00) & 0.083 (0.00) & 0.183 (0.00) & 0.201 (0.02) & 0.330 (0.02) \\
 & 24 & \textcolor{red}{0.081 (0.00)} & \textcolor{red}{0.180 (0.00)} & 0.087 (0.00) & 0.190 (0.00) & 0.127 (0.00) & 0.245 (0.00) & 0.127 (0.00) & 0.234 (0.00) & 0.105 (0.01) & 0.221 (0.01) & 0.189 (0.00) & 0.302 (0.00) & 0.101 (0.00) & 0.204 (0.00) & 0.304 (0.04) & 0.402 (0.03) \\
 & 48 & \textcolor{red}{0.113 (0.01)} & \textcolor{red}{0.218 (0.01)} & 0.892 (0.12) & 0.764 (0.08) & 0.267 (0.01) & 0.380 (0.01) & 0.238 (0.00) & 0.325 (0.00) & 0.157 (0.01) & 0.265 (0.00) & 0.375 (0.00) & 0.436 (0.00) & 0.133 (0.00) & 0.236 (0.00) & 0.422 (0.13) & 0.472 (0.08) \\
 & 96 & \textcolor{red}{0.156 (0.02)} & \textcolor{red}{0.262 (0.01)} & 0.972 (0.19) & 0.789 (0.12) & 0.736 (0.02) & 0.673 (0.01) & 0.394 (0.00) & 0.432 (0.00) & 0.268 (0.02) & 0.342 (0.02) & 0.579 (0.00) & 0.540 (0.00) & \underline{0.211 (0.06)} & \underline{0.308 (0.06)} & 0.519 (0.10) & 0.546 (0.05) \\
\cline{1-18}
\multirow[t]{4}{*}{PEMS08} & 12 & \textcolor{red}{0.074 (0.00)} & \textcolor{red}{0.175 (0.00)} & 0.089 (0.00) & 0.193 (0.00) & 0.125 (0.00) & 0.239 (0.00) & 0.091 (0.00) & 0.195 (0.00) & 0.095 (0.00) & 0.203 (0.00) & 0.112 (0.00) & 0.223 (0.00) & 0.110 (0.00) & 0.208 (0.00) & 0.467 (0.07) & 0.503 (0.05) \\
 & 24 & \textcolor{red}{0.107 (0.01)} & \textcolor{red}{0.213 (0.01)} & 0.138 (0.00) & 0.243 (0.00) & 0.238 (0.00) & 0.336 (0.00) & 0.144 (0.00) & 0.247 (0.00) & 0.150 (0.01) & 0.257 (0.01) & 0.195 (0.00) & 0.299 (0.00) & 0.139 (0.00) & 0.234 (0.00) & 0.503 (0.07) & 0.512 (0.05) \\
 & 48 & \textcolor{red}{0.178 (0.02)} & \textcolor{red}{0.277 (0.02)} & 0.237 (0.01) & \underline{0.277 (0.01)} & 0.528 (0.00) & 0.534 (0.00) & 0.254 (0.00) & 0.332 (0.00) & 0.256 (0.01) & 0.344 (0.01) & 0.382 (0.00) & 0.431 (0.00) & \underline{0.194 (0.00)} & \underline{0.277 (0.00)} & 0.964 (0.23) & 0.729 (0.11) \\
 & 96 & \textcolor{red}{0.318 (0.04)} & \underline{0.360 (0.04)} & \underline{0.346 (0.07)} & \underline{0.363 (0.05)} & 1.150 (0.00) & 0.808 (0.00) & 0.435 (0.00) & 0.441 (0.00) & 0.399 (0.02) & 0.415 (0.02) & 0.634 (0.00) & 0.550 (0.01) & \underline{0.322 (0.01)} & \textcolor{red}{0.349 (0.01)} & 1.021 (0.14) & 0.763 (0.06) \\
\cline{1-18}
\bottomrule
\end{tabular}

%% file: tables/statistical_test_LTSF.tex

\begin{tabular}{llllllll}
\toprule
 &  &  & DARTS-TS & \multicolumn{2}{c}{Baseline} & statistic & p-value \\
 &  &  &  & Name & Value &  &  \\
\midrule
\multirow[t]{8}{*}{ECL} & \multirow[t]{2}{*}{96} & MSE & 0.129 (0.00) & PatchTST & 0.130 (0.00) & -2.114 & 0.067 \\
 &  & MAE & 0.217 (0.00) & PatchTST & 0.223 (0.00) & -3.954 & 0.004 \\
\cline{2-8}
 & \multirow[t]{2}{*}{192} & MSE & 0.147 (0.00) & PatchTST & 0.148 (0.00) & -3.985 & 0.004 \\
 &  & MAE & 0.234 (0.00) & PatchTST & 0.241 (0.00) & -7.983 & 0.000 \\
\cline{2-8}
 & \multirow[t]{2}{*}{336} & MSE & 0.164 (0.00) & PatchTST & 0.165 (0.00) & -1.445 & 0.186 \\
 &  & MAE & 0.253 (0.00) & PatchTST & 0.259 (0.00) & -2.992 & 0.017 \\
\cline{2-8}
 & \multirow[t]{2}{*}{720} & MSE & 0.186 (0.00) & PatchTST & 0.202 (0.00) & -8.918 & 0.000 \\
 &  & MAE & 0.275 (0.00) & PatchTST & 0.292 (0.00) & -15.057 & 0.000 \\
\cline{1-8} \cline{2-8}
\multirow[t]{8}{*}{ETTh1} & \multirow[t]{2}{*}{96} & MSE & 0.365 (0.00) & ModernTCN & 0.369 (0.00) & -1.735 & 0.121 \\
 &  & MAE & 0.385 (0.00) & ModernTCN & 0.394 (0.00) & -5.077 & 0.001 \\
\cline{2-8}
 & \multirow[t]{2}{*}{192} & MSE & 0.405 (0.00) & ModernTCN & 0.407 (0.00) & -1.129 & 0.291 \\
 &  & MAE & 0.411 (0.00) & ModernTCN & 0.415 (0.00) & -2.168 & 0.062 \\
\cline{2-8}
 & \multirow[t]{2}{*}{336} & MSE & 0.437 (0.01) & ModernTCN & 0.392 (0.00) & 11.610 & 0.000 \\
 &  & MAE & 0.433 (0.01) & ModernTCN & 0.413 (0.00) & 6.234 & 0.000 \\
\cline{2-8}
 & \multirow[t]{2}{*}{720} & MSE & 0.448 (0.01) & PatchTST & 0.444 (0.00) & 0.920 & 0.385 \\
 &  & MAE & 0.462 (0.00) & ModernTCN & 0.461 (0.00) & 0.331 & 0.749 \\
\cline{1-8} \cline{2-8}
\multirow[t]{8}{*}{ETTh2} & \multirow[t]{2}{*}{96} & MSE & 0.276 (0.00) & ModernTCN & 0.264 (0.00) & 25.684 & 0.000 \\
 &  & MAE & 0.332 (0.00) & ModernTCN & 0.333 (0.00) & -2.560 & 0.034 \\
\cline{2-8}
 & \multirow[t]{2}{*}{192} & MSE & 0.344 (0.00) & ModernTCN & 0.322 (0.00) & 18.238 & 0.000 \\
 &  & MAE & 0.377 (0.00) & ModernTCN & 0.377 (0.00) & 1.436 & 0.189 \\
\cline{2-8}
 & \multirow[t]{2}{*}{336} & MSE & 0.377 (0.01) & ModernTCN & 0.315 (0.00) & 22.508 & 0.000 \\
 &  & MAE & 0.406 (0.00) & ModernTCN & 0.377 (0.00) & 23.442 & 0.000 \\
\cline{2-8}
 & \multirow[t]{2}{*}{720} & MSE & 0.406 (0.01) & PatchTST & 0.378 (0.00) & 10.461 & 0.000 \\
 &  & MAE & 0.435 (0.00) & PatchTST & 0.420 (0.00) & 7.311 & 0.000 \\
\cline{1-8} \cline{2-8}
\multirow[t]{8}{*}{ETTm1} & \multirow[t]{2}{*}{96} & MSE & 0.283 (0.00) & PatchTST & 0.290 (0.00) & -2.776 & 0.024 \\
 &  & MAE & 0.330 (0.00) & PatchTST & 0.341 (0.00) & -6.752 & 0.000 \\
\cline{2-8}
 & \multirow[t]{2}{*}{192} & MSE & 0.320 (0.00) & PatchTST & 0.333 (0.00) & -4.973 & 0.001 \\
 &  & MAE & 0.354 (0.00) & DLinear & 0.368 (0.00) & -9.211 & 0.000 \\
\cline{2-8}
 & \multirow[t]{2}{*}{336} & MSE & 0.357 (0.00) & PatchTST & 0.367 (0.00) & -5.200 & 0.001 \\
 &  & MAE & 0.377 (0.00) & PatchTST & 0.391 (0.00) & -20.458 & 0.000 \\
\cline{2-8}
 & \multirow[t]{2}{*}{720} & MSE & 0.418 (0.01) & PatchTST & 0.417 (0.00) & 0.267 & 0.797 \\
 &  & MAE & 0.412 (0.00) & ModernTCN & 0.421 (0.00) & -9.198 & 0.000 \\
\cline{1-8} \cline{2-8}
\bottomrule
\end{tabular}
\begin{tabular}{llllllll}
\toprule
 &  &  & DARTS-TS & \multicolumn{2}{c}{Baseline} & statistic & p-value \\
 &  &  &  & Name & Value &  &  \\
\midrule
\multirow[t]{8}{*}{ETTm2} & \multirow[t]{2}{*}{96} & MSE & 0.162 (0.00) & PatchTST & 0.165 (0.00) & -3.616 & 0.007 \\
 &  & MAE & 0.244 (0.00) & PatchTST & 0.254 (0.00) & -14.964 & 0.000 \\
\cline{2-8}
 & \multirow[t]{2}{*}{192} & MSE & 0.223 (0.00) & PatchTST & 0.222 (0.00) & 0.735 & 0.484 \\
 &  & MAE & 0.285 (0.00) & PatchTST & 0.293 (0.00) & -12.254 & 0.000 \\
\cline{2-8}
 & \multirow[t]{2}{*}{336} & MSE & 0.274 (0.00) & PatchTST & 0.277 (0.00) & -2.276 & 0.052 \\
 &  & MAE & 0.320 (0.00) & PatchTST & 0.329 (0.00) & -32.523 & 0.000 \\
\cline{2-8}
 & \multirow[t]{2}{*}{720} & MSE & 0.354 (0.00) & PatchTST & 0.364 (0.00) & -7.163 & 0.000 \\
 &  & MAE & 0.373 (0.00) & PatchTST & 0.383 (0.00) & -10.913 & 0.000 \\
\cline{1-8} \cline{2-8}
\multirow[t]{8}{*}{Exchange} & \multirow[t]{2}{*}{96} & MSE & 0.088 (0.00) & DLinear & 0.084 (0.00) & 2.040 & 0.076 \\
 &  & MAE & 0.210 (0.00) & DLinear & 0.203 (0.00) & 3.046 & 0.016 \\
\cline{2-8}
 & \multirow[t]{2}{*}{192} & MSE & 0.186 (0.00) & DLinear & 0.164 (0.01) & 6.302 & 0.000 \\
 &  & MAE & 0.308 (0.00) & DLinear & 0.293 (0.00) & 5.449 & 0.001 \\
\cline{2-8}
 & \multirow[t]{2}{*}{336} & MSE & 0.350 (0.01) & PatchTST & 0.350 (0.00) & 0.010 & 0.992 \\
 &  & MAE & 0.428 (0.01) & PatchTST & 0.431 (0.00) & -0.724 & 0.490 \\
\cline{2-8}
 & \multirow[t]{2}{*}{720} & MSE & 0.888 (0.03) & TSMixer & 0.671 (0.11) & 3.802 & 0.005 \\
 &  & MAE & 0.689 (0.01) & TSMixer & 0.653 (0.04) & 1.634 & 0.141 \\
\cline{1-8} \cline{2-8}
\multirow[t]{8}{*}{Traffic} & \multirow[t]{2}{*}{96} & MSE & 0.358 (0.00) & iTransformer & 0.356 (0.00) & 1.327 & 0.221 \\
 &  & MAE & 0.240 (0.01) & PatchTST & 0.250 (0.00) & -3.321 & 0.011 \\
\cline{2-8}
 & \multirow[t]{2}{*}{192} & MSE & 0.386 (0.01) & iTransformer & 0.376 (0.00) & 2.652 & 0.029 \\
 &  & MAE & 0.256 (0.00) & PatchTST & 0.259 (0.00) & -1.174 & 0.274 \\
\cline{2-8}
 & \multirow[t]{2}{*}{336} & MSE & 0.398 (0.01) & iTransformer & 0.389 (0.00) & 3.337 & 0.010 \\
 &  & MAE & 0.262 (0.00) & PatchTST & 0.267 (0.00) & -2.009 & 0.079 \\
\cline{2-8}
 & \multirow[t]{2}{*}{720} & MSE & 0.434 (0.00) & iTransformer & 0.426 (0.00) & 1.963 & 0.107 \\
 &  & MAE & 0.284 (0.00) & PatchTST & 0.292 (0.01) & -1.488 & 0.175 \\
\cline{1-8} \cline{2-8}
\multirow[t]{8}{*}{Weather} & \multirow[t]{2}{*}{96} & MSE & 0.148 (0.00) & TSMixer & 0.146 (0.00) & 1.586 & 0.151 \\
 &  & MAE & 0.194 (0.01) & PatchTST & 0.199 (0.00) & -0.880 & 0.404 \\
\cline{2-8}
 & \multirow[t]{2}{*}{192} & MSE & 0.195 (0.00) & TSMixer & 0.192 (0.00) & 1.331 & 0.220 \\
 &  & MAE & 0.239 (0.01) & PatchTST & 0.242 (0.00) & -0.482 & 0.643 \\
\cline{2-8}
 & \multirow[t]{2}{*}{336} & MSE & 0.252 (0.01) & TSMixer & 0.240 (0.00) & 2.741 & 0.025 \\
 &  & MAE & 0.282 (0.01) & PatchTST & 0.283 (0.00) & -0.102 & 0.921 \\
\cline{2-8}
 & \multirow[t]{2}{*}{720} & MSE & 0.323 (0.00) & TSMixer & 0.311 (0.01) & 2.937 & 0.019 \\
 &  & MAE & 0.331 (0.01) & PatchTST & 0.335 (0.00) & -0.941 & 0.374 \\
\cline{1-8} \cline{2-8}
\bottomrule
\end{tabular}

%% file: tables/statisitical_test_pems.tex

\begin{tabular}{llllllll}
\toprule
 &  &  & DARTS-TS & \multicolumn{2}{c}{Baseline} & statistic & p-value \\
 &  &  &  & Name & Value &  &  \\
\midrule
\multirow[t]{8}{*}{PEMS03} & \multirow[t]{2}{*}{12} & MSE & 0.066 (0.00) & iTransformer & 0.069 (0.00) & -2.449 & 0.040 \\
 &  & MAE & 0.171 (0.00) & iTransformer & 0.175 (0.00) & -2.651 & 0.029 \\
\cline{2-8}
 & \multirow[t]{2}{*}{24} & MSE & 0.097 (0.00) & iTransformer & 0.098 (0.00) & -0.679 & 0.516 \\
 &  & MAE & 0.206 (0.00) & iTransformer & 0.209 (0.00) & -1.589 & 0.151 \\
\cline{2-8}
 & \multirow[t]{2}{*}{48} & MSE & 0.152 (0.01) & TimesNet & 0.168 (0.01) & -2.215 & 0.058 \\
 &  & MAE & 0.257 (0.01) & TimesNet & 0.263 (0.00) & -1.166 & 0.277 \\
\cline{2-8}
 & \multirow[t]{2}{*}{96} & MSE & 0.234 (0.02) & TimesNet & 0.242 (0.01) & -0.588 & 0.573 \\
 &  & MAE & 0.331 (0.02) & TimesNet & 0.321 (0.00) & 1.170 & 0.276 \\
\cline{1-8} \cline{2-8}
\multirow[t]{8}{*}{PEMS04} & \multirow[t]{2}{*}{12} & MSE & 0.073 (0.00) & iTransformer & 0.081 (0.00) & -15.978 & 0.000 \\
 &  & MAE & 0.176 (0.00) & iTransformer & 0.188 (0.00) & -16.158 & 0.000 \\
\cline{2-8}
 & \multirow[t]{2}{*}{24} & MSE & 0.091 (0.00) & TimesNet & 0.104 (0.00) & -8.493 & 0.000 \\
 &  & MAE & 0.198 (0.00) & TimesNet & 0.216 (0.00) & -7.884 & 0.000 \\
\cline{2-8}
 & \multirow[t]{2}{*}{48} & MSE & 0.120 (0.00) & iTransformer & 0.135 (0.00) & -8.544 & 0.000 \\
 &  & MAE & 0.232 (0.00) & iTransformer & 0.248 (0.00) & -7.388 & 0.000 \\
\cline{2-8}
 & \multirow[t]{2}{*}{96} & MSE & 0.165 (0.00) & iTransformer & 0.169 (0.00) & -2.519 & 0.036 \\
 &  & MAE & 0.278 (0.00) & iTransformer & 0.280 (0.00) & -1.439 & 0.188 \\
\cline{1-8} \cline{2-8}
\bottomrule
\end{tabular}
\begin{tabular}{llllllll}
\toprule
 &  &  & DARTS-TS & \multicolumn{2}{c}{Baseline} & statistic & p-value \\
 &  &  &  & Name & Value &  &  \\
\midrule
     \multirow[t]{8}{*}{PEMS07} & \multirow[t]{2}{*}{12} & MSE & 0.060 (0.00) & iTransformer & 0.066 (0.00) & -6.855 & 0.000 \\
 &  & MAE & 0.155 (0.00) & iTransformer & 0.164 (0.00) & -4.497 & 0.002 \\
\cline{2-8}
 & \multirow[t]{2}{*}{24} & MSE & 0.081 (0.00) & iTransformer & 0.087 (0.00) & -4.173 & 0.003 \\
 &  & MAE & 0.180 (0.00) & iTransformer & 0.190 (0.00) & -4.539 & 0.002 \\
\cline{2-8}
 & \multirow[t]{2}{*}{48} & MSE & 0.113 (0.01) & TimesNet & 0.133 (0.00) & -3.977 & 0.004 \\
 &  & MAE & 0.218 (0.01) & TimesNet & 0.236 (0.00) & -3.299 & 0.011 \\
\cline{2-8}
 & \multirow[t]{2}{*}{96} & MSE & 0.156 (0.02) & TimesNet & 0.211 (0.06) & -1.799 & 0.110 \\
 &  & MAE & 0.262 (0.01) & TimesNet & 0.308 (0.06) & -1.555 & 0.158 \\
\cline{1-8} \cline{2-8}
\multirow[t]{8}{*}{PEMS08} & \multirow[t]{2}{*}{12} & MSE & 0.074 (0.00) & iTransformer & 0.089 (0.00) & -11.044 & 0.000 \\
 &  & MAE & 0.175 (0.00) & iTransformer & 0.193 (0.00) & -7.653 & 0.000 \\
\cline{2-8}
 & \multirow[t]{2}{*}{24} & MSE & 0.107 (0.01) & iTransformer & 0.138 (0.00) & -9.285 & 0.000 \\
 &  & MAE & 0.213 (0.01) & TimesNet & 0.234 (0.00) & -4.376 & 0.002 \\
\cline{2-8}
 & \multirow[t]{2}{*}{48} & MSE & 0.178 (0.02) & TimesNet & 0.194 (0.00) & -1.710 & 0.126 \\
 &  & MAE & 0.277 (0.02) & iTransformer & 0.277 (0.01) & -0.016 & 0.988 \\
\cline{2-8}
 & \multirow[t]{2}{*}{96} & MSE & 0.318 (0.04) & TimesNet & 0.322 (0.01) & -0.160 & 0.877 \\
 &  & MAE & 0.360 (0.04) & TimesNet & 0.349 (0.01) & 0.643 & 0.538 \\
\cline{1-8} \cline{2-8}
\bottomrule
\end{tabular}

%% file: tables/llm_based/res_with_llm.tex

\begin{tabular}{llllllll}
\toprule
 &  & \multicolumn{2}{c}{DARTS-TS} & \multicolumn{2}{c}{TimesFM} & \multicolumn{2}{c}{Moirai-MoE} \\
 &  & MSE & MAE & MSE & MAE & MSE & MAE \\
\midrule
\multirow[t]{4}{*}{ECL} & 96 & 0.129 & 0.217 & \textcolor{red}{0.126} & \textcolor{red}{0.217} & 0.213 & 0.280 \\
 & 192 & 0.147 & \textcolor{red}{0.234} & \textcolor{red}{0.146} & 0.236 & 0.228 & 0.296 \\
 & 336 & \textcolor{red}{0.164} & \textcolor{red}{0.253} & 0.168 & 0.257 & NaN & NaN \\
 & 720 & \textcolor{red}{0.186} & \textcolor{red}{0.275} & 0.214 & 0.295 & NaN & NaN \\
\cline{1-8}
\multirow[t]{4}{*}{ETTh1} & 96 & \textcolor{red}{0.365} & \textcolor{red}{0.385} & 0.453 & 0.412 & 0.472 & 0.415 \\
 & 192 & \textcolor{red}{0.405} & \textcolor{red}{0.411} & 0.490 & 0.439 & 0.544 & 0.450 \\
 & 336 & \textcolor{red}{0.437} & \textcolor{red}{0.433} & 0.518 & 0.457 & 0.612 & 0.479 \\
 & 720 & \textcolor{red}{0.448} & \textcolor{red}{0.462} & 0.512 & 0.480 & 0.621 & 0.508 \\
\cline{1-8}
\multirow[t]{4}{*}{ETTh2} & 96 & \textcolor{red}{0.276} & \textcolor{red}{0.332} & 0.326 & 0.353 & 0.345 & 0.356 \\
 & 192 & \textcolor{red}{0.344} & \textcolor{red}{0.377} & 0.392 & 0.398 & 0.413 & 0.404 \\
 & 336 & \textcolor{red}{0.377} & \textcolor{red}{0.406} & 0.422 & 0.423 & 0.429 & 0.426 \\
 & 720 & \textcolor{red}{0.406} & \textcolor{red}{0.435} & 0.462 & 0.462 & 0.441 & 0.439 \\
\cline{1-8}
\multirow[t]{4}{*}{ETTm1} & 96 & \textcolor{red}{0.283} & \textcolor{red}{0.330} & 0.363 & 0.371 & 1.097 & 0.606 \\
 & 192 & \textcolor{red}{0.320} & \textcolor{red}{0.354} & 0.427 & 0.411 & 1.009 & 0.600 \\
 & 336 & \textcolor{red}{0.357} & \textcolor{red}{0.377} & 0.482 & 0.447 & 0.940 & 0.592 \\
 & 720 & \textcolor{red}{0.418} & \textcolor{red}{0.412} & 0.544 & 0.487 & 0.906 & 0.594 \\
\cline{1-8}
\bottomrule
\end{tabular}
\begin{tabular}{llllllll}
\toprule
 &  & \multicolumn{2}{c}{DARTS-TS} & \multicolumn{2}{c}{TimesFM} & \multicolumn{2}{c}{Moirai-MoE} \\
 &  & MSE & MAE & MSE & MAE & MSE & MAE \\
\midrule
    \multirow[t]{4}{*}{ETTm2} & 96 & \textcolor{red}{0.162} & \textcolor{red}{0.244} & 0.208 & 0.273 & 0.256 & 0.315 \\
 & 192 & \textcolor{red}{0.223} & \textcolor{red}{0.285} & 0.287 & 0.323 & 0.319 & 0.353 \\
 & 336 & \textcolor{red}{0.274} & \textcolor{red}{0.320} & 0.361 & 0.369 & 0.368 & 0.382 \\
 & 720 & \textcolor{red}{0.354} & \textcolor{red}{0.373} & 0.460 & 0.434 & 0.464 & 0.432 \\
\cline{1-8}
\multirow[t]{4}{*}{Exchange} & 96 & 0.088 & 0.210 & 0.109 & 0.231 & \textcolor{red}{0.083} & \textcolor{red}{0.201} \\
 & 192 & 0.186 & 0.308 & 0.228 & 0.340 & \textcolor{red}{0.184} & \textcolor{red}{0.302} \\
 & 336 & 0.350 & 0.428 & 0.412 & 0.465 & \textcolor{red}{0.350} & \textcolor{red}{0.424} \\
 & 720 & \textcolor{red}{0.888} & \textcolor{red}{0.689} & 1.035 & 0.754 & 0.897 & 0.715 \\
\cline{1-8}
\multirow[t]{4}{*}{Traffic} & 96 & 0.358 & 0.240 & \textcolor{red}{0.340} & \textcolor{red}{0.225} & NaN & NaN \\
 & 192 & 0.386 & 0.256 & \textcolor{red}{0.368} & \textcolor{red}{0.241} & NaN & NaN \\
 & 336 & 0.398 & 0.262 & \textcolor{red}{0.391} & \textcolor{red}{0.254} & NaN & NaN \\
 & 720 & 0.434 & 0.284 & \textcolor{red}{0.432} & \textcolor{red}{0.279} & NaN & NaN \\
\cline{1-8}
\multirow[t]{4}{*}{Weather} & 96 & 0.148 & 0.194 & \textcolor{red}{0.136} & \textcolor{red}{0.169} & 0.244 & 0.250 \\
 & 192 & 0.195 & 0.239 & \textcolor{red}{0.177} & \textcolor{red}{0.209} & 0.291 & 0.289 \\
 & 336 & 0.252 & 0.282 & \textcolor{red}{0.236} & \textcolor{red}{0.257} & 0.351 & 0.330 \\
 & 720 & \textcolor{red}{0.323} & \textcolor{red}{0.331} & 0.340 & 0.337 & 0.418 & 0.376 \\
\cline{1-8}
\bottomrule
\end{tabular}

%% file: tables/llm_based/res_with_llm_pems.tex

\begin{tabular}{llllllll}
\toprule
 &  & \multicolumn{2}{c}{DARTS-TS} & \multicolumn{2}{c}{TimesFM} & \multicolumn{2}{c}{Moirai-MoE} \\
 &  & MSE & MAE & MSE & MAE & MSE & MAE \\
\midrule
\multirow[t]{4}{*}{PEMS03} & 12 & \textcolor{red}{0.066} & \textcolor{red}{0.171} & 0.158 & 0.266 & 0.124 & 0.232 \\
 & 24 & \textcolor{red}{0.097} & \textcolor{red}{0.206} & 0.326 & 0.391 & 0.250 & 0.330 \\
 & 48 & \textcolor{red}{0.152} & \textcolor{red}{0.257} & 0.767 & 0.632 & 0.589 & 0.526 \\
 & 96 & \textcolor{red}{0.234} & \textcolor{red}{0.331} & 1.673 & 1.008 & 1.378 & 0.867 \\
\cline{1-8}
\multirow[t]{4}{*}{PEMS04} & 12 & \textcolor{red}{0.073} & \textcolor{red}{0.176} & 0.176 & 0.282 & 0.134 & 0.241 \\
 & 24 & \textcolor{red}{0.091} & \textcolor{red}{0.198} & 0.363 & 0.416 & 0.261 & 0.340 \\
 & 48 & \textcolor{red}{0.120} & \textcolor{red}{0.232} & 0.854 & 0.670 & 0.616 & 0.543 \\
 & 96 & \textcolor{red}{0.165} & \textcolor{red}{0.278} & 1.824 & 1.055 & 1.476 & 0.903 \\
\cline{1-8}
\bottomrule
\end{tabular}
\begin{tabular}{llllllll}
\toprule
 &  & \multicolumn{2}{c}{DARTS-TS} & \multicolumn{2}{c}{TimesFM} & \multicolumn{2}{c}{Moirai-MoE} \\
 &  & MSE & MAE & MSE & MAE & MSE & MAE \\
\midrule
     \multirow[t]{4}{*}{PEMS07} & 12 & \textcolor{red}{0.060} & \textcolor{red}{0.155} & 0.149 & 0.260 & 0.110 & 0.218 \\
 & 24 & \textcolor{red}{0.081} & \textcolor{red}{0.180} & 0.340 & 0.397 & 0.232 & 0.316 \\
 & 48 & \textcolor{red}{0.113} & \textcolor{red}{0.218} & 0.824 & 0.647 & 0.593 & 0.521 \\
 & 96 & \textcolor{red}{0.156} & \textcolor{red}{0.262} & 1.701 & 1.009 & 1.445 & 0.871 \\
\cline{1-8}
\multirow[t]{4}{*}{PEMS08} & 12 & \textcolor{red}{0.074} & \textcolor{red}{0.175} & 0.160 & 0.270 & 0.128 & 0.234 \\
 & 24 & \textcolor{red}{0.107} & \textcolor{red}{0.213} & 0.331 & 0.397 & 0.247 & 0.331 \\
 & 48 & \textcolor{red}{0.178} & \textcolor{red}{0.277} & 0.814 & 0.649 & 0.597 & 0.536 \\
 & 96 & \textcolor{red}{0.318} & \textcolor{red}{0.360} & 1.855 & 1.050 & 1.500 & 0.902 \\
\cline{1-8}
\bottomrule
\end{tabular}

%% file: tables/res_ablation_components.tex
\begin{tabular}{llllllllllllllllll}
\toprule
 &  & \multicolumn{2}{c}{DARTS-TS} & \multicolumn{2}{c}{DARTS-TS CV} & \multicolumn{2}{c}{Parallel} & \multicolumn{2}{c}{Seq Only} & \multicolumn{4}{c}{Flat Only} & \multicolumn{2}{c}{Concat Seq First} & \multicolumn{2}{c}{No Weights} \\
 &  & MSE & MAE & MSE & MAE & MSE & MAE & MSE & MAE & MSE & MAE & MSE & MAE & MSE & MAE & MSE & MAE \\
\midrule
ECL & 96 & 0.129 (0.00) & 0.217 (0.0) & 0.129 (0.00) & 0.220 (0.00) & 0.128 (0.00) & 0.219 (0.00) & 0.206 (0.05) & 0.305 (0.04) & 0.128 (0.00) & 0.221 (0.00) & 0.128 (0.00) & 0.221 (0.00) & 0.139 (0.01) & 0.234 (0.02) & 0.139 (0.01) & 0.234 (0.02) \\
\cline{1-18}
ETTh1 & 96 & 0.365 (0.00) & 0.385 (0.0) & 0.402 (0.02) & 0.420 (0.02) & 0.371 (0.01) & 0.392 (0.01) & 0.466 (0.04) & 0.463 (0.02) & 0.383 (0.00) & 0.406 (0.00) & 0.383 (0.00) & 0.406 (0.00) & 0.364 (0.00) & 0.385 (0.00) & 0.366 (0.00) & 0.385 (0.00) \\
\cline{1-18}
ETTh2 & 96 & 0.276 (0.00) & 0.332 (0.0) & 0.284 (0.00) & 0.339 (0.00) & 0.282 (0.00) & 0.341 (0.00) & 0.382 (0.02) & 0.421 (0.01) & 0.312 (0.00) & 0.361 (0.00) & 0.312 (0.00) & 0.361 (0.00) & 0.283 (0.00) & 0.338 (0.00) & 0.279 (0.00) & 0.338 (0.01) \\
\cline{1-18}
ETTm1 & 96 & 0.283 (0.00) & 0.330 (0.0) & 0.291 (0.01) & 0.335 (0.01) & 0.287 (0.00) & 0.338 (0.00) & 0.366 (0.02) & 0.402 (0.02) & 0.295 (0.00) & 0.345 (0.00) & 0.295 (0.00) & 0.345 (0.00) & 0.283 (0.00) & 0.331 (0.00) & 0.286 (0.00) & 0.334 (0.01) \\
\cline{1-18}
ETTm2 & 96 & 0.162 (0.00) & 0.244 (0.0) & 0.163 (0.00) & 0.245 (0.00) & 0.166 (0.00) & 0.249 (0.00) & 0.221 (0.01) & 0.301 (0.01) & 0.173 (0.00) & 0.255 (0.00) & 0.173 (0.00) & 0.255 (0.00) & 0.164 (0.00) & 0.246 (0.00) & 0.162 (0.00) & 0.246 (0.00) \\
\cline{1-18}
Exchange & 96 & 0.088 (0.00) & 0.210 (0.0) & 0.088 (0.00) & 0.209 (0.00) & 0.098 (0.00) & 0.220 (0.00) & 0.227 (0.03) & 0.347 (0.02) & 0.107 (0.01) & 0.234 (0.01) & 0.107 (0.01) & 0.234 (0.01) & 0.091 (0.00) & 0.213 (0.00) & 0.092 (0.00) & 0.214 (0.00) \\
\cline{1-18}
Traffic & 96 & 0.358 (0.00) & 0.240 (0.0) & 0.364 (0.01) & 0.246 (0.01) & 0.370 (0.02) & 0.244 (0.01) & 0.572 (0.02) & 0.312 (0.01) & 0.359 (0.00) & 0.249 (0.00) & 0.359 (0.00) & 0.249 (0.00) & 0.417 (0.01) & 0.275 (0.02) & 0.452 (0.00) & 0.300 (0.01) \\
\cline{1-18}
Weather & 96 & 0.148 (0.00) & 0.194 (0.0) & 0.149 (0.00) & 0.195 (0.01) & 0.150 (0.00) & 0.195 (0.01) & 0.191 (0.00) & 0.244 (0.00) & 0.151 (0.00) & 0.201 (0.00) & 0.151 (0.00) & 0.201 (0.00) & 0.149 (0.00) & 0.188 (0.00) & 0.157 (0.00) & 0.209 (0.00) \\
\cline{1-18}
\bottomrule
\end{tabular}

%% file: tables/res_ablation_ops.tex
\begin{tabular}{lllllllllllllllllllllll}
\toprule
 & \multicolumn{2}{c}{Full Ops} & \multicolumn{2}{c}{MLP} & \multicolumn{2}{c}{N-BEATS G} & \multicolumn{2}{c}{N-BEATS S} & \multicolumn{2}{c}{N-BEATS T} & \multicolumn{2}{c}{GRU} & \multicolumn{2}{c}{LSTM} & \multicolumn{2}{c}{MLP-Mixer} & \multicolumn{2}{c}{Sep. TCN} & \multicolumn{2}{c}{TCN} & \multicolumn{2}{c}{Transformer} \\
 & MSE & MAE & MSE & MAE & MSE & MAE & MSE & MAE & MSE & MAE & MSE & MAE & MSE & MAE & MSE & MAE & MSE & MAE & MSE & MAE & MSE & MAE \\
\midrule
ECL & 0.129 & 0.217 & 0.131 & 0.223 & 0.130 & 0.222 & 0.133 & 0.227 & 0.136 & 0.233 & 0.132 & 0.224 & 0.131 & 0.221 & 0.129 & 0.220 & 0.128 & 0.216 & 0.131 & 0.220 & 0.129 & 0.219 \\
ETTh1 & 0.365 & 0.385 & 0.390 & 0.408 & 0.385 & 0.401 & 0.378 & 0.398 & 0.378 & 0.399 & 0.372 & 0.393 & 0.393 & 0.409 & 0.371 & 0.391 & 0.381 & 0.398 & 0.366 & 0.387 & 0.384 & 0.403 \\
ETTh2 & 0.276 & 0.332 & 0.289 & 0.341 & 0.286 & 0.339 & 0.286 & 0.338 & 0.291 & 0.347 & 0.292 & 0.345 & 0.292 & 0.346 & 0.286 & 0.340 & 0.286 & 0.340 & 0.287 & 0.340 & 0.287 & 0.341 \\
ETTm1 & 0.283 & 0.330 & 0.284 & 0.333 & 0.278 & 0.327 & 0.278 & 0.327 & 0.284 & 0.336 & 0.280 & 0.328 & 0.285 & 0.332 & 0.285 & 0.333 & 0.284 & 0.332 & 0.282 & 0.330 & 0.283 & 0.329 \\
ETTm2 & 0.162 & 0.244 & 0.163 & 0.245 & 0.161 & 0.245 & 0.163 & 0.246 & 0.166 & 0.250 & 0.163 & 0.248 & 0.162 & 0.245 & 0.162 & 0.245 & 0.162 & 0.245 & 0.167 & 0.250 & 0.161 & 0.244 \\
Traffic & 0.358 & 0.240 & 0.379 & 0.251 & 0.381 & 0.258 & 0.378 & 0.255 & 0.402 & 0.278 & 0.381 & 0.244 & 0.371 & 0.245 & 0.364 & 0.238 & 0.367 & 0.235 & 0.377 & 0.244 & 0.369 & 0.247 \\
Exchange Rate & 0.088 & 0.210 & 0.095 & 0.218 & 0.093 & 0.215 & 0.094 & 0.218 & 0.105 & 0.228 & 0.089 & 0.210 & 0.090 & 0.213 & 0.089 & 0.211 & 0.091 & 0.213 & 0.092 & 0.214 & 0.089 & 0.211 \\
Weather & 0.148 & 0.194 & 0.150 & 0.198 & 0.153 & 0.207 & 0.154 & 0.208 & 0.150 & 0.198 & 0.152 & 0.195 & 0.152 & 0.197 & 0.149 & 0.192 & 0.155 & 0.201 & 0.148 & 0.192 & 0.152 & 0.195 \\
\bottomrule
\end{tabular}

%% file: tables/res_ablation_revin.tex
\begin{tabular}{llllllllll}
\toprule
 &  & ECL & ETTh1 & ETTh2 & ETTm1 & ETTm2 & Traffic & Exchange Rate & Weather \\
\midrule
\multirow[t]{2}{*}{w/ RevIN} & MSE & 0.129(0.001) & 0.365(0.004) & 0.276(0.001) & 0.283(0.005) & 0.162(0.001) & 0.358(0.003) & 0.088(0.002) & 0.148(0.003) \\
 & MAE & 0.217(0.003) & 0.385(0.003) & 0.332(0.000) & 0.330(0.003) & 0.244(0.001) & 0.240(0.006) & 0.210(0.002) & 0.194(0.011) \\
\cline{1-10}
\multirow[t]{2}{*}{w/o RevIN} & MSE & 0.130(0.004) & 0.400(0.022) & 0.287(0.004) & 0.288(0.002) & 0.175(0.020) & 0.375(0.009) & 0.093(0.004) & 0.147(0.003) \\
 & MAE & 0.225(0.009) & 0.421(0.018) & 0.341(0.002) & 0.334(0.002) & 0.267(0.030) & 0.246(0.008) & 0.219(0.003) & 0.193(0.007) \\
\cline{1-10}
\bottomrule
\end{tabular}

%% file: main.bbl
\begin{thebibliography}{101}
\providecommand{\natexlab}[1]{#1}
\providecommand{\url}[1]{\texttt{#1}}
\expandafter\ifx\csname urlstyle\endcsname\relax
  \providecommand{\doi}[1]{doi: #1}\else
  \providecommand{\doi}{doi: \begingroup \urlstyle{rm}\Url}\fi

\bibitem[Abdelfattah et~al.(2021)Abdelfattah, Mehrotra, Dudziak, and Lane]{abdelfattah-iclr22a}
M.~Abdelfattah, A.~Mehrotra, L.~Dudziak, and N.~Lane.
\newblock Zero-cost proxies for lightweight {NAS}.
\newblock In \emph{The Ninth International Conference on Learning Representations ({ICLR}'21)} \citet{iclr21}.

\bibitem[Alexandrov et~al.(2020)Alexandrov, Benidis, Bohlke-Schneider, Flunkert, Gasthaus, Januschowski, Maddix, Rangapuram, Salinas, Schulz, Stella, Türkmen, and Wang]{alexandrov-jmlr20a}
A.~Alexandrov, K.~Benidis, M.~Bohlke-Schneider, V.~Flunkert, J.~Gasthaus, T.~Januschowski, D.~Maddix, S.~Rangapuram, D.~Salinas, J.~Schulz, L.~Stella, A.~Türkmen, and Y.~Wang.
\newblock Gluonts: Probabilistic and neural time series modeling in python.
\newblock \emph{Journal of Machine Learning Research}, 21:\penalty0 116:1--116:6, 2020.

\bibitem[Ansari et~al.(2024)Ansari, Stella, T{\"{u}}rkmen, Zhang, Mercado, Shen, Shchur, Rangapuram, Pineda{-}Arango, Kapoor, Zschiegner, Maddix, Mahoney, Torkkola, Wilson, Bohlke{-}Schneider, and Wang]{ansari-arxiv24a}
A.~Ansari, L.~Stella, A.~T{\"{u}}rkmen, X.~Zhang, P.~Mercado, H.~Shen, O.~Shchur, S.~Rangapuram, S.~Pineda{-}Arango, S.~Kapoor, J.~Zschiegner, D.~Maddix, M.~Mahoney, K.~Torkkola, A.~Wilson, M.~Bohlke{-}Schneider, and Y.~Wang.
\newblock Chronos: Learning the language of time series.
\newblock \emph{arXiv:2403.07815 [cs.LG]}, 2024.

\bibitem[Athanasopoulos(2021)]{hyndman-book21a}
R.~Hyndmanand~G. Athanasopoulos.
\newblock \emph{Forecasting: principles and practice}.
\newblock OTexts, 3. edition, 2021.

\bibitem[Ba et~al.(2016)Ba, Kiros, and Hinton]{ba-arxiv16a}
J.~Ba, J.~Kiros, and G.~Hinton.
\newblock Layer normalization, 2016.

\bibitem[Bai et~al.(2018)Bai, Kolter, and Koltun]{bai-arxiv18a}
S.~Bai, J.~Kolter, and V.~Koltun.
\newblock An empirical evaluation of generic convolutional and recurrent networks for sequence modeling.
\newblock \emph{arXiv:1803.01271 [cs.LG]}, 2018.

\bibitem[Beitner(2020)]{beitner-github20a}
J.~Beitner.
\newblock {PyTorch Forecasting: Time series forecasting with PyTorch}.
\newblock \url{github.com/jdb78/pytorch-forecasting}, 2020.

\bibitem[Bian et~al.(2024)Bian, Ju, Li, Xu, Cheng, and Xu]{bian-icml24a}
Y.~Bian, X.~Ju, J.~Li, Z.~Xu, D.~Cheng, and Q.~Xu.
\newblock Multi-patch prediction: Adapting llms for time series representation learning.
\newblock In R.~Salakhutdinov, Z.~Kolter, K.~Heller, A.~Weller, N.~Oliver, J.~Scarlett, and F.~Berkenkamp, editors, \emph{Proceedings of the 41st International Conference on Machine Learning ({ICML}'24)}, volume 251 of \emph{Proceedings of Machine Learning Research}. PMLR, 2024.

\bibitem[Box et~al.(2015)Box, Jenkins, Reinsel, and Ljung]{box-book15a}
G.~Box, G.~Jenkins, G.~Reinsel, and G.~Ljung.
\newblock \emph{Time series analysis: forecasting and control}.
\newblock John Wiley \& Sons, 2015.

\bibitem[Brown et~al.(2020)Brown, Mann, Ryder, Subbiah, Kaplan, Dhariwal, Neelakantan, Shyam, Sastry, Askell, Agarwal, Herbert-Voss, Krueger, Henighan, Child, Ramesh, Ziegler, Wu, Winter, Hesse, Chen, Sigler, Litwin, Gray, Chess, Clark, Berner, McCandlish, Radford, Sutskever, and Amodei]{brown-neurips20a}
T.~Brown, B.~Mann, N.~Ryder, M.~Subbiah, J.~Kaplan, P.~Dhariwal, A.~Neelakantan, P.~Shyam, G.~Sastry, A.~Askell, S.~Agarwal, A.~Herbert-Voss, G.~Krueger, T.~Henighan, R.~Child, A.~Ramesh, D.~Ziegler, J.~Wu, C.~Winter, C.~Hesse, M.~Chen, E.~Sigler, M.~Litwin, S.~Gray, B.~Chess, J.~Clark, C.~Berner, S.~McCandlish, A.~Radford, I.~Sutskever, and D.~Amodei.
\newblock Language models are few-shot learners.
\newblock In  \citet{neurips20}, pages 1877--1901.

\bibitem[Chen et~al.(2001)Chen, Petty, Skabardonis, Varaiya, and Jia]{chen-trr01a}
C.~Chen, K.~Petty, A.~Skabardonis, P.~Varaiya, and Z.~Jia.
\newblock Freeway performance measurement system: Mining loop detector data.
\newblock \emph{Transportation Research Record}, 2001.
\newblock URL \url{https://doi.org/10.3141/1748-12}.

\bibitem[Chen et~al.(2021{\natexlab{a}})Chen, Chen, Shang, Zhang, Wen, and Yang]{chen-arxiv21b}
D.~Chen, L.~Chen, Z.~Shang, Y.~Zhang, B.~Wen, and C.~Yang.
\newblock Scale-aware neural architecture search for multivariate time series forecasting, 2021{\natexlab{a}}.

\bibitem[Chen et~al.(2022)Chen, Lin, Sun, and Li]{chen-openr22a}
H.~Chen, M.~Lin, X.~Sun, and H.~Li.
\newblock {NAS}-bench-zero: A large scale dataset for understanding zero-shot neural architecture search, 2022.
\newblock URL \url{https://openreview.net/forum?id=hP-SILoczR}.

\bibitem[Chen et~al.(2021{\natexlab{b}})Chen, Peng, Fu, and Ling]{Chen-iccv21b}
M.~Chen, H.~Peng, J.~Fu, and H.~Ling.
\newblock Autoformer: Searching transformers for visual recognition.
\newblock In \emph{Proceedings of the 24nd IEEE/CVF International Conference on Computer Vision ({ICCV}'21)} \citet{iccv21}, pages 12270--12280.

\bibitem[Chen et~al.(2023)Chen, Li, Arik, Yoder, and Pfister]{chen-tmlr23a}
S.~Chen, C.~Li, S.~Arik, N.~Yoder, and T.~Pfister.
\newblock {TSM}ixer: An all-{MLP} architecture for time series forecast-ing.
\newblock \emph{Transactions on Machine Learning Research}, 2023.
\newblock ISSN 2835-8856.
\newblock URL \url{https://openreview.net/forum?id=wbpxTuXgm0}.

\bibitem[Cho et~al.(2014)Cho, van Merrienboer, G{\"{u}}l{\c{c}}ehre, Bahdanau, Bougares, Schwenk, and Bengio]{cho-emnlp14a}
K.~Cho, B.~van Merrienboer, {\c{C}}.~G{\"{u}}l{\c{c}}ehre, D.~Bahdanau, F.~Bougares, H.~Schwenk, and Y.~Bengio.
\newblock Learning phrase representations using {RNN} encoder-decoder for statistical machine translation.
\newblock In A.~Moschitti, B.~Pang, and W.~Daelemans, editors, \emph{Proceedings of the 2014 Conference on Empirical Methods in Natural Language Processing}, pages 1724--1734. Association for Computational Linguistics, 2014.

\bibitem[Chu et~al.(2020)Chu, Zhou, Zhang, and Li]{chu-eccv20a}
X.~Chu, T.~Zhou, B.~Zhang, and J.~Li.
\newblock Fair darts: Eliminating unfair advantages in differentiable architecture search.
\newblock In A.~Vedaldi, H.~Bischof, T.~Brox, and J.~Frahm, editors, \emph{16th European Conference on Computer Vision ({ECCV}'20)}, pages 465--480. Springer, Springer, 2020.

\bibitem[Chung et~al.(2015)Chung, Kastner, Dinh, Goel, Courville, and Bengio]{chung-arxiv15a}
J.~Chung, K.~Kastner, L.~Dinh, K.~Goel, A.~Courville, and Y.~Bengio.
\newblock A recurrent latent variable model for sequential data.
\newblock \emph{arXiv:1506.02216v6 [cs.LG]}, 2015.

\bibitem[Com(2021)]{iccv21}
\emph{Proceedings of the 24nd IEEE/CVF International Conference on Computer Vision ({ICCV}'21)}, 2021. Computer Vision Foundation and IEEE Computer Society, IEEE.

\bibitem[Das et~al.(2023)Das, Kong, Sen, and Zhou]{das-arxiv23a}
A.~Das, W.~Kong, R.~Sen, and Y.~Zhou.
\newblock A decoder-only foundation model for time-series forecasting.
\newblock \emph{arXiv:2310.10688 [cs.LG]}, 2023.

\bibitem[Deng et~al.(2022)Deng, Karl, Hutter, Bischl, and Lindauer]{deng-ecml22a}
D.~Deng, F.~Karl, F.~Hutter, B.~Bischl, and M.~Lindauer.
\newblock Efficient automated deep learning for time series forecasting.
\newblock In M.-R. Amini, S.~Canu, A.~Fischer, T.~Guns P.~K. Novak, and G.~Tsoumakas, editors, \emph{Machine Learning and Knowledge Discovery in Databases. Research Track - European Conference, {ECML} {PKDD}}. {ACM}, 2022.

\bibitem[Devlin et~al.(2019)Devlin, Chang, Lee, and Toutanova]{devlin-acl19a}
J.~Devlin, M.~Chang, K.~Lee, and K.~Toutanova.
\newblock {BERT}: Pre-training of deep bidirectional transformers for language understanding.
\newblock In J.~Burstein, C.~Doran, and T.~Solorio, editors, \emph{Proceedings of the 2019 Conference of the North {A}merican Chapter of the Association for Computational Linguistics: Human Language Technologies}, pages 4171--4186. Association for Computational Linguistics, 2019.

\bibitem[Didolkar et~al.(2022)Didolkar, Gupta, Goyal, Gundavarapu, Lamb, Ke, and Bengio]{didolkar-neurips22a}
A.~Didolkar, K.~Gupta, A.~Goyal, N.~Gundavarapu, A.~Lamb, N.~Rosemary Ke, and Y.~Bengio.
\newblock Temporal latent bottleneck: Synthesis of fast and slow processing mechanisms in sequence learning.
\newblock In \emph{Advances in Neural Information Processing Systems 35: Annual Conference on Neural Information Processing Systems 2022, NeurIPS}, 2022.

\bibitem[Godahewa et~al.(2021)Godahewa, Bergmeir, Webb, Hyndman, and Montero-Manso]{godahewa-neuripsdbt21a}
R.~Godahewa, C.~Bergmeir, G.~Webb, R.~Hyndman, and P.~Montero-Manso.
\newblock Monash time series forecasting archive.
\newblock In J.~Vanschoren and S.~Yeung, editors, \emph{Proceedings of the Neural Information Processing Systems Track on Datasets and Benchmarks}. Curran Associates, 2021.

\bibitem[Guyon et~al.(2017)Guyon, von Luxburg, Bengio, Wallach, Fergus, Vishwanathan, and Garnett]{nips17}
I.~Guyon, U.~von Luxburg, S.~Bengio, H.~Wallach, R.~Fergus, S.~Vishwanathan, and R.~Garnett, editors.
\newblock \emph{Proceedings of the 31st International Conference on Advances in Neural Information Processing Systems ({N}eur{IPS}'17)}, 2017. Curran Associates.

\bibitem[He et~al.(2016)He, Zhang, Ren, and Sun]{he-cvpr16a}
K.~He, X.~Zhang, S.~Ren, and J.~Sun.
\newblock Deep residual learning for image recognition.
\newblock In \emph{Proceedings of the International Conference on Computer Vision and Pattern Recognition ({CVPR}'16)}, pages 770--778. Computer Vision Foundation and IEEE Computer Society, IEEE, 2016.

\bibitem[Hewamalage et~al.(2021)Hewamalage, Bergmeir, and Bandara]{hewamalage-ijf21a}
H.~Hewamalage, C.~Bergmeir, and K.~Bandara.
\newblock Recurrent neural networks for time series forecasting: Current status and future directions.
\newblock \emph{International Journal of Forecasting}, pages 388--427, 2021.

\bibitem[Hochreiter and Schmidhuber(1997)]{hochreiter-nc97a}
S.~Hochreiter and J.~Schmidhuber.
\newblock {Long Short-Term Memory}.
\newblock \emph{Neural Computation}, 9\penalty0 (8):\penalty0 1735--1780, 1997.
\newblock Based on TR FKI-207-95, TUM (1995).

\bibitem[Hochreiter et~al.(2001)Hochreiter, Younger, and Conwell]{hochreiter-icann01a}
S.~Hochreiter, A.~Younger, and P.~Conwell.
\newblock Learning to learn using gradient descent.
\newblock In G.~Dorffner, H.~Bischof, and K.~Hornik, editors, \emph{Proceedings of the 11th International Conference on Artificial Neural Networks ({ICANN}'01)}, pages 87--94. Springer, 2001.

\bibitem[Hvarfner et~al.(2022)Hvarfner, Stoll, Souza, Nardi, Lindauer, and Hutter]{hvarfner-iclr22a}
C.~Hvarfner, D.~Stoll, A.~Souza, L.~Nardi, M.~Lindauer, and F.~Hutter.
\newblock $\pi${BO}: {A}ugmenting {A}cquisition {F}unctions with {U}ser {B}eliefs for {B}ayesian {O}ptimization.
\newblock In \emph{The Tenth International Conference on Learning Representations ({ICLR}'22)} \citet{iclr22}.

\bibitem[ICL(2017)]{iclr17}
\emph{Proceedings of the International Conference on Learning Representations ({ICLR}'17)}, 2017. ICLR.

\bibitem[ICL(2019)]{iclr19}
\emph{Proceedings of the International Conference on Learning Representations ({ICLR}'19)}, 2019. ICLR.

\bibitem[ICL(2020)]{iclr20}
\emph{Proceedings of the International Conference on Learning Representations ({ICLR}'20)}, 2020. ICLR.

\bibitem[ICL(2021)]{iclr21}
\emph{Proceedings of the International Conference on Learning Representations ({ICLR}'21)}, 2021. ICLR.

\bibitem[ICL(2022)]{iclr22}
\emph{Proceedings of the International Conference on Learning Representations ({ICLR}'22)}, 2022. ICLR.

\bibitem[ICL(2023)]{iclr23}
\emph{Proceedings of the International Conference on Learning Representations ({ICLR}'23)}, 2023. ICLR.

\bibitem[ICL(2024)]{iclr24}
\emph{Proceedings of the International Conference on Learning Representations ({ICLR}'24)}, 2024. ICLR.

\bibitem[Ioffe and Szegedy(2015)]{ioffe-icml15a}
S.~Ioffe and C.~Szegedy.
\newblock Batch normalization: Accelerating deep network training by reducing internal covariate shift.
\newblock In F.~Bach and D.~Blei, editors, \emph{Proceedings of the 32nd International Conference on Machine Learning ({ICML}'15)}, volume~37. Omnipress, 2015.

\bibitem[Jiang et~al.(2023)Jiang, Ji, Zhu, Yuan, and Huang]{jiang-neurips23a}
S.~Jiang, Z.~Ji, G.~Zhu, C.~Yuan, and Y.~Huang.
\newblock Operation-level early stopping for robustifying differentiable {NAS}.
\newblock In \emph{Proceedings of the 36th International Conference on Advances in Neural Information Processing Systems ({N}eur{IPS}'23)}, 2023.
\newblock URL \url{https://openreview.net/forum?id=yAOwkf4FyL}.

\bibitem[Jin et~al.(2019)Jin, Song, and Hu]{jin-sigkdd19a}
H.~Jin, Q.~Song, and X.~Hu.
\newblock {Auto-Keras}: An efficient neural architecture search system.
\newblock In A.~Teredesai, V.~Kumar, Y.~Li, R.~Rosales, E.~Terzi, and G.~Karypis, editors, \emph{Proceedings of the 25th {ACM} {SIGKDD} International Conference on Knowledge Discovery {\&} Data Mining ({KDD}'19)}, pages 1946--1956. ACM Press, 2019.

\bibitem[Ke et~al.(2017)Ke, Meng, Finley, Wang, Chen, Ma, Ye, and Liu]{ke-neurips17a}
G.~Ke, Q.~Meng, T.~Finley, T.~Wang, W.~Chen, W.~Ma, Q.~Ye, and T.-Y. Liu.
\newblock Lightgbm: A highly efficient gradient boosting decision tree.
\newblock In  \citet{nips17}.

\bibitem[Kim et~al.(2022)Kim, Kim, Tae, Park, Choi, and Choo]{kim-iclr22a}
T.~Kim, J.~Kim, Y.~Tae, C.~Park, J.~Choi, and J.~Choo.
\newblock Reversible instance normalization for accurate time-series forecasting against distribution shift.
\newblock In \emph{The Tenth International Conference on Learning Representations ({ICLR}'22)} \citet{iclr22}.

\bibitem[Klyuchnikov et~al.(2020)Klyuchnikov, Trofimov, Artemova, Salnikov, Fedorov, and Burnaev]{klyuchnikov-arxiv20a}
N.~Klyuchnikov, I.~Trofimov, E.~Artemova, M.~Salnikov, M.~Fedorov, and E.~Burnaev.
\newblock {NAS-Bench-NLP}: {N}eural {A}rchitecture {S}earch benchmark for {N}atural {L}anguage {P}rocessing.
\newblock \emph{arXiv:2006.07116v1 [cs.LG]}, 2020.

\bibitem[Krishnakumar et~al.(2022)Krishnakumar, White, Zela, R.~Tu, Safari, and Hutter]{krishnakumar-neurips22a}
A.~Krishnakumar, C.~White, A.~Zela, Renbo R.~Tu, M.~Safari, and F.~Hutter.
\newblock Nas-bench-suite-zero: Accelerating research on zero cost proxies.
\newblock In S.~Koyejo, S.~Mohamed, A.~Agarwal, D.~Belgrave, K.~Cho, and A.~Oh, editors, \emph{Proceedings of the 35th International Conference on Advances in Neural Information Processing Systems ({N}eur{IPS}'22)}. Curran Associates, 2022.

\bibitem[Lai et~al.(2018)Lai, Chang, Yang, and Liu]{lai-sigir18a}
G.~Lai, W.~Chang, Y.~Yang, and H.~Liu.
\newblock Modeling long- and short-term temporal patterns with deep neural networks.
\newblock In K.~Thompson, Q.~Mei, B.Davison, Y.~Liu, and E.~Yilmaz, editors, \emph{International {ACM} {SIGIR} Conference on Research {\&} Development in Information Retrieval}, pages 95--104. ACM, 2018.
\newblock URL \url{https://doi.org/10.1145/3209978.3210006}.

\bibitem[Lana et~al.(2018)Lana, Ser, V{\'{e}}lez, and Vlahogianni]{lana-itsm18a}
I.~Lana, J.~Del Ser, M.~V{\'{e}}lez, and E.~Vlahogianni.
\newblock Road traffic forecasting: Recent advances and new challenges.
\newblock \emph{{IEEE} Intell. Transp. Syst. Mag.}, 10\penalty0 (2):\penalty0 93--109, 2018.

\bibitem[Larochelle et~al.(2020)Larochelle, Ranzato, Hadsell, Balcan, and Lin]{neurips20}
H.~Larochelle, M.~Ranzato, R.~Hadsell, M.-F. Balcan, and H.~Lin, editors.
\newblock \emph{Proceedings of the 33rd International Conference on Advances in Neural Information Processing Systems ({N}eur{IPS}'20)}, 2020. Curran Associates.

\bibitem[Lee et~al.(2019)Lee, Ajanthan, and Torr]{lee-iclr19a}
N.~Lee, T.~Ajanthan, and P.~Torr.
\newblock Snip: single-shot network pruning based on connection sensitivity.
\newblock In \emph{The Seventh International Conference on Learning Representations ({ICLR}'19)} \citet{iclr19}.

\bibitem[Li et~al.(2019)Li, Jin, Xuan, Zhou, Chen, Wang, and Yan]{li-neurips19a}
S.~Li, X.~Jin, Y.~Xuan, X.~Zhou, W.~Chen, Y.~Wang, and X.~Yan.
\newblock Enhancing the locality and breaking the memory bottleneck of transformer on time series forecasting.
\newblock In  \citet{neurips19}, pages 5244--5254.

\bibitem[Lim et~al.(2021)Lim, Arık, Loeff, and Pfister]{lim-ijf21a}
B.~Lim, S.~Arık, N.~Loeff, and T.~Pfister.
\newblock Temporal fusion transformers for interpretable multi-horizon time series forecasting.
\newblock \emph{International Journal of Forecasting}, 37\penalty0 (4):\penalty0 1748--1764, 2021.

\bibitem[Lin et~al.(2021)Lin, Wang, Sun, Chen, Sun, Qian, Li, and Jin]{lin21-iccv21a}
M.~Lin, P.~Wang, Z.~Sun, H.~Chen, X.~Sun, Q.~Qian, H.~Li, and R.~Jin.
\newblock Zen-nas: A zero-shot nas for high-performance image recognition.
\newblock In \emph{Proceedings of the 24nd IEEE/CVF International Conference on Computer Vision ({ICCV}'21)} \citet{iccv21}, pages 347--356.

\bibitem[Liu et~al.(2018)Liu, Simonyan, Vinyals, Fernando, and Kavukcuoglu]{liu-iclr18a}
H.~Liu, K.~Simonyan, O.~Vinyals, C.~Fernando, and K.~Kavukcuoglu.
\newblock Hierarchical representations for efficient architecture search.
\newblock In \emph{The Sixth International Conference on Learning Representations ({ICLR}'18)}. ICLR, 2018.

\bibitem[Liu et~al.(2019{\natexlab{a}})Liu, Simonyan, and Yang]{liu-iclr19a}
H.~Liu, K.~Simonyan, and Y.~Yang.
\newblock {DARTS}: Differentiable architecture search.
\newblock In \emph{The Seventh International Conference on Learning Representations ({ICLR}'19)} \citet{iclr19}.

\bibitem[Liu et~al.(2022)Liu, Yu, Liao, Li, Lin, Liu, and Dustdar]{liu-iclr22a}
S.~Liu, H.~Yu, C.~Liao, J.~Li, W.~Lin, A.~Liu, and S.~Dustdar.
\newblock Pyraformer: Low-complexity pyramidal attention for long-range time series modeling and forecasting.
\newblock In \emph{The Tenth International Conference on Learning Representations ({ICLR}'22)} \citet{iclr22}.
\newblock URL \url{https://openreview.net/forum?id=0EXmFzUn5I}.

\bibitem[Liu et~al.(2024{\natexlab{a}})Liu, Liu, Woo, Aksu, Liang, Zimmermann, Liu, Savarese, Xiong, and Sahoo]{liu-arxiv24a}
X.~Liu, J.~Liu, G.~Woo, T.~Aksu, Y.~Liang, R.~Zimmermann, C.~Liu, S.~Savarese, C.~Xiong, and D.~Sahoo.
\newblock Moirai-moe: Empowering time series foundation models with sparse mixture of experts.
\newblock \emph{arXiv:2410.10469[cs.LG]}, 2024{\natexlab{a}}.

\bibitem[Liu et~al.(2019{\natexlab{b}})Liu, Ott, Goyal, Du, Joshi, Chen, Levy, Lewis, Zettlemoyer, and Stoyanov]{liu-arxiv19a}
Y.~Liu, M.~Ott, N.~Goyal, J.~Du, M.~Joshi, D.~Chen, O.~Levy, M.~Lewis, L.~Zettlemoyer, and V.~Stoyanov.
\newblock Roberta: A robustly optimized bert pretraining approach.
\newblock \emph{arXiv:1907.11692 [cs.CL]}, 2019{\natexlab{b}}.

\bibitem[Liu et~al.(2024{\natexlab{b}})Liu, Hu, Zhang, Wu, Wang, Ma, and Long]{liu-iclr24c}
Y.~Liu, T.~Hu, H.~Zhang, H.~Wu, S.~Wang, L.~Ma, and M.~Long.
\newblock itransformer: Inverted transformers are effective for time series forecasting.
\newblock In \emph{The Twelfth International Conference on Learning Representations ({ICLR}'24)} \citet{iclr24}.

\bibitem[Liu et~al.(2021)Liu, Lin, Cao, Hu, Wei, Zhang, Lin, and Guo]{liu-iccv21a}
Z.~Liu, Y.~Lin, Y.~Cao, H.~Hu, Y.~Wei, Z.~Zhang, S.~Lin, and B.~Guo.
\newblock Swin transformer: Hierarchical vision transformer using shifted windows.
\newblock In \emph{Proceedings of the 24nd IEEE/CVF International Conference on Computer Vision ({ICCV}'21)} \citet{iccv21}, pages 10012--10022.

\bibitem[Lopes et~al.(2023)Lopes, Degardin, and Alexandre]{lopes-arxiv23a}
V.~Lopes, B.~Degardin, and L.~Alexandre.
\newblock Are neural architecture search benchmarks well designed? a deeper look into operation importance.
\newblock \emph{arXiv:2303.16938[cs.LG]}, 2023.

\bibitem[Loshchilov and Hutter(2017)]{loshchilov-iclr17a}
I.~Loshchilov and F.~Hutter.
\newblock {SGDR}: Stochastic gradient descent with warm restarts.
\newblock In \emph{The Fifth International Conference on Learning Representations ({ICLR}'17)} \citet{iclr17}.

\bibitem[Luo and Wang(2024)]{luo-iclr24a}
D.~Luo and X.~Wang.
\newblock Modern{TCN}: A modern pure convolution structure for general time series analysis.
\newblock In \emph{The Twelfth International Conference on Learning Representations ({ICLR}'24)} \citet{iclr24}.
\newblock URL \url{https://openreview.net/forum?id=vpJMJerXHU}.

\bibitem[Lyu et~al.(2023)Lyu, Ororbia, and Desell]{lyu-asc23a}
Z.~Lyu, A.~Ororbia, and T.~Desell.
\newblock Online evolutionary neural architecture search for multivariate non-stationary time series forecasting.
\newblock \emph{Appl. Soft Comput.}, 2023.
\newblock URL \url{https://doi.org/10.1016/j.asoc.2023.110522}.

\bibitem[Makridakis et~al.(2020)Makridakis, Spiliotis, and Assimakopoulos]{makridakis-ijf20a}
S.~Makridakis, E.~Spiliotis, and V.~Assimakopoulos.
\newblock The m4 competition: 100,000 time series and 61 forecasting methods.
\newblock \emph{International Journal of Forecasting}, pages 54--74, 2020.

\bibitem[Makridakis et~al.(2022)Makridakis, Spiliotis, and Assimakopoulos]{makridakis-ijf22a}
S.~Makridakis, E.~Spiliotis, and V.~Assimakopoulos.
\newblock M5 accuracy competition: Results, findings, and conclusions.
\newblock \emph{International Journal of Forecasting}, pages 1346--1364, 2022.
\newblock URL \url{https://www.sciencedirect.com/science/article/pii/S0169207021001874}.

\bibitem[Mallik et~al.(2023)Mallik, Bergman, Hvarfner, Stoll, Janowski, Lindauer, Nardi, and Hutter]{malik-neurips23a}
N.~Mallik, E.~Bergman, C.~Hvarfner, D.~Stoll, M.~Janowski, M.~Lindauer, L.~Nardi, and F.~Hutter.
\newblock Priorband: practical hyperparameter optimization in the age of deep learning.
\newblock In A.~Oh, T.~Naumann, A.~Globerson, K.~Saenko, M.~Hardt, and S.~Levine, editors, \emph{Proceedings of the 36th International Conference on Advances in Neural Information Processing Systems ({N}eur{IPS}'23)}. Curran Associates, 2023.

\bibitem[Mehta et~al.(2022)Mehta, White, Zela, Krishnakumar, Zabergja, Moradian, Safari, Yu, and Hutter]{mehta-iclr22a}
Y.~Mehta, C.~White, A.~Zela, A.~Krishnakumar, G.~Zabergja, S.~Moradian, M.~Safari, K.~Yu, and F.~Hutter.
\newblock {NAS-Bench-Suite}: {NAS} evaluation is (now) surprisingly easy.
\newblock In \emph{The Tenth International Conference on Learning Representations ({ICLR}'22)} \citet{iclr22}.

\bibitem[Mellor et~al.(2020)Mellor, Turner, Storkey, and Crowley]{mellor-icml20a}
J.~Mellor, J.~Turner, A.~Storkey, and E.~Crowley.
\newblock {N}eural {A}rchitecture {S}earch without training.
\newblock In H.~{Daume III} and A.~Singh, editors, \emph{Proceedings of the 37th International Conference on Machine Learning ({ICML}'20)}, volume~98. Proceedings of Machine Learning Research, 2020.

\bibitem[Nie et~al.(2023)Nie, Nguyen, Sinthong, and Kalagnanam]{nie-iclr23a}
Y.~Nie, N.~Nguyen, P.~Sinthong, and J.~Kalagnanam.
\newblock A time series is worth 64 words: Long-term forecasting with transformers.
\newblock In \emph{The Eleventh International Conference on Learning Representations ({ICLR}'23)} \citet{iclr23}.
\newblock URL \url{https://openreview.net/pdf?id=Jbdc0vTOcol}.

\bibitem[Ning et~al.(2021)Ning, Tang, Li, Zhou, Liang, Yang, and Wang]{ning-neurips21a}
X.~Ning, C.~Tang, W.~Li, Z.~Zhou, S.~Liang, H.~Yang, and Y.~Wang.
\newblock Evaluating efficient performance estimators of neural architectures.
\newblock In  \citet{neurips21}.

\bibitem[Oord et~al.(2016)Oord, Dieleman, Zen, Simonyan, Vinyals, Graves, Kalchbrenner, Senior, and Kavukcuoglu]{Oord-iscassw16a}
A.~Oord, S.~Dieleman, H.~Zen, K.~Simonyan, O.~Vinyals, A.~Graves, N.~Kalchbrenner, A.~Senior, and K.~Kavukcuoglu.
\newblock Wavenet: {A} generative model for raw audio.
\newblock In \emph{the 9th {ISCA} Speech Synthesis Workshop}, page 125, 2016.

\bibitem[Oreshkin et~al.(2020)Oreshkin, Carpov, Chapados, and Bengio]{oreshkin-iclr20a}
B.~Oreshkin, D.~Carpov, N.~Chapados, and Y.~Bengio.
\newblock {N-BEATS:} neural basis expansion analysis for interpretable time series forecasting.
\newblock In \emph{The Eigth International Conference on Learning Representations ({ICLR}'20)} \citet{iclr20}.

\bibitem[Pan et~al.(2023)Pan, Cai, and Zhuang]{pan-cvpr23a}
Z.~Pan, J.~Cai, and B.~Zhuang.
\newblock Stitchable neural networks.
\newblock In \emph{Proceedings of the International Conference on Computer Vision and Pattern Recognition ({CVPR}'23)}, pages 16102--16112. Computer Vision Foundation and IEEE Computer Society, IEEE, 2023.
\newblock URL \url{https://doi.org/10.1109/CVPR52729.2023.01545}.

\bibitem[Paszke et~al.(2019)Paszke, Gross, Massa, Lerer, et~al.]{pytorch-neurips19a}
A.~Paszke, S.~Gross, F.~Massa, A.~Lerer, et~al.
\newblock {PyTorch}: An imperative style, high-performance deep learning library.
\newblock In  \citet{neurips19}, pages 8024--8035.

\bibitem[Pham et~al.(2018)Pham, Guan, Zoph, Le, and Dean]{pham-icml18a}
H.~Pham, M.~Guan, B.~Zoph, Q.~Le, and J.~Dean.
\newblock Efficient {N}eural {A}rchitecture {S}earch via parameter sharing.
\newblock In J.~Dy and A.~Krause, editors, \emph{Proceedings of the 35th International Conference on Machine Learning ({ICML}'18)}, volume~80. Proceedings of Machine Learning Research, 2018.

\bibitem[Ranzato et~al.(2021)Ranzato, Beygelzimer, Nguyen, Liang, Vaughan, and Dauphin]{neurips21}
M.~Ranzato, A.~Beygelzimer, K.~Nguyen, P.~Liang, J.~Vaughan, and Y.~Dauphin, editors.
\newblock \emph{Proceedings of the 34th International Conference on Advances in Neural Information Processing Systems ({N}eur{IPS}'21)}, 2021. Curran Associates.

\bibitem[Ronneberger et~al.(2015)Ronneberger, Fischer, and Brox]{ronneberger-miccai15a}
O.~Ronneberger, P.~Fischer, and T.~Brox.
\newblock {U-Net} convolutional networks for biomedical image segmentation.
\newblock \emph{Medical Image Computing and Computer-Assisted Intervention}, pages 234--241, 2015.

\bibitem[Salinas et~al.(2020)Salinas, Flunkert, Gasthaus, and Januschowski]{salinas-ijf20a}
D.~Salinas, V.~Flunkert, J.~Gasthaus, and T.~Januschowski.
\newblock Deepar: Probabilistic forecasting with autoregressive recurrent networks.
\newblock \emph{International Journal of Forecasting}, pages 1181--1191, 2020.

\bibitem[Shi et~al.(2015)Shi, Chen, Wang, Yeung, Wong, and Woo]{shi-neurips15a}
X.~Shi, Z.~Chen, H.~Wang, D.~Yeung, W.~Wong, and W.~Woo.
\newblock Convolutional {LSTM} network: {A} machine learning approach for precipitation nowcasting.
\newblock In C.~Cortes, N.~Lawrence, D.~Lee, M.~Sugiyama, and R.~Garnett, editors, \emph{Proceedings of the 29th International Conference on Advances in Neural Information Processing Systems ({N}eur{IPS}'15)}, pages 802--810. Curran Associates, 2015.

\bibitem[Srivastava et~al.(2014)Srivastava, Hinton, Krizhevsky, Sutskever, and Salakhutdinov]{srivastava-jmlr14a}
N.~Srivastava, G.~Hinton, A.~Krizhevsky, I.~Sutskever, and R.~Salakhutdinov.
\newblock Dropout: A simple way to prevent neural networks from overfitting.
\newblock \emph{Journal of Machine Learning Research}, 15:\penalty0 1929--1958, 2014.

\bibitem[Sukthanker et~al.(2024)Sukthanker, Zela, Staffler, Dooley, Grabocka, and Hutter]{sukthanker-arxiv24a}
R.~Sukthanker, A.~Zela, B.~Staffler, S.~Dooley, J.~Grabocka, and Frank Hutter.
\newblock Multi-objective differentiable neural architecture search.
\newblock \emph{arXiv:2402.18213[cs.LG]}, 2024.

\bibitem[Sutskever et~al.(2014)Sutskever, Vinyals, and Le]{sutskever-neurips14a}
I.~Sutskever, O.~Vinyals, and Q.~Le.
\newblock Sequence to sequence learning with neural networks.
\newblock In Z.~Ghahramani, M.~Welling, C.~Cortes, N.~Lawrence, and K.~Weinberger, editors, \emph{Proceedings of the 28th International Conference on Advances in Neural Information Processing Systems ({N}eur{IPS}'14)}. Curran Associates, 2014.

\bibitem[Tanaka et~al.(2020)Tanaka, Kunin, Yamins, and Ganguli]{tanaka-neurips20a}
H.~Tanaka, D.~Kunin, D.~L. Yamins, and S.~Ganguli.
\newblock Pruning neural networks without any data by iteratively conserving synaptic flow.
\newblock In  \citet{neurips20}.

\bibitem[Trindade(2015)]{trindade-uci15a}
A.~Trindade.
\newblock {ElectricityLoadDiagrams20112014}.
\newblock UCI Machine Learning Repository, 2015.
\newblock {DOI}: https://doi.org/10.24432/C58C86.

\bibitem[Turner(2019)]{turner-19a}
R.~Turner.
\newblock The bayes opt benchmark documentation.
\newblock \url{https://bayesmark.readthedocs.io/en/latest/}, 2019.

\bibitem[Vaswani et~al.(2017)Vaswani, Shazeer, Parmar, Uszkoreit, Jones, Gomez, Kaiser, and Polosukhin]{Vaswani-neurips17a}
A.~Vaswani, N.~Shazeer, N.~Parmar, J.~Uszkoreit, L.~Jones, A.~Gomez, L.~Kaiser, and I.~Polosukhin.
\newblock Attention is all you need.
\newblock In  \citet{nips17}.

\bibitem[Wallach et~al.(2019)Wallach, Larochelle, Beygelzimer, d'Alche Buc, Fox, and Garnett]{neurips19}
H.~Wallach, H.~Larochelle, A.~Beygelzimer, F.~d'Alche Buc, E.~Fox, and R.~Garnett, editors.
\newblock \emph{Proceedings of the 32nd International Conference on Advances in Neural Information Processing Systems ({N}eur{IPS}'19)}, 2019. Curran Associates.

\bibitem[Wan et~al.(2022)Wan, Ru, Esperança, and Li]{wan-iclr22a}
X.~Wan, B.~Ru, P.~Esperança, and Z.~Li.
\newblock On redundancy and diversity in cell-based neural architecture search.
\newblock In \emph{The Tenth International Conference on Learning Representations ({ICLR}'22)} \citet{iclr22}.

\bibitem[Wang et~al.(2019)Wang, Zhang, and Grosse]{wang-iclr19a}
C.~Wang, G.~Zhang, and R.~Grosse.
\newblock Picking winning tickets before training by preserving gradient flow.
\newblock In \emph{International Conference on Learning Representations} \citet{iclr19}.

\bibitem[Wang et~al.(2021)Wang, Cheng, Chen, Tang, and Hsieh]{wang-iclr21a}
R.~Wang, M.~Cheng, X.~Chen, X.~Tang, and C.~Hsieh.
\newblock Rethinking architecture selection in differentiable nas.
\newblock In \emph{International Conference on Learning Representation} \citet{iclr21}.

\bibitem[Wen et~al.(2017)Wen, Torkkola, Narayanaswamy, and Madeka]{wen-tsw17a}
R.~Wen, K.~Torkkola, B.~Narayanaswamy, and D.~Madeka.
\newblock A multi-horizon quantile recurrent forecaster.
\newblock In \emph{31st Conference on Neural Information Processing Systems, Time Series Workshop}, 2017.

\bibitem[White et~al.(2021)White, Neiswanger, and Savani]{white-aaai21a}
C.~White, W.~Neiswanger, and Y.~Savani.
\newblock {BANANAS}: {Bayesian} optimization with neural architectures for neural architecture search.
\newblock In  \citet{aaai21}, pages 10293--10301.

\bibitem[Wu et~al.(2021)Wu, Xu, Wang, and Long]{wu-neurips21a}
H.~Wu, J.~Xu, J.~Wang, and M.~Long.
\newblock Autoformer: Decomposition transformers with {Auto-Correlation} for long-term series forecasting.
\newblock In  \citet{neurips21}.

\bibitem[Wu et~al.(2023)Wu, Hu, Liu, Zhou, Wang, and Long]{wu-iclr23a}
H.~Wu, T.~Hu, Y.~Liu, H.~Zhou, J.~Wang, and M.~Long.
\newblock Timesnet: Temporal 2d-variation modeling for general time series analysis.
\newblock In \emph{The Eleventh International Conference on Learning Representations ({ICLR}'23)} \citet{iclr23}.
\newblock URL \url{https://openreview.net/pdf?id=ju\_Uqw384Oq}.

\bibitem[Yang et~al.(2021)Yang, Leyton-Brown, and Mausam]{aaai21}
Q.~Yang, K.~Leyton-Brown, and Mausam, editors.
\newblock \emph{Proceedings of the Thirty-Fifth Conference on Artificial Intelligence ({AAAI}'21)}, 2021. Association for the Advancement of Artificial Intelligence, {AAAI} Press.

\bibitem[Ying et~al.(2019)Ying, Klein, Christiansen, Real, Murphy, and Hutter]{ying-icml19a}
C.~Ying, A.~Klein, E.~Christiansen, E.~Real, K.~Murphy, and F.~Hutter.
\newblock {NAS-Bench-101}: Towards reproducible {N}eural {A}rchitecture {S}earch.
\newblock In K.~Chaudhuri and R.~Salakhutdinov, editors, \emph{Proceedings of the 36th International Conference on Machine Learning ({ICML}'19)}, volume~97, pages 7105--7114. Proceedings of Machine Learning Research, 2019.

\bibitem[Zela et~al.(2020)Zela, Elsken, Saikia, Marrakchi, Brox, and Hutter]{zela-iclr20a}
A.~Zela, T.~Elsken, T.~Saikia, Y.~Marrakchi, T.~Brox, and F.~Hutter.
\newblock Understanding and robustifying differentiable architecture search.
\newblock In \emph{The Eigth International Conference on Learning Representations ({ICLR}'20)} \citet{iclr20}.
\newblock URL \url{https://openreview.net/forum?id=H1gDNyrKDS}.

\bibitem[Zeng et~al.(2023)Zeng, Chen, Zhang, and Xu]{zeng-aaai23a}
A.~Zeng, M.~Chen, L.~Zhang, and Q.~Xu.
\newblock Are transformers effective for time series forecasting?
\newblock In B.~Williams, S.~Bernardini, Y.~Chen, and J.~Neville, editors, \emph{Proceedings of the Thirty-Seventh Conference on Artificial Intelligence ({AAAI}'23)}, pages 11121--11128. Association for the Advancement of Artificial Intelligence, {AAAI} Press, 2023.
\newblock URL \url{https://doi.org/10.1609/aaai.v37i9.26317}.

\bibitem[Zhou et~al.(2021)Zhou, Zhang, Peng, Zhang, Li, Xiong, and Zhang]{zhou-aaai21a}
Haoyi Zhou, Shanghang Zhang, Jieqi Peng, Shuai Zhang, Jianxin Li, Hui Xiong, and Wancai Zhang.
\newblock Informer: Beyond efficient transformer for long sequence time-series forecasting.
\newblock In  \citet{aaai21}.
\newblock URL \url{https://doi.org/10.1609/aaai.v35i12.17325}.

\bibitem[Zimmer et~al.(2021)Zimmer, Lindauer, and Hutter]{zimmer-tpami21a}
L.~Zimmer, M.~Lindauer, and F.~Hutter.
\newblock {Auto-Pytorch}: Multi-fidelity metalearning for efficient and robust {AutoDL}.
\newblock \emph{IEEE Transactions on Pattern Analysis and Machine Intelligence}, 43:\penalty0 3079--3090, 2021.

\bibitem[Zoph and Le(2017)]{zoph-iclr17a}
B.~Zoph and Q.~V. Le.
\newblock {N}eural {A}rchitecture {S}earch with reinforcement learning.
\newblock In \emph{The Fifth International Conference on Learning Representations ({ICLR}'17)} \citet{iclr17}.

\bibitem[Zoph et~al.(2018)Zoph, Vasudevan, Shlens, and Le]{zoph-cvpr18a}
B.~Zoph, V.~Vasudevan, J.~Shlens, and Q.~Le.
\newblock Learning transferable architectures for scalable image recognition.
\newblock In \emph{Proceedings of the International Conference on Computer Vision and Pattern Recognition ({CVPR}'18)}. Computer Vision Foundation and IEEE Computer Society, IEEE, 2018.

\end{thebibliography}
